\documentclass{article} % For LaTeX2e
\usepackage[table,dvipsnames]{xcolor}
\usepackage{iclr2025_conference,times}

\usepackage{amsmath,amsfonts,bm}

\def\eqref#1{equation~\ref{#1}}
\def\1{\bm{1}}

\DeclareMathAlphabet{\mathsfit}{\encodingdefault}{\sfdefault}{m}{sl}
\SetMathAlphabet{\mathsfit}{bold}{\encodingdefault}{\sfdefault}{bx}{n}

\newcommand{\E}{\mathbb{E}}

\newcommand{\KL}{D_{\mathrm{KL}}}

\usepackage{scalefnt,letltxmacro}
\LetLtxMacro{\oldtextsc}{\textsc}
\renewcommand{\textsc}[1]{\oldtextsc{\scalefont{1.10}#1}}
\usepackage[acronym,nowarn]{glossaries}
\glsdisablehyper
\makeglossaries
\usepackage{xspace}
\usepackage{float}
\usepackage{physics}
\usepackage{soul}
\usepackage[normalem]{ulem}

\usepackage{enumitem}

\usepackage{wasysym}
\usepackage{mathtools}
\usepackage{amsfonts}
\usepackage{amsmath}
\usepackage{amsthm,thmtools}
\usepackage{booktabs}
\usepackage[]{microtype}

\usepackage{hyperref}

\usepackage[capitalize,nameinlink,]{cleveref} %noabbrev
\crefname{section}{\S}{\S\S}
\Crefname{section}{\S}{\S\S}
\makeatletter
\@ifundefined{c@rownum}{%
	\let\c@rownum\rownum
}{}
\@ifundefined{therownum}{%
	\def\therownum{\@arabic\rownum}%
}{}
\makeatother

\makeatletter
\newcommand*{\addFileDependency}[1]{% argument=file name and extension
	\typeout{(#1)}
	\@addtofilelist{#1}
	\IfFileExists{#1}{}{\typeout{No file #1.}}
}
\makeatother

\usepackage[font=footnotesize,labelfont=bf,tableposition=top]{caption}
\usepackage{tikz}
\usepackage{graphicx}
\usepackage[]{subcaption}   %% Already defined

\usepackage{pgfplots}
\pgfplotsset{compat=1.6}
\usepgfplotslibrary{groupplots}
\tikzstyle{every picture}+=[font=\sffamily]
\tikzstyle{optimized} = [circle,fill=white,draw=black, dashed,inner sep=1pt, minimum size=20pt, font=\fontsize{10}{10}\selectfont, node distance=1]
\pgfkeys{/pgf/number format/.cd,1000 sep={}}
\pgfplotsset{
	tick label style = {font=\sffamily},
	every axis label/.append style={font=\sffamily},
	typeset ticklabels with strut,
}
\pgfplotsset{every axis/.append style={
			every x tick label/.append style={font=\fontsize{6pt}{6pt}\sffamily, yshift=.5ex,},
			every y tick label/.append style={font=\fontsize{6pt}{6pt}\sffamily, xshift=.5ex},
			every y label/.append style={xshift=10ex, font=\sffamily},
			every x label/.append style={yshift=3ex, font=\sffamily},
			every title/.append style={font=\sffamily}
		},
}
\pgfplotsset{
	xticklabel={$\mathsf{\pgfmathprintnumber{\tick}}$},
	yticklabel={$\mathsf{\pgfmathprintnumber{\tick}}$},
}
\pgfplotsset{every axis title/.append style={yshift=-1ex}}
\newlength\figureheight
\newlength\figurewidth
\usepgfplotslibrary{external}
\tikzexternaldisable

\newcommand{\tinytiny}[1]{\fontsize{7}{10}\selectfont #1}
\newcommand{\TINY}[1]{\fontsize{8.5}{8.5}\selectfont #1}

\usepackage[colorinlistoftodos,
	textsize=scriptsize,
	linecolor=red!30,
	bordercolor=red!30,
	backgroundcolor=red!10]{todonotes}
\renewcommand{\todo}[2][]{\tikzexternaldisable\@todo[#1]{#2}\tikzexternalenable}
\usepackage[most]{tcolorbox}
{\begin{tcolorbox}[toprule=2mm,left=4pt,right=4pt,top=0pt,bottom=1pt,boxsep=2pt, before skip = 2ex, after skip =2ex]}{\end{tcolorbox}}
\tcbset{boxsep=4pt,left=2pt,right=2pt,top=-0pt,bottom=0pt}

\usepackage{url}

\usepackage{subcaption}

\usepackage{tikz}
\usepackage{graphicx}
\usepackage{amsmath}
\usepackage{amssymb}
\usepackage{bm}
\usepackage{amsthm} % theorems and proofs
\usepackage{nicefrac}

\usepackage[normalem]{ulem}

\usepackage{adjustbox}
\usepackage{multirow}
\usepackage{mathtools}

\usepackage{xspace}

\usepackage[]{mdframed}

\usepackage{multirow} 
\usepackage[ruled,linesnumbered]{algorithm2e} 
\usepackage{amsmath}
\usepackage{float}
\usepackage{colortbl}
\usepackage{caption}
\usepackage[utf8]{inputenc} % allow utf-8 input
\usepackage[T1]{fontenc}    % use 8-bit T1 fonts
\usepackage{hyperref}       % hyperlinks
\usepackage{url}            % simple URL typesetting
\usepackage{booktabs}       % professional-quality tables
\usepackage{amsfonts}       % blackboard math symbols
\usepackage{nicefrac}       % compact symbols for 1/2, etc.
\usepackage{microtype}      % microtypography
\usepackage{xcolor}         % colors
\usepackage{makecell}      % make cells in table i.e., to use


\usepackage{utfsym}
\usepackage{setspace}

\hypersetup{
linktocpage=true,
linktoc=page
}

\newlength{\commentWidth}
\usepackage[page,header]{appendix}
\usepackage{titletoc}
\usepackage{geometry}

\newcommand{\CELCOLALT}[1]{\cellcolor{gray!30}#1}

\newcommand{\GRAY}[1]{\textcolor{gray!80}{#1}}

\newcommand{\UL}[1]{#1}
\newcommand{\DIA}{$\diamond$}

\definecolor{GRADLEVEL1}{HTML}{ff99c1}
\definecolor{GRADLEVEL2}{HTML}{ffc2c4}
\definecolor{GRADLEVEL3}{HTML}{ffdf68}
\definecolor{GRADLEVEL4}{HTML}{fffa0f}
\definecolor{GRADLEVEL5}{HTML}{fff100}
\definecolor{GRADLEVEL6}{HTML}{ffe000}
\definecolor{GRADLEVEL7}{HTML}{d8d50a}
\definecolor{GRADLEVEL8}{HTML}{84d11e}
\definecolor{GRADLEVEL9}{HTML}{32cd32}
\definecolor{GRADLEVEL10}{HTML}{32cd32}
\colorlet{GRAD1}{GRADLEVEL1!50}
\colorlet{GRAD2}{GRADLEVEL2!50}
\colorlet{GRAD3}{GRADLEVEL3!50}
\colorlet{GRAD4}{GRADLEVEL4!50}
\colorlet{GRAD5}{GRADLEVEL5!50}
\colorlet{GRAD6}{GRADLEVEL6!50}
\colorlet{GRAD7}{GRADLEVEL7!50}
\colorlet{GRAD8}{GRADLEVEL8!50}
\colorlet{GRAD9}{GRADLEVEL9!50}
\colorlet{GRAD10}{GRADLEVEL10!50}
\definecolor{MONOGRADLEVEL1}{HTML}{e2fee2}
\definecolor{MONOGRADLEVEL2}{HTML}{cafdca}
\definecolor{MONOGRADLEVEL3}{HTML}{b2fcb2}
\definecolor{MONOGRADLEVEL4}{HTML}{9cfb9c}
\definecolor{MONOGRADLEVEL5}{HTML}{88ee91}
\definecolor{MONOGRADLEVEL6}{HTML}{74df89}
\definecolor{MONOGRADLEVEL7}{HTML}{61d081}
\definecolor{MONOGRADLEVEL8}{HTML}{4fc179}
\definecolor{MONOGRADLEVEL9}{HTML}{3cb371}
\definecolor{MONOGRADLEVEL10}{HTML}{3cb371}
\colorlet{MONOGRAD1}{MONOGRADLEVEL1!30}
\colorlet{MONOGRAD2}{MONOGRADLEVEL2!30}
\colorlet{MONOGRAD3}{MONOGRADLEVEL3!30}
\colorlet{MONOGRAD4}{MONOGRADLEVEL4!30}
\colorlet{MONOGRAD5}{MONOGRADLEVEL5!30}
\colorlet{MONOGRAD6}{MONOGRADLEVEL6!30}
\colorlet{MONOGRAD7}{MONOGRADLEVEL7!30}
\colorlet{MONOGRAD8}{MONOGRADLEVEL8!30}
\colorlet{MONOGRAD9}{MONOGRADLEVEL9!30}
\colorlet{MONOGRAD10}{MONOGRADLEVEL10!30}
\newcommand{\MONOCELGRADa}{\cellcolor{MONOGRAD1}}
\newcommand{\MONOCELGRADb}{\cellcolor{MONOGRAD2}}
\newcommand{\MONOCELGRADc}{\cellcolor{MONOGRAD3}}
\newcommand{\MONOCELGRADd}{\cellcolor{MONOGRAD4}}
\newcommand{\MONOCELGRADe}{\cellcolor{MONOGRAD5}}
\newcommand{\MONOCELGRADf}{\cellcolor{MONOGRAD6}}
\newcommand{\MONOCELGRADg}{\cellcolor{MONOGRAD7}}
\newcommand{\MONOCELGRADh}{\cellcolor{MONOGRAD8}}

\newcommand{\MONOCELGRADj}{\cellcolor{MONOGRAD10}}

\newcommand{\HIGHLIGHT}[2]{%
  \begingroup\setlength{\fboxsep}{1pt}%
  \colorbox{#1}{\hspace*{2pt}\vphantom{Ay}#2\hspace*{2pt}}%
  \endgroup
}

\newacronym{FID}{\textsc{fid}}{Fr\'echet Inception Distance}
\newacronym{VAE}{\textsc{vae}}{Variational Autoencoder}
\newacronym{GAN}{\textsc{gan}}{Generative Adversarial Network}
\newacronym{SDE}{\textsc{sde}}{Stochastic Differential Equation}
\newacronym{ELBO}{\textsc{elbo}}{evidence lower bound}
\newacronym{KL}{\textsc{kl}}{Kullback-Leibler}
\newacronym{MLP}{\textsc{mlp}}{multilayer perceptron}
\newacronym{NN}{\textsc{nn}}{neural network}
\newacronym{FDP}{\textsc{fdp}}{Functional Diffusion Process}
\newacronym{NFO}{\textsc{nfo}}{Neural Fourier Operator}
\newacronym{INR}{\textsc{inr}}{implicit neural representation}
\newacronym{LSGM}{\textsc{lsgm}}{Generative Modeling in Latent Space}
\newacronym{IDIFF}{\textsc{$\infty$-diff}}{Infinite Diffusion}
\newacronym{FD2F}{\textsc{fd2f}}{From Data To Functa}
\newacronym{SBD}{\textsc{sbd}}{Score Based Diffusion}
\newacronym{EDM}{\textsc{edm-g++}}{Generative Process with Discriminator Guidance}
\newacronym{NLF}{\textsc{nlf}}{Nonlinear Filtering}
\newacronym{T2I}{\textsc{t2i}}{Text-to-Image}
\newacronym{MI}{\textsc{mi}}{Mutual Information}
\newacronym{MITUNE}{\textsc{mi-tune}}{Mutual Information Fine Tuning}
\newacronym{HPSTUNE}{\textsc{hps-tune}}{Human Preference Score Fine Tuning}
\newacronym{MINDE}{\textsc{minde}}{Mutual Information Neural Diffusion Estimation}
\newacronym{SD}{\textsc{sd}}{Stable Diffusion}
\newacronym{SDXL}{\textsc{sdxl}}{Stable Diffusion XL}
\newacronym{BLIP}{\textsc{blip-vqa}}{BLIP}
\newacronym{UNET}{\textsc{unet}}{UNet}
\newacronym{CLIP}{\textsc{clip}}{CLIP}
\newacronym{VQA}{\textsc{vqa}}{Visual Question Answering}
\newacronym{DPO}{\textsc{dpo}}{Direct Preference Optimization}
\newacronym{LLM}{\textsc{llm}}{Large Language Model}
\newacronym{HPS}{\textsc{hps}}{Human Preference Score}
\newacronym{UNIDET}{\textsc{unidet}}{UniDet}

\newacronym{AE}{\textsc{a\&e}}{Attend and Excite}
\newacronym{SDG}{\textsc{sdg}}{Structured Diffusion Guidance}
\newacronym{SCG}{\textsc{scg}}{Semantic-aware Classifier-Free Guidance}
\newacronym{DPOK}{\textsc{dpok}}{Diffusion Policy Optimization with KL regularization}
\newacronym{GORS}{\textsc{gors}}{Generative mOdel finetuning with Reward-driven Sample selection}
\newacronym{HN}{\textsc{hn-itm}}{Hard-Negatives Image-Text-Matching}

\newacronym{DALLE}{\textsc{dalle-3}}{}
\newacronym{IMAGEN}{\textsc{imagen-3}}{}
\newacronym{DINO}{\textsc{fd-dino}}{}
\newacronym{CMMD}{\textsc{cmmd}}{MMD distance on CLIP-L features}
\newacronym{DIFFUSIONDB}{DiffusionDB}{}
\newacronym{SDBASE}{\textsc{\gls{SD}}-2.1-base}{}
\newacronym{HUMANTUNE}{\textsc{human-tune}}{}
\newacronym{MITUNESTAR}{\acrshort{MITUNE}$\star$}{}
\newacronym{T2ICOMPBENCH}{\gls{T2I}-CompBench}{}
\newacronym{SPRIGHT}{\textsc{SpRight}}{}
\newacronym{MITUNEXL}{\acrshort{MITUNE}-XL}{}
\newacronym{CFG}{\textsc{cfg}}{Classifier Free Guidance}

\newcommand{\mathbold}[1]{{\boldsymbol{{#1}}}}

\newcommand{\g}{\,|\,}

\newcommand{\nestedmathbold}[1]{{\mathbold{#1}}}

\newcommand{\mbp}{\nestedmathbold{p}}

\newcommand{\mbw}{\nestedmathbold{w}}
\newcommand{\mbx}{\nestedmathbold{x}}

\newcommand{\mbz}{\nestedmathbold{z}}

\newcommand{\mbI}{\nestedmathbold{I}}

\newcommand{\mbepsilon}{\nestedmathbold{\epsilon}}

\newcommand{\mbtheta}{\nestedmathbold{\theta}}

\DeclareRobustCommand{\KL}[2]{\ensuremath{\textsc{kl}\left[#1\;\|\;#2\right]}}
\DeclarePairedDelimiterX{\infdivx}[2]{[}{]}{%
#1\;\delimsize\|\;#2%
}

\newcommand{\cL}{\mathcal{L}}
\newcommand{\cN}{\mathcal{N}}

\newcommand{\mt}{m_t}
\newcommand{\ft}{\frac{\dd \log \mt}{\dd t}}

\usepackage{hyperref}
\usepackage{url}

\iclrfinalcopy

\title{Information Theoretic Text-to-Image Alignment}

\author{%
    $\bf \textrm{Chao Wang}^{1,2}$, 
    $\bf \textrm{Giulio Franzese}^{1}$, 
    $\bf \textrm{Alessandro Finamore}^{2}$, 
    $\bf \textrm{Massimo Gallo}^{2}$, 
    $\bf \textrm{Pietro Michiardi}^{1}$
    \\
    $\textrm{EURECOM}^{1}$, $\textrm{Huawei Technologies SASU, France}^{2}$
    \\
    \fontsize{8}{8}\selectfont
    \tt \textrm{$^1$\{}chao.wang, giulio.franzeze, pietro.michiardi\textrm{\}}@eurecom.fr
    \\
    \fontsize{8}{8}\selectfont
    \tt \textrm{$^2$\{}wang.chao3, alessandro.finamore, massimo.gallo\textrm{\}}@huawei.com
}

\begin{document}

\maketitle

\begin{abstract}
Diffusion models for \gls{T2I} conditional generation have recently achieved tremendous success. Yet, aligning these models with user's intentions still involves a laborious trial-and-error process, and this challenging alignment problem has attracted considerable attention from the research community.
In this work, instead of relying on fine-grained linguistic analyses of prompts, human annotation, or auxiliary vision-language models, we use \gls{MI} to guide model alignment.
In brief, our method uses self-supervised fine-tuning and relies on a point-wise \gls{MI} estimation between prompts and images to create a synthetic fine-tuning set for improving model alignment. 
Our analysis indicates that our method is superior to the state-of-the-art, yet it only requires the pre-trained denoising network of the \gls{T2I} model itself to estimate \acrshort{MI}, and a simple fine-tuning strategy that improves alignment while maintaining image quality. 
Code available at \href{https://github.com/Chao0511/mitune}{\textcolor{magenta}{\url{https://github.com/Chao0511/mitune}}}.

\end{abstract}

\section{Introduction}\label{sec:intro}
Generative models used for \acrfull{T2I} conditional generation~\citep{rombach2022, ramesh2022hierarchical, saharia2022, balaji2022, gafni2022, podell2024} have reached impressive performance. In particular, diffusion models~\citep{song2019, ho2020, kingma2021, song2020, song2021a, dhariwal2021} generate extremely high-quality images by specifying a natural text prompt that acts as a guiding signal~\citep{ho2022classifier, nichol2022glide, rombach2022}. %for the generative process. 
% Such synthetic images are of extremely high-quality, diversity and realism.
Yet, accurately translating prompts into images with the intended semantics is still complex~\citep{conwell2022, feng2023, wang2023}. Issues include catastrophic neglecting (i.e., prompt elements are not generated), incorrect attribute binding (i.e., elements attributes such as color, shape, and texture %describing elements 
are missing or wrongly assigned), incorrect spatial layout  %generated 
(i.e., elements are not correctly positioned), and a general difficulty in handling complex prompts~\citep{wu2024conceptmixcompositionalimagegeneration}.

%Such \textit{model alignment} problems have attracted a lot of attention recently. 
%MAX moved alignment evaluation metrics here
On the one hand, quantifying \textit{model alignment} is not trivial. Various works~\citep{hu2023tifa, gordon2023mismatch, grimal2024tiam} propose different metrics, most of which use complementary \gls{VQA} models or \glspl{LLM} to create scores measuring and explaining alignment. Moreover, a recent work \citep{huang2023t2icompbench} introduces a comprehensive benchmark suite to ease comparison among different metrics and modeling techniques via ``categories'', i.e., a pre-defined set of attribute binding, spatial-related, and other tasks. 

On the other hand, addressing \gls{T2I} model alignment is even more challenging than measuring it. Broadly, we can group the related literature into two main families: \emph{inference-time} and \emph{fine-tuning} methods.
For inference-time methods, the key intuition is that the generative process can be optimized by modifying the reverse path of the latent variables. Some works~\citep{chefer2023, li2023, rassin2023} mitigate failures by refining the cross-attention units \citep{tang2023} of the denoising network of \gls{SD}~\citep{rombach2022} on-the-fly, ensuring they attend to all subject tokens in the prompt (typically directly specified as a complementary prompt-specific input for the alignment process) and strengthen their activations. Other inference-time methods (\cite{agarwal2023, liu2022, kang2023counting, dahary2024, meral2024, feng2023Structured, kim2023, wu2023, zhang2024realcompo, zhang2024spdiffusion}), focus on individual failure cases. These approaches ($i$) require a linguistic analysis of prompts, leading to specialized solutions that rely on auxiliary models for prompt understanding, and ($ii$) result in considerably longer image generation time due to extra optimization costs during sampling.

Considering fine-tuning methods, 
% Model fine-tuning is another class of methods used for \gls{T2I} alignment. 
some works~\citep{wu2023better, lee2023aligning} require human annotations to prepare a fine-tuning set, while others \citep{fan2023dpok, wallace2023diffusion, clark2024directly} rely on Reinforcement Learning (RL), \gls{DPO}, or a differentiable reward function to steer model behavior. Recent methods use self-playing~\citep{yuan2024selfplay, xu2023imagereward, sun2023dreamsync, wang2023tokencompose, ma2023subjectdiffusionopen}, auxiliary models such as \gls{VQA}~\citep{li2023blip2,jiang2024CoMat} or segmentation maps~\citep{kirillov2023ICCV} in a semi-supervised fine-tuning setting. While these methods do not introduce extra inference time costs, they still require human annotation (which is subjective, costly, and does not scale well) and/or auxiliary models to guide the fine-tuning. %However, fine-tuning methods do not incur in additional costs at image generation time.

Complementary to both families are \emph{heuristic}-based methods that rely on a variety of ``tricks'', such as 
% More generally, the bulk of \gls{T2I} alignment research develops along two main lines: inference-time and fine-tuning methods.
% and a wide range of methods have been studied, including heuristics for 
prompt engineering~\citep{witteveen2022investigating, liu2022design, wang2023}, negative prompting~\citep{hfnegative, mahajan2023prompting, ogezi2024optimizing}, prompt rewriting~\citep{manas2024OPT2I} or brute force an appropriate seed selection~\citep{Samuel2023SeedSelect, karthik2023if}. While these methods can be beneficial in specific cases, they fundamentally shift the alignment problem to users.

Overall, current approaches require extra information (human input, auxiliary models, and additional data).
To the best of our knowledge, no previous work investigates \emph{self-supervised} approaches for T2I alignment, 
i.e., the use of a pre-trained model to generate images given a specific set of prompts, and select the most aligned ones to prepare a fine-tuning set, without using auxiliary models.
% \noteAF{@chao: this is not true, right?}
% In other words, since state of the art models are trained on huge volumes of prompt-image pairs, \emph{``is there any way one could tap into what already learned to foster task-specific alignment?''.}
In this work, we investigate this strategy  
% Our objective is to empirically study the benefits of a self-supervised fine-tuning approach. 
% In this paper, 
from an information theoretic perspective, by using \gls{MI} to quantify the non-linear prompt-image relationship. In particular, we focus on the estimation of \textit{point-wise} \gls{MI} using neural estimators~\citep{belghazi2018mine, song2019understanding, brekelmans2023improving, franzese2024minde, kong2024interpretable}, and study if and how \gls{MI} can be used as a meaningful signal to improve \gls{T2I} alignment, without relying on linguistic analysis of prompts, nor auxiliary models or heuristics. 
Our method unfolds as follows. We build upon the work in~\citep{franzese2024minde} and extend it to compute point-wise \gls{MI}.
% This adaptation enables the use of the \gls{T2I} model itself for estimating \gls{MI}. 
We then proceed with a self-supervised fine-tuning approach, whereby we use point-wise \gls{MI} to construct a fine-tuning set using synthetic data generated by the \gls{T2I} model itself. We then use the recent adapter presented in~\citep{liu2024dora} to fine-tune a small fraction of weights injected in the \gls{T2I} model denoising network. 
In summary, our work presents the following contributions:
\textbf{(1) We define a point-wise \gls{MI} estimator suitable for a discrete-time setting} (\Cref{sec:preliminaries}). We empirically study whether \gls{MI} between natural prompts and corresponding images considering both qualitative and quantitative approaches. Specifically, we show that \gls{MI} provides a meaningful indication of alignment with respect to both alignment metrics (\acrshort{BLIP} and \acrshort{HPS}) as well as a users study~(\Cref{sec:mialign}). 

%\gls{MI} is a simple and general measure of similarity between arbitrary distributions, and we leverage it to propose a new \gls{T2I} alignment method.

\textbf{(2) We design a self-supervised fine-tuning approach}, called \acrshort{MITUNE}~(\Cref{sec:finetuning}), that uses a small number of fine-tuning samples to align a pre-trained \gls{T2I} model without extra auxiliary models or inference overhead.

\textbf{(3) We perform an extensive experimental campaign} using a recent \gls{T2I} benchmark suite~\citep{huang2023t2icompbench}  and \acrshort{SDBASE} as base model obtaining sizable improvement compared to six alternative methods~(\Cref{sec:experiments}). Those benefits hold also when considering more complex tasks (based on DiffusionDB~\citep{diffusiondb}) and alternative base models (namely, \acrshort{SDXL}~\citep{podell2024}). Moreover we study the trade-off between T2I alignment and image quality that has been overlooked in the literature. Specifically, while the well-known \acrshort{FID}, \acrshort{DINO} and \acrshort{CMMD} metrics 
suggest a modest image quality/variety deterioration as a consequence of alignment objectives, optimizing the \acrfull{CFG} hyper-parameter of the fine-tuned model at generation time, enables finding a ``sweet spot'' between T2I alignment and image quality. 

% \noteMG{I'd move this to the beginning of the previous paragraph. It sounds highly repetitive.} Previous work requires an accurate analysis of which misalignment problem to address, either through linguistic analysis and prompt decomposition or through costly human-annotation campaigns. Instead, our approach relies on a theoretically sound measure, and it is lightweight, as it does not require large amounts of fine-tuning.
% In conclusion, \acrshort{MITUNE} is extremely simple to use, completely self-contained as it does not require external assistance, can be easily adapted to recent \gls{T2I} models, and offers a considerable advantage compared to the state of the art even with ``in the wild'' prompts that do not represented by available benchmarks.

\section{Preliminaries}\label{sec:preliminaries}
\noindent \textbf{Diffusion models. }
Denoising diffusion models \citep{ho2020, sohl-dickstein15} are generative models characterized by a forward process, that is fixed to a Markov chain that gradually adds Gaussian noise to the data according to a carefully selected variance schedule $\beta_t$, and a corresponding discrete-time reverse process, that has a Markov structure as well. Intuitively, diffusion models rely on the principle of iterative denoising: starting from a simple distribution $\mbx_T \sim \cN(\mathbf{0}, \mbI)$, samples are generated by iterative applications of a denoising network $\mbepsilon_\mbtheta$, that removes noise over $T$ denoising steps.
A simple way to learn the denoising network $\mbepsilon_\mbtheta$ is to consider a re-weighted variational lower bound of the marginal likelihood:

\vspace{-3pt}
{\small
\begin{equation}\label{eq:lsimple}
    \cL_{\text{simple}}(\mbtheta) = 
    \E_{t \sim U(0,T), \mbx_0 \sim p_{\text{data}}, \mbepsilon \sim \cN(\mathbf{0}, \mbI)}
    \left[
        ||
            \mbepsilon - \mbepsilon_\mbtheta (\sqrt{\bar{\alpha}_t} \mbx_0 + (\sqrt{1-\bar{\alpha}_t}) \mbepsilon, t)
        ||^2
    \right],
\end{equation}
}
where $\alpha_t = 1 - \beta_t$, $\bar{\alpha}_t = \prod_{s=1}^{t} \alpha_s$. For sampling, we let $\sigma_t^2 = \beta_t$. A similar variational objective can be obtained by switching perspective from discrete to continuous time \citep{song2021a}, whereby the denoising network approximates a score function of the data distribution. For image data, the denoising network is typically parameterized by a \acrshort{UNET} \citep{ronneberger2015, rombach2022}.

This simple formulation has been extended to conditional generation \citep{ho2021classifierfree}, whereby a conditioning signal $\mbp$ injects ``external information'' in the iterative denoising process. This requires a simple extension to the denoising network such that it can accept the conditioning signal: $\mbepsilon_\mbtheta(\mbx_t, \mbp, t)$. Then, during training, a randomized approach allows to learn both the conditional and unconditional variants of the denoising network, for example by assigning a null value to the conditioning signal. At sampling time, a weighted linear combination of the conditional and unconditional networks, such as $\tilde{\mbepsilon}_\mbtheta(\mbx_t, \mbp, t) = \mbepsilon_\mbtheta(\mbx_t, \emptyset, t) + \gamma (\mbepsilon_\mbtheta(\mbx_t, \mbp, t) - \mbepsilon_\mbtheta(\mbx_t, \emptyset, t))$ can be used. 

In this work, we use pre-trained latent diffusion models operating on a learned projection of the input data $\mbx_0$ into a corresponding latent variable $\mbz_0$ which is lower-dimensional compared to the original data. Moreover, the conditioning signal $\mbp$ is obtained by a text encoder such as \acrshort{CLIP} \citep{radford2021}. 

\noindent \textbf{MI estimation. }
\gls{MI} is a central measure to study the non-linear dependence between random variables~\citep{shannon1948mathematical, mackay2003information}, and has been extensively used in machine learning for representation learning~\citep{bell1995information, stratos2018mutual, belghazi2018mine, oord2019representation, hjelm2019learning}, and for both training~\citep{alemi2019deep, chen2016infogan, zhao2018information} and evaluating generative models~\citep{alemi2019gilbo, huang2020evaluating}. 

For many problems of interest, precise computation of \gls{MI} is not trivial~\citep{mcallester2020formal, paninski2003estimation}. Consequently, a wide range of techniques for \gls{MI} estimation have flourished. In this work, we focus on realistic and high-dimensional data, which calls for recent advances in \gls{MI} estimation~\citep{papamakarios2017masked, belghazi2018mine, oord2019representation, song2019understanding, rhodes2020telescoping, letizia2022copula, brekelmans2023improving, kong2024interpretable}. 
In particular, we capitalize on a recent method~\citep{franzese2024minde}, that relies on the theory behind continuous-time diffusion processes \citep{song2021a} and uses the Girsanov Theorem \citep{oksendal2003stochastic} to show that score functions can be used to compute the \gls{KL} divergence between two distributions. 
In what follows, we use a simplified notation and gloss over several mathematical details to favor intuition over rigor. 
Here we consider discrete-time diffusion models, which are equivalent to the continuous-time counterpart under the variational formulation, up to constants and discretization errors \citep{song2021a}. 

We begin by considering the two arbitrary random variables $\mbz$ and $\mbp$ which are sampled from the joint distribution $p_{\text{latent},\text{prompt}}$, where the former corresponds to the distribution of the projections in a latent space of the image distribution, and the latter to the distribution of prompts used for conditional generation. 
Then, following the approach in~\citep{franzese2024minde}, with the necessary adaptation to the discrete domain (see \Cref{app:minde} for details), point-wise \gls{MI} estimation can be obtained as follows: 

{\small
\begin{equation}\label{eq:mi}
    \textsc{I}(\mbz,\mbp) = 
    \E_{t,\mbepsilon\sim \cN(\mathbf{0}, \mbI)}
    \left[
    \kappa_t
    ||
        \mbepsilon_\mbtheta(\mbz_t, \mbp, t) - \mbepsilon_\mbtheta(\mbz_t, \emptyset, t)
    ||^2
    \right],\,\,\, \kappa_t = \frac{\beta_t T}{2\alpha_t (1- \bar{\alpha}_t)}.
\end{equation}
}

Given a pre-trained diffusion model, we compute an expectation (over diffusion times $t$) of the scaled squared norm of the difference between the conditional $\mbepsilon_\mbtheta(\mbz_t, \mbp, t)$ and unconditional networks $\mbepsilon_\mbtheta(\mbz_t, \emptyset, t)$, which corresponds to an estimate of the point-wise \gls{MI} between an image and a prompt. 
Intuitively, the difference between these scores quantifies how much extra knowledge of the prompt helps in denoising the perturbed images. This is both a key ingredient and a competitive advantage of our method, as it enables a self-contained approach 
to alignment based on the \gls{T2I} model alone without auxiliary models or human~feedback.

\section{Our method: \acrshort{MITUNE}}\label{sec:method}
The \gls{T2I} alignment problem arises when user's intentions, as expressed through natural text prompts, fail to materialize in the generated image. 
Our novel approach aims to address alignment using a theoretically grounded \gls{MI} estimation, that applies across various contexts. To improve model alignment, we introduce a self-supervised fine-tuning method. Leveraging the T2I model itself, we estimate \gls{MI} and generate an information-theoretic enhanced fine-tuning dataset. While our focus in this work is on \gls{T2I} alignment, our framework remains extensible to other modalities.

\begin{figure}[!t]
\centering
\hspace{0.01\textwidth}
\minipage{0.18\textwidth}
{\small \textbf{Color binding}: \\
``\textit{A blue car and \\
a red horse}''}
\vspace{+1.5cm}
\endminipage
\minipage{0.14\textwidth}
\includegraphics[width=\linewidth]{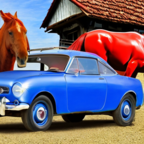}
\vspace{-0.65cm}
\caption*{
    \hspace{-0.43em}
    \fontsize{6}{6}\selectfont
    \begin{tabular}{@{}r@{$\,=\,$}l@{}}
    \acrshort{BLIP} & 0.93 \\
    \acrshort{HPS} & 0.319 \\
    \gls{MI} & 36.28\\
    \end{tabular}
}
\endminipage
\hspace{0.01\textwidth}
\minipage{0.14\textwidth}
\includegraphics[width=\linewidth]{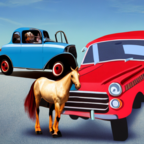}
\vspace{-0.65cm}
\caption*{
    \hspace{-0.43em}
    \fontsize{6}{6}\selectfont
    \begin{tabular}{@{}r@{$\,=\,$}l@{}}
    \acrshort{BLIP} & 0.40 \\
    \acrshort{HPS}  & 0.310 \\
    \gls{MI}        & 24.56 \\
    \end{tabular}
}
\endminipage
\hspace{0.01\textwidth}
\minipage{0.14\textwidth}
\includegraphics[width=\linewidth]{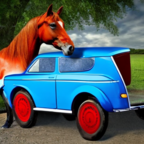}
\vspace{-0.65cm}
\caption*{
    \hspace{-0.43em}
    \fontsize{6}{6}\selectfont
    \begin{tabular}{@{}r@{$\,=\,$}l@{}}
    \acrshort{BLIP} & 0.17 \\
    \acrshort{HPS}  & 0.312 \\
    \gls{MI}        & 22.05 \\
    \end{tabular}
}
\endminipage
\hspace{0.01\textwidth}
\minipage{0.14\textwidth}
\includegraphics[width=\linewidth]{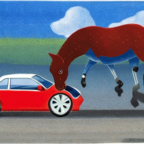}
\vspace{-0.65cm}
\caption*{
    \hspace{-0.43em}
    \fontsize{6}{6}\selectfont
    \begin{tabular}{@{}r@{$\,=\,$}l@{}}
    \acrshort{BLIP} & 0.06 \\
    \acrshort{HPS}  & 0.263 \\ 
    \gls{MI}        & 15.44 \\
    \end{tabular}
}
\endminipage
\hspace{0.01\textwidth}
\minipage{0.14\textwidth}
\includegraphics[width=\linewidth]{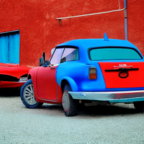}
\vspace{-0.65cm}
\caption*{
    \hspace{-0.43em}
    \fontsize{6}{6}\selectfont
    \begin{tabular}{@{}r@{$\,=\,$}l@{}}
    \acrshort{BLIP} & 0.05 \\
    \acrshort{HPS}  & 0.258 \\
    \gls{MI}        & 14.67\\
    \end{tabular}
}
\endminipage

\hspace{0.005\textwidth}
\minipage{0.18\textwidth}
{\small \textbf{Texture binding}: \\
``\textit{A fabric dress and \\
a glass table}''}
\vspace{+1.5cm}
\endminipage
\minipage{0.14\textwidth}
\includegraphics[width=\linewidth]{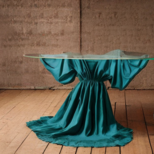}
\vspace{-0.65cm}
\caption*{
    \hspace{-0.43em}
    \fontsize{6}{6}\selectfont
    \begin{tabular}{@{}r@{$\,=\,$}l@{}}
    \acrshort{BLIP} & 0.90 \\
    \acrshort{HPS}  & 0.257 \\
    \gls{MI}        & 44.6\\
    \end{tabular}
}
\endminipage
\hspace{0.01\textwidth}
\minipage{0.14\textwidth}
\includegraphics[width=\linewidth]{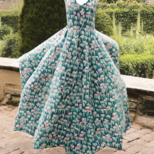}
\vspace{-0.65cm}
\caption*{
    \hspace{-0.43em}
    \fontsize{6}{6}\selectfont
    \begin{tabular}{@{}r@{$\,=\,$}l@{}}
    \acrshort{BLIP} & 0.46 \\
    \acrshort{HPS}  & 0.213 \\
    \gls{MI}        & 28.1 \\
    \end{tabular}
}
\endminipage
\hspace{0.01\textwidth}
\minipage{0.14\textwidth}
\includegraphics[width=\linewidth]{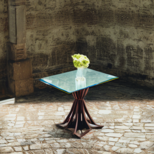}
\vspace{-0.65cm}
\caption*{
    \hspace{-0.43em}
    \fontsize{6}{6}\selectfont
    \begin{tabular}{@{}r@{$\,=\,$}l@{}}
    \acrshort{BLIP} & 0.17 \\
    \acrshort{HPS}  & 0.201 \\
    \gls{MI}        & 19.86\\
    \end{tabular}
}
\endminipage
\hspace{0.01\textwidth}
\minipage{0.14\textwidth}
\includegraphics[width=\linewidth]{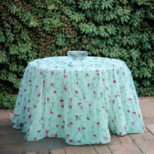}
\vspace{-0.65cm}
\caption*{
    \hspace{-0.43em}
    \fontsize{6}{6}\selectfont
    \begin{tabular}{@{}r@{$\,=\,$}l@{}}
    \acrshort{BLIP} & 0.12 \\
    \acrshort{HPS}  & 0.231 \\
    \gls{MI}        & 15.41\\
    \end{tabular}
}
\endminipage
\hspace{0.01\textwidth}
\minipage{0.14\textwidth}
\includegraphics[width=\linewidth]{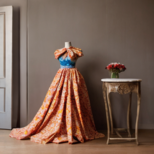}
\vspace{-0.65cm}
\caption*{
    \hspace{-0.43em}
    \fontsize{6}{6}\selectfont
    \begin{tabular}{@{}r@{$\,=\,$}l@{}}
    \acrshort{BLIP} & 0.07 \\
    \acrshort{HPS}  & 0.295 \\
    \gls{MI}        & 9.34
    \end{tabular}
}
\endminipage

\hspace{0.005\textwidth}
\minipage{0.18\textwidth}
{\small \textbf{Shape binding}:\\
``\textit{A round bag and \\
a rectangular \\
wallet}''}
\vspace{+1.5cm}
\endminipage
\minipage{0.14\textwidth}
\includegraphics[width=\linewidth]{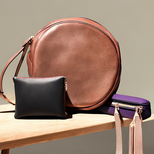}
\vspace{-0.65cm}
\caption*{
    \hspace{-0.43em}
    \fontsize{6}{6}\selectfont
    \begin{tabular}{@{}r@{$\,=\,$}l@{}}
    \acrshort{BLIP} & 0.82 \\
    \acrshort{HPS}  & 0.262 \\
    \gls{MI}        & 18.61\\
    \end{tabular}
}
\endminipage
\hspace{0.01\textwidth}
\minipage{0.14\textwidth}
\includegraphics[width=\linewidth]{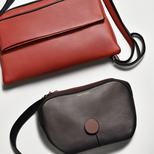}
\vspace{-0.65cm}
\caption*{
    \hspace{-0.43em}
    \fontsize{6}{6}\selectfont
    \begin{tabular}{@{}r@{$\,=\,$}l@{}}
    \acrshort{BLIP} & 0.64 \\
    \acrshort{HPS}  & 0.247 \\
    \gls{MI}        & 17.16\\
    \end{tabular}
}
\endminipage
\hspace{0.01\textwidth}
\minipage{0.14\textwidth}
\includegraphics[width=\linewidth]{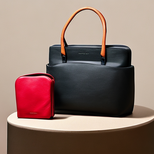}
\vspace{-0.65cm}
\caption*{
    \hspace{-0.43em}
    \fontsize{6}{6}\selectfont
    \begin{tabular}{@{}r@{$\,=\,$}l@{}}
    \acrshort{BLIP} & 0.27 \\
    \acrshort{HPS}  & 0.262 \\
    \gls{MI}        & 14.84\\
    \end{tabular}
}
\endminipage
\hspace{0.01\textwidth}
\minipage{0.14\textwidth}
\includegraphics[width=\linewidth]{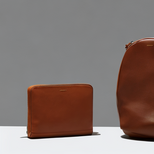}
\vspace{-0.65cm}
\caption*{
    \hspace{-0.43em}
    \fontsize{6}{6}\selectfont
    \begin{tabular}{@{}r@{$\,=\,$}l@{}}
    \acrshort{BLIP} & 0.24 \\
    \acrshort{HPS}  & 0.216 \\
    \gls{MI}        & 12.50
    \end{tabular}
}
\endminipage
\hspace{0.005\textwidth}
\minipage{0.14\textwidth}
\includegraphics[width=\linewidth]{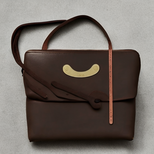}
\vspace{-0.65cm}
\caption*{
    \hspace{-0.43em}
    \fontsize{6}{6}\selectfont
    \begin{tabular}{@{}r@{$\,=\,$}l@{}}
    \acrshort{BLIP} & 0.01 \\
    \acrshort{HPS}  & 0.160 \\
    \gls{MI}        & 11.57    
    \end{tabular}
}
\endminipage 
\caption{Qualitative analysis of \gls{MI} as an alignment measure (all metrics decrease from left to right). See also \Cref{app:MI-qualitative-all-categories}.
}
\label{fig:MI-qualitative}
\end{figure}

\subsection{Is Mutual Information meaningful for alignment?}\label{sec:mialign}
To the best of our knowledge, \gls{MI} has never been evaluated as a \emph{meaningful signal} for \gls{T2I} alignment. 
As such, in this section we perform both qualitative and quantitative analyses to investigate this aspect.

\noindent\textbf{Qualitative analysis.}
Starting with a qualitative analysis,
we select a set of simple prompts to probe color, texture, and shape attribute binding from \acrshort{T2ICOMPBENCH}~\citep{huang2023t2icompbench} using \gls{SD}~\citep{rombach2022} (specifically \acrshort{SDBASE}) to generate the corresponding images.
We then measure the well-known \acrshort{BLIP}~\citep{huang2023t2icompbench} and \gls{HPS}~\citep{wu2023human} alignment metrics as well as point-wise \gls{MI} estimates. \acrshort{BLIP} uses a large vision-language model to compute an alignment score, by casting questions against an image to verify that the prompt used to generate it is well represented. \gls{HPS} is an elaborate metric that uses an auxiliary pre-trained model, blending alignment with aesthetics according to human perception, which are factors that can sometimes be in conflict. 
\Cref{fig:MI-qualitative} collects some examples and related metric scores revealing a substantial agreement among all measures: all metrics decrease from left to right in the figure, as prompt-image alignment deteriorates.

\noindent\textbf{Quantitative analysis.}
To quantitatively measure the agreement between \gls{MI} and well-established alignment metrics, we use all 700 prompts from \acrshort{T2ICOMPBENCH} and use \gls{SD} (again, \acrshort{SDBASE}) to generate 50 images per prompt. We use point-wise \gls{MI} to rank such images and select the 1st, 25th, and 50th. For these three representative images, we compute \acrshort{BLIP} and \gls{HPS} scores and re-rank them according to both metrics. 
Last, we measure agreement between the three rankings using Kendall's $\tau$ method~\citep{kendall38}, and average results across all prompts. Results indicate good agreement between \gls{MI} and \acrshort{BLIP} ($\tau=0.4$), and a strong agreement between \gls{MI} and \gls{HPS} ($\tau=0.68$).

To strengthen our analysis, we also perform a users study eliciting human preference (see \Cref{app:survey-metrics} for details). Given a randomly selected prompt from \acrshort{T2ICOMPBENCH} that users can read, we present the top-ranked generated image (among the 50) according to \gls{MI}, \acrshort{BLIP} and \gls{HPS}, in a randomized order. Users can select one or more images to indicate their preference regarding alignment and aesthetics, for a total of 10 random prompts per user. From the 102 surveys from 46 users, we find that human preference for prompt-image pairs goes to \gls{MI} for 69.1\%, \acrshort{BLIP} for 73.5\% and \gls{HPS} for 52.2\% of the cases, respectively.

\noindent\textbf{Relevant literature.}
Overall, our analyses support our intuition by which \emph{\gls{MI} is a meaningful signal for alignment} (and possibly aesthetics too), setting the stage for our \gls{T2I} alignment method. 
Our intuition is also supported by recent studies investigating the information flow in the generative process of diffusion models. Specifically, \cite{kong2024interpretable} estimates pixel-wise mutual information between natural prompts and the images generated at each time-step of a backward diffusion process. They compare such ``information maps'' to cross-attention maps~\citep{tang2023} in an experiment involving prompt manipulation -- modifications of the initial prompt during reverse diffusion --  and conclude that \gls{MI} is much more sensitive to information flow from prompt to images. In a similar vein, \cite{franzese2024minde} compute \gls{MI} between prompt and images at different stages of the reverse process of image generation. Experimental evidence indicates that \gls{MI} can be used to analyze various reverse diffusion phases: noise, semantic, and denoising stages~\citep{balaji2022ediff}.
While previous studies do not explicitly focus on alignment, they \emph{indirectly} support our intuition that \gls{MI} estimated using a diffusion model   
gauges the amount of information a text prompt conveys about an image (and vice-versa) which is key for T2I alignment.

\subsection{Self-supervised fine-tuning with \acrshort{MITUNE}}\label{sec:finetuning}

In summary, given a pre-trained diffusion model such as \gls{SD}~\citep{rombach2022} or any variant, such as \acrshort{SDXL}~\citep{podell2024}, we leverage our point-wise \gls{MI} estimation method to select a small fine-tuning dataset set of information-theoretic aligned examples. 

Our self-supervised alignment method \textbf{relies on the pre-trained model only} to produce a given amount of fine-tuning data, which is then filtered to retain prompt-image pairs with a high degree of alignment, according to pair-wise \gls{MI} \textbf{estimates obtained using only the pre-trained model}. We begin with a set of fine-tuning prompts $\mathcal{P}$, which can be either manually crafted, or borrowed from available prompt collections~\citep{wang2023, huang2023t2icompbench}. Ideally, fine-tuning prompts should be conceived to stress the pre-trained model with challenging attribute and spatial bindings, or complex rendering tasks.

\vspace{-1em}
\IncMargin{0.8em} 

\begin{minipage}[t]{0.5\textwidth}
\fontsize{7.2}{7.2}\selectfont
\begin{algorithm}[H]
    \SetKwFunction{main}{PointWise\gls{MI}}
    \SetKwFunction{tk}{Top-$k$}
    \SetKwFunction{ft}{FineTune}
    \SetKwFunction{app}{append}
    \SetKwProg{Fn}{Function}{:}{}
    \SetCommentSty{mycommstyle}
    \DontPrintSemicolon
    \SetAlgoLined
    \SetKwInOut{Input}{Input}
    \SetKwInOut{Hyperparameters}{Hyper par}
    \SetKwInOut{Output}{Output}
    \Input{Pre-trained model: $\mbepsilon_\theta$, Prompt set: $\mathcal{P}$}
    \Hyperparameters{Image pool size: $M$; Top \gls{MI}-aligned samples: $k$}
    \Output{Fine-tuned diffusion model $\mbepsilon_{\theta^*}$}
    \BlankLine

    \tcp{Fine-tuning set}
    $\mathcal{S} \leftarrow [\,\,]$   
    
    \For{$\mbp^{(i)}$ in $\mathcal{P}$}{
        \For{$j \in \{1,\cdots,M\}$} { 
            \tcp{Generate and compute \gls{MI}}
            $\mbz^{(j)}$, $\textsc{I}(\mbz^{(j)},\mbp^{(i)})$ = \main{$\mbepsilon_\theta$, $\mbp^{(i)}$}
            
            \tcp{Append samples and \gls{MI}}
            $\mathcal{S}[\mbp^{(i)}]$.\app{$\mbz^{(j)}, \textsc{I}(\mbz^{(j)},\mbp^{(i)})$}  
        }
        \tcp{Retain only Top-$k$ elements}
        $\mathcal{S}[\mbp^{(i)}]$ = \tk{$\mathcal{S}[\mbp^{(i)}]$} 
    }
    \Return $\mbepsilon_{\theta^*}$ = \ft{$\mbepsilon_\theta$, $\mathcal{S}$}
\caption{
\fontsize{8.2}{8.2}\selectfont
\textsc{mi-tune} 
\label{algo:mitune}
}
\end{algorithm}
\end{minipage}
%%%%%%%%%%%%%%%%%%%%%%%%%%%%
\hspace{1em}
%%%%%%%%%%%%%%%%%%%%%%%%%%%%
\begin{minipage}[t]{0.45\textwidth}
\fontsize{7.5}{7.5}\selectfont
\begin{algorithm}[H]
    \SetKwFunction{main}{PointWise\gls{MI}}
    \SetKwProg{Fn}{Function}{:}{}
    \SetKwInOut{Input}{Input}
    \SetKwInOut{Output}{Output}
    \DontPrintSemicolon
    \SetAlgoLined
    \SetCommentSty{mycommstyle}
    \Input{Pre-trained model: $\mbepsilon_\theta$; Prompt: $\mbp$}
    \Output{Generated latent: $\mbz$; Point-wise \gls{MI}: $\textsc{I}(\mbz,\mbp)$}
    \BlankLine
      
    \Fn{\main{$\mbepsilon_\theta$, $\mbp$}}{
        \tcp{Initial latent sample}
        $\mbz_T \sim \mathcal{N}(\mathbf{0}, \mbI)$ 
        \For{$t$ in $T,...,0$}{ 
            \tcp{\gls{MI} estimation (\cref{eq:mi})}
            $\textsc{I}(\mbz_t,\mbp) \mathrel{+}= \left[ \kappa_t || \mbepsilon_\mbtheta(\mbz_t, \mbp, t) - \mbepsilon_\mbtheta(\mbz_t, \emptyset, t) ||^2\right]$ 

            \tcp{Noise sample}
            $\mbw \sim \mathcal{N}(\mathbf{0}, \mbI)$ if $t>1$, else $\mbw = \mathbf{0}$
            
            \tcp{Sampling step}% (\cref{eq:guidance})}
            $\mbz_{t-1} =
                \frac{1}{\sqrt{\alpha_t}}
                \left( \mbz_t - \frac{1-\alpha_t}{\sqrt{1-\bar{\alpha}_t}} {\mbepsilon}_\mbtheta(\mbz_t, \mbp, t) \right) + \sigma_t \mbw
                $ 
        }
    \Return $\mbz$, $\textsc{I}(\mbz,\mbp)$
    }    
    \caption{
    \fontsize{8.5}{8.5}\selectfont
    Point-wise \gls{MI} Estimation
    \label{algo:miestimate}
    }
\end{algorithm}
\end{minipage}

As described in Algorithm \ref{algo:mitune}, for each prompt $p^{(i)}$ in the fine-tuning set $\mathcal{P}$, we use the pre-trained model to generate a fixed number $M$ of synthetic 
images. Given prompt-image pairs $(p^{(i)} ; z^{(j)}),\, j \in [1,M]$, we estimate pair-wise \gls{MI} and select the top $k$ pairs, which will be part of the model fine-tuning dataset $\mathcal{S}$. 
Finally, we augment the pre-trained model with adapters~\citep{hu2021lora, liu2024dora}, and proceed with fine-tuning. 
We study the impact of the adapter choice, and whether only the denoising network or both the denoising and text encoder networks should be fine-tuned (\Cref{app:abl_model}). Moreover, we measure the impact of the number of fine-tuning rounds $R$ to the pre-trained model, i.e., we renew the fine-tuning dataset $\mathcal{S}$ using the fine-tuned model, and re-fine-tune it using Algorithm~\ref{algo:mitune} (\Cref{sec:results}). Our efficient implementation combines latent generation and point-wise \gls{MI} computation as shown in Algorithm~\ref{algo:miestimate}. Since \gls{MI} estimation involves computing an expectation over diffusion times $t$, it is easy to combine generation and estimation in the same loop. Moreover, the function is easy to parallelize to significantly speed up the fine-tuning set $\mathcal{S}$ composition.

\vspace{-0.9em}
\section{Experimental evaluation}\label{sec:experiments}
\subsection{Benchmark and metrics
\label{sec:benchmark-and-metrics}
}

\noindent \textbf{Benchmark.}
We compare all techniques using \acrshort{T2ICOMPBENCH}~\citep{huang2023t2icompbench}, a benchmark composed of 700/300 (train/test) prompts across 6 \emph{categories} including attribute binding (color, shape, and texture categories), object relationships (2D-spatial and non-spatial associations), and complex composition tasks. These prompts were generated with predefined rules or ChatGPT~\citep{chatgpt}. We also assess \acrshort{MITUNE} performance on more realistic prompts by sampling 5,000/1,250 (train/test) prompt-image pairs from \acrshort{DIFFUSIONDB}~\citep{diffusiondb}, a large-scale dataset composed of complex human-crafted prompts paired with the corresponding images generated from a \acrshort{SD} model.

\noindent\textbf{Alignment Metrics.}
Evaluating %compositional properties of 
\gls{T2I} alignment is difficult as it requires a detailed understanding of prompt-images pairs, and many metrics have been proposed, e.g., \acrshort{CLIP}~\citep{hessel2021-clipscore, radford2021}, \textsc{minigpt-4}~\citep{zhu2024minigpt}, and human evaluation. In our work we use \acrshort{BLIP}~\citep{huang2023t2icompbench}, \gls{HPS}~\citep{wu2023HPS} and UniDet~\citep{zhou2021simple}. 
While \acrshort{BLIP} computes a score with a questions-answers approach -- a given prompt is decomposed and each part is transformed into a question for an auxiliary VQA model; then, answers are aggregated into a single score -- only based on alignment, \acrshort{HPS} includes both alignment and aesthetics -- this is enabled by an auxiliary model pre-trained using human-annotated data.
As in \citep{huang2023t2icompbench}, the 2D-spatial category is evaluated using the UniDet object detection model.

We complement these metrics with a user study. 
We randomly select 100 prompts per category, and generate 10 pictures per prompt for each method we consider in our evaluation. Then, we run surveys composed of 12 rounds (2 for each category), each showing to the user a randomly selected prompt and a randomly selected image for each method, randomly arranged in a grid. At each round, users need to select zero or more images they consider aligned with the prompt. Overall, we collected 42 surveys from 5 users, from which we computed the total percentage of times each method was selected for each category (\Cref{app:survey-methods}).

\noindent\textbf{Image quality metrics.}
Assessing performance only considering alignment metrics can hide undesired effects.
Intuitively, a strong adherence to a given prompt reduces the generative process ``degrees of freedom'' and this trade-off might not be visible even to a trained eye. To investigate these dynamics we compute \acrshort{FID}~\citep{fid}, \acrshort{DINO}~\citep{dino} and \acrshort{CMMD}~\citep{cmmd} scores -- \acrshort{FID} favors natural colors and textures but struggles to detect objects/shapes distortion, while \acrshort{DINO} and \acrshort{CMMD} favor image content. 
Following \citep{imagenteamgoogle2024imagen3}, rather than using the \acrshort{T2ICOMPBENCH} test set, we compute the metrics using 30k samples of the \textsc{ms-coco-2014}~\citep{lin2015microsoftcococommonobjects} validation set.

\vspace{-5pt}
\subsection{\acrshort{MITUNE} Fine-tuning
\label{sec:mitune-finetuning}
}

\noindent \textbf{Base models.} 
We mainly run our benchmark using \acrshort{SDBASE} as base model, but 
we also report results of the application of \acrshort{MITUNE} on \acrshort{SDXL} to demonstrate its flexibility. 

\noindent \textbf{Fine-tuning sets.} 
\acrshort{T2ICOMPBENCH} contains 700 training prompts for each category.
When using \acrshort{MITUNE}, we generate $M\!=\!50$ images for each prompt using the pre-trained model, %$\mbp^{(i)}$, 
compute their point-wise \gls{MI}, and select the top $k\!=\!1$\footnote{We remark that, albeit in a different context, this selection resembles an image retrieval task~\citep{krojer2023HNitm}} 
(sensitivity to $M$ and $k$ in \Cref{app:abl_ft_set_strategies}).
For the 2D-Spatial category, we also compose fine-tuning sets generating images from  
\acrshort{SPRIGHT}~\citep{chatterjee2024SpRight} -- a model optimized for this (more challenging) category and fine-tuned from \gls{SD}-2.1 (a higher resolution version of \acrshort{SDBASE}). Last, we also contrast \acrshort{MITUNE} fine-tuning set composition against ($i$) using \acrshort{HPS} rather than MI for image selection,\footnote{We exclude \acrshort{BLIP} for the fine-tuning set composition to avoid  biasing the evaluation~\citep{huang2023t2icompbench}.} ($ii$) using both MI-selected and real-pictures and ($iii$) images from \acrshort{DIFFUSIONDB}.

\noindent \textbf{Fine-tuning weights.}
In our work, fine-tuning corresponds to injecting DoRA~\citep{liu2024dora} adapters (rank and scaling factor $\alpha$ are set to 32) only into the attention layers and fully connected layers of the denoising \acrshort{UNET} network, whereas other layers 
are frozen.\footnote{
LoRA adapters~\citep{hu2021lora} and fine-tuning also the \acrshort{CLIP}-based text encoder do not provide performance improvements (\Cref{app:abl_model}). Likewise, creating a multi-category model by ``merging'' different per-category models 
or using a fine-tuning set composed with images from all categories do not provide performance gains
(\Cref{app:abl_Merge}).
}

\noindent \textbf{Other hyperparams search.} 
We consider up to $R\!\in\![1,3]$ rounds of fine-tuning i.e., using as base model the one obtained from previous round and apply Algorithm  \ref{algo:mitune}, and \acrfull{CFG} $\in [2.5, 7.5]$. For each fine-tuned model we then compute all alignment and image quality metrics. More fine-grained hyperparams details and computational costs considerations in \Cref{app:exp_protocol}.

\setlength{\tabcolsep}{0pt}
\begin{table}[ht]
\centering
\caption{
Alignment results (\%). 
\HIGHLIGHT{gray!20}{Gray highlighted} style when \acrshort{MITUNE} outperforms all competitors;
\GRAY{Grayed text} for under-performing methods per-family; 
\HIGHLIGHT{MONOGRAD10}{Green heatmaps} show per-category absolute gains w.r.t. the base model.
}
\label{tab:results_blip}
\vspace{-5pt}
\scalebox{0.62}{
\begin{tabular}{
    @{} 
    c
    @{}
    r
    %%%%%%%%% blip-vqa
    @{$\:\:\:\:$}
    r
    @{$\:\:$}
    r
    @{$\:\:$}
    r
    @{$\:\:$}
    r
    @{$\:\:$}
    r
    @{$\:\:$}
    r
    @{$\:\:$}
    r
    %%%%%%%%% hps
    @{$\:\:\:$}
    r
    @{$\:\:$}
    r
    @{$\:\:$}
    r
    @{$\:\:$}
    r
    @{$\:\:$}
    r
    @{$\:\:$}
    r
    @{$\:\:$}
    r
    %%%%%%%%% human
    @{$\:\:\:$}
    r
    @{$\:\:$}
    r
    @{$\:\:$}
    r
    @{$\:\:$}
    r
    @{$\:\:$}
    r
    @{$\:\:$}
    r
    @{$\:\:$}
    r
    %%%%
    @{}
}
%%%%%%%%%%%%%%%%%%%%%%%%%%%
\toprule
&  
& \multicolumn{7}{c}{\bf \acrshort{BLIP}} 
& \multicolumn{7}{c}{\bf \acrshort{HPS}}
& \multicolumn{7}{c}{\textbf{Human} (\textit{user study})}
\\
\cmidrule(r){1-2}
\cmidrule(r){3-9}
\cmidrule(r){10-16}
\cmidrule{17-23}
& 
\bf Method &
    \TINY{Color} & \TINY{Shape} & \TINY{Texture} & \TINY{2D-Sp.} & \TINY{Non-Sp.} & \TINY{Compl.} & \TINY{(\textit{avg})} &
    \TINY{Color} & \TINY{Shape} & \TINY{Texture} & \TINY{2D-Sp.} & \TINY{Non-Sp.} & \TINY{Compl.} & \TINY{(\textit{avg})} &
    \TINY{Color} & \TINY{Shape} & \TINY{Texture} & \TINY{2D-Sp.} & \TINY{Non-Sp.} & \TINY{Compl.} & \TINY{(\textit{avg})}
\\
\cmidrule(r){1-2}
\cmidrule(r){3-8}
\cmidrule(r){9-9}
\cmidrule(r){10-16}
\cmidrule{17-23}
    & 
    \TINY{\acrshort{SDBASE}}
    %color      %shape   %texture   %2d-spatial %non-spatial    %complex    %avg
    &49.65      &42.71   &49.99     &15.77      &66.23          &50.53      &(\it45.81)
    &27.64      &24.56   &24.99     &27.50      &26.66          &25.70      &(\it26.17)
    &29.76      &11.90   &40.48     &35.71      &66.67          &29.76      &(\it35.71)
\\
\cmidrule(r){1-2}
\cmidrule(r){3-9}
\cmidrule(r){10-16}
\cmidrule{17-23}
\multirow{3}{*}{\rotatebox[origin=c]{90}{Infer.}} & 
    \acrshort{AE}
        %color              %shape          %texture            %2d-spatial         %non-spatial        %complex        %avg
        &\UL{61.43}         &\UL{47.39}     &\UL{64.10}         &\GRAY{16.18}       &\GRAY{66.21}       &\UL{51.69}     &(\it51.17)
        &\UL{28.44}         &\GRAY{24.43}   &\UL{25.88}         &\UL{28.42}         &\GRAY{26.60}       &\GRAY{25.60}   &(\it26.56)
        &\UL{31.95}         &\UL{15.48}     &\UL{52.38}         &\GRAY{32.14}       &\GRAY{65.48}       &\GRAY{30.95}   &(\it38.06)
\\
    & \acrshort{SDG}
        %color              %shape          %texture            %2d-spatial         %non-spatial        %complex        %avg
        &\GRAY{47.15}       &\GRAY{45.24}   &\GRAY{47.13}       &\GRAY{15.25}       &\GRAY{66.17}       &\GRAY{47.41}   &\GRAY{(\it 44.72)}
        &\GRAY{27.25}       &\GRAY{24.40}   &\GRAY{24.71}       &\GRAY{27.10}       &\GRAY{26.12}       &\GRAY{25.83}   &\GRAY{(\it25.90)}
        &\GRAY{26.19}       &\UL{15.48}     &\GRAY{38.10}       &\GRAY{38.10}       &\GRAY{61.90}       &\GRAY{29.76}   &\GRAY{(\it34.92)}
\\ 
    & \acrshort{SCG}
        %color              %shape          %texture            %2d-spatial         %non-spatial        %complex        %avg
        &\GRAY{49.82}       &\GRAY{43.28}   &\GRAY{50.16}       &\UL{16.31}         &\UL{66.60}         &\GRAY{51.07}   &\GRAY{(\it 46.21)}
        &\GRAY{27.86}       &\UL{24.85}     &\GRAY{25.57}       &\GRAY{27.76}       &\UL{26.98}         &26.03          &\GRAY{(\it26.51)}
        &\GRAY{20.24}       &\GRAY{11.90}   &\GRAY{33.33}       &\UL{40.48}         &\UL{69.05}         &\UL{39.29}     &\GRAY{(\it35.71)}
\\ 
\cmidrule(r){1-2}
\cmidrule(r){3-9}
\cmidrule(r){10-16}
\cmidrule{17-23}
\multirow{3}{*}{\rotatebox[origin=c]{90}{FT}} 
    & \acrshort{DPOK}
        %color              %shape          %texture            %2d-spatial         %non-spatial        %complex        %avg
        &\GRAY{53.28}       &\UL{45.63}     &\GRAY{52.84}       &\UL{17.19}         &\GRAY{66.95}       &\GRAY{51.97}   &(\it47.98)
        &\UL{28.20}         &\UL{24.99}     &\GRAY{25.44}       &\UL28.12           &\GRAY{26.80}       &\GRAY{25.88}   &(\it26.57)
        &\GRAY{23.81}       &\GRAY{16.67}   &\GRAY{47.62}       &\GRAY{34.52}       &\UL{70.24}         &\UL{38.10}     &(\it38.49)
\\
    & \acrshort{GORS}
        %color              %shape          %texture            %2d-spatial         %non-spatial        %complex        %avg
        &\UL{53.59}         &\GRAY{43.82}   &\UL{54.47}         &\GRAY{15.66}       &\UL{67.47}         &\UL{52.28}     &\GRAY{(\it47.88)}
        &\GRAY{28.15}       &\GRAY{24.79}   &\UL{25.56}         &\GRAY{27.90}       &\UL{26.88}         &\UL{26.07}     &\GRAY{(\it26.56)}
        &\UL{34.52}         &\GRAY{14.29}   &\UL{48.81}         &\UL{36.90}         &\GRAY{65.48}       &\GRAY{30.95}   &(\it38.49)
\\
    & \acrshort{HN}
        %color              %shape          %texture            %2d-spatial         %non-spatial        %complex        %avg
        &\GRAY{46.51}       &\GRAY{39.99}   &\GRAY{48.78}       &\GRAY{15.24}       &\GRAY{65.31}       &\GRAY{49.84}   &\GRAY{(\it44.28)}
        &\GRAY{26.90}       &\GRAY{24.33}   &\GRAY{24.63}       &\GRAY{27.15}       &\GRAY{25.40}       &\GRAY{25.22}   &\GRAY{(\it25.60)}
        &\GRAY{23.81}       &\UL{19.05}     &\GRAY{30.95}       &\GRAY{20.24}       &\GRAY{47.62}       &\GRAY{23.81}   &\GRAY{(\it27.58)}
\\ 
\cmidrule(r){1-2}
\cmidrule(r){3-9}
\cmidrule(r){10-16}
\cmidrule{17-23}
%%%%%%%%%%%%%%%%%%%%%%%%%%%%%%%%%%%%%%%
%%%%%%%%%%%%%%%%%%%%%%%%%%%%%%%%%%%%%%%
%%%%%%%%%%%%%%%%%%%%%%%%%%%%%%%%%%%%%%%
\multicolumn{2}{@{}r@{$\:\:\:\:$}}{\acrshort{MITUNE}}
        %color              %shape              %texture            %2d-spatial         %non-spatial        %complex            %avg
        &\CELCOLALT{65.04}  &\CELCOLALT{50.08}  &\CELCOLALT{65.82}  &$^\dagger\!$\CELCOLALT{18.51}  &\CELCOLALT{67.77}  &\CELCOLALT{54.17}  &\CELCOLALT{(\it53.56)}
        &\CELCOLALT{29.13}  &\CELCOLALT{25.57}  &\CELCOLALT{26.20}  &$^\dagger\!$\CELCOLALT{28.50}  &\CELCOLALT{27.15}  &\CELCOLALT{26.70}  &\CELCOLALT{(\it27.21)}
        &\CELCOLALT{46.43}  &\CELCOLALT{25.01}  &\CELCOLALT{53.19}  &$^\dagger\!$\CELCOLALT{45.24}  &\CELCOLALT{73.81}  &\CELCOLALT{46.43}  &\CELCOLALT{(\it48.35)}
\\
%%%%%%%%%%%%%%%%%%%%%%%%%%%%%%%%%%%%%%%%%%%%%%%%%%%%
%%%% ABS GAINS WRT BASE
%%%%%%%%%%%%%%%%%%%%%%%%%%%%%%%%%%%%%%%%%%%%%%%%%%%%
\cmidrule(r){1-2}
\cmidrule(r){3-9}
\cmidrule(r){10-16}
\cmidrule{17-23}
\multicolumn{2}{@{}r@{$\:\:\:\:$}}{\fontsize{8}{8}\selectfont{\textit{best} Infer.$\,\boxminus\,$base}}
	&\MONOCELGRADf11.78  &\MONOCELGRADd4.68   &\MONOCELGRADh14.11  &\MONOCELGRADa0.54   &\MONOCELGRADa0.37   &\MONOCELGRADa1.16   &\MONOCELGRADe(\it5.44)
	&\MONOCELGRADb0.80   &\MONOCELGRADa0.29   &\MONOCELGRADe0.89   &\MONOCELGRADg0.92   &\MONOCELGRADc0.32   &\MONOCELGRADa0.33   &\MONOCELGRADb(\it0.59)
	&\MONOCELGRADa2.19   &\MONOCELGRADa3.58   &\MONOCELGRADh11.90  &\MONOCELGRADd4.77   &\MONOCELGRADa2.38   &\MONOCELGRADa9.53   &\MONOCELGRADa(\it5.72)
\\
\multicolumn{2}{@{}r@{$\:\:\:\:$}}{\fontsize{8}{8}\selectfont{\textit{best} FT$\,\boxminus\,$base}}
	&\MONOCELGRADa3.94   &\MONOCELGRADa2.92   &\MONOCELGRADa4.48   &\MONOCELGRADd1.42   &\MONOCELGRADg1.24   &\MONOCELGRADb1.75   &\MONOCELGRADa(\it2.62)
	&\MONOCELGRADa0.56   &\MONOCELGRADb0.43   &\MONOCELGRADa0.57   &\MONOCELGRADa0.62   &\MONOCELGRADa0.22   &\MONOCELGRADa0.37   &\MONOCELGRADa(\it0.46)
	&\MONOCELGRADa4.76   &\MONOCELGRADc7.15   &\MONOCELGRADa8.33   &\MONOCELGRADa1.19   &\MONOCELGRADb3.57   &\MONOCELGRADa8.34   &\MONOCELGRADa(\it5.56)
\\
\multicolumn{2}{@{}r@{$\:\:\:\:$}}{\fontsize{8}{8}\selectfont{{\acrshort{MITUNE}$\,\boxminus$\,base}}}
	&\MONOCELGRADj15.39  &\MONOCELGRADj7.37   &\MONOCELGRADj15.83  &\MONOCELGRADj2.74   &\MONOCELGRADj1.54   &\MONOCELGRADj3.64   &\MONOCELGRADj(\it7.75)
	&\MONOCELGRADj1.49   &\MONOCELGRADj1.01   &\MONOCELGRADj1.21   &\MONOCELGRADj1.00   &\MONOCELGRADj0.49   &\MONOCELGRADj1.00   &\MONOCELGRADj(\it1.03)
	&\MONOCELGRADj16.67  &\MONOCELGRADj13.11  &\MONOCELGRADj12.71  &\MONOCELGRADj9.53   &\MONOCELGRADj7.14   &\MONOCELGRADj16.67  &\MONOCELGRADj(\it12.64)
\\
\cmidrule(r){1-2}
\cmidrule(r){3-9}
\cmidrule(r){10-16}
\cmidrule{17-23}
\multicolumn{2}{@{}r@{$\:\:\:\:$}}{\fontsize{8}{8}\selectfont{{\acrshort{MITUNE}$\,\boxminus$\,$\it best\star$}}}
	&3.61                &2.69                &1.72                &1.32                &0.30                &1.89                &(1.92)
	&0.69                &0.58                &0.32                &0.08                &0.17                &0.63                &(0.41)
	&11.91               &5.96                &0.81                &4.76                &3.57                &7.14                &(5.69)
\\
\multicolumn{2}{@{}r@{$\:\:\:\:$}}{\fontsize{8}{8}\selectfont{{\acrshort{MITUNE}$\,\%\,$\it best$\star$}}}
	&5.88                &5.68                &2.68                &7.68                &0.44                &3.62                &(4.33)
	&2.43                &2.32                &1.24                &0.28                &0.63                &2.42                &(1.55)
	&34.50               &31.29               &1.55                &11.76               &5.08                &18.17               &(17.06)
\\ 
\bottomrule
\end{tabular}
}
\\
\raggedright
\fontsize{7}{7}\selectfont
A$\,\boxminus\,$B indicates the absolute difference between A and B; A$\,\%\,$B corresponds to the percentage difference (A - B) / B; $\dagger$: Fine-tuning set obtained \\
from \acrshort{SPRIGHT} rather than \acrshort{SDBASE};
\vspace{-1pt}
Human scores do not sum to 100 in each category as users can select multiple methods for each question.
\end{table}

\subsection{Alternative methods}

\noindent \textbf{Inference-time methods.} Pre-trained model alignment can be improved at inference by
optimizing the latent variables $\mbz_t$ throughout the numerical integration used to generate the (latent) image. This process steers model alignment with an auxiliary loss based on attention maps and fine-grained linguistic analysis of the prompt (e.g., additional input is used to explicitly indicate which words to focus on). In this family, we consider 3 methods: \gls{AE}~\citep{chefer2023attendandexcite}, \gls{SDG}~\citep{feng2023Structured} and \gls{SCG}~\citep{shen2024SCFG}.

\noindent \textbf{Fine-tuning methods.} Alternatively, a pre-trained model can be fine-tuned with adapters~\citep{hu2021lora} optimized via a variety of RL or supervision methods. Specifically, we consider 3 approaches: \gls{DPOK}~\citep{fan2023dpok}, \gls{GORS}~\citep{huang2023t2icompbench} and \gls{HN}~\citep{krojer2023HNitm}. Notice that since results in the literature for both families do not
necessarily refer to same base models, to guarantee a fair comparison, we adapted and evaluated all methods on \acrshort{SDBASE}.

\subsection{Results}\label{sec:results}

\noindent \textbf{Comparing methods.}
\Cref{tab:results_blip} reports the alignment results on \acrshort{T2ICOMPBENCH}.
To simplify its reading, the bottom part of the table summarizes ($i$) the absolute gain with respect to the \acrshort{SDBASE} model for each of the best methods in each family and ($ii$) the percentage gains of \acrshort{MITUNE} with respect to the alternative method for each category.
We also summarize performance as averages across categories for each metric.

Despite performance varies, \acrshort{MITUNE} achieves a new state of the art across all categories/metrics,
often by a sizable margin. While this is more evident for \acrshort{BLIP} and Human,
the literature shows that \acrshort{HPS} has natural small variations
(see \Cref{app:hps_range}), hence \acrshort{MITUNE} gains are significant also for this metric.

\Cref{tab:results_blip} results are obtained generating fine-tuning sets from 
\acrshort{SDBASE} for all tasks but 2D-Spatial. For this category, we were able to obtain (at best) \acrshort{BLIP}=15.93 and \acrshort{HPS}=28.13. Conversely, generating the fine-tuning images from \acrshort{SPRIGHT} resulted beneficial.
We can link this result to the self-supervision nature of \acrshort{MITUNE}. On the one hand, our methodology is not bounded to a specific model. On the other hand, the filtering operated via point-wise \gls{MI} estimation can benefit from ``pre-alignment'' -- \acrshort{MITUNE} can strengthen existing alignment but might not be sufficient to ``induce'' it. 
Notice that all competitors suffer from this trade-off too as no single winner emerges.
In particular, despite \gls{AE} and \gls{GORS} are the most frequent best method in their family
(winning in 10-out-of-18 scenarios), all competitors show less consistent performance across categories and metrics than \acrshort{MITUNE}.
For instance, for attribute binding (color, shape and texture), 
fine-tuning methods under-perform
according to \acrshort{BLIP} and Human, but
the performance gaps are very close considering \acrshort{HPS}. 
Yet, \acrshort{MITUNE} achieves consistently higher performance across all categories, outperforming alternative fine-tuning methods by a large margin. 

Raw alignment performance apart, it is important to highlight \acrshort{MITUNE} key differences compared to the alternative fine-tuning methods. \acrshort{DPOK} uses RL with a reward model (pre-trained with human-labeled real images) to define a prompt-image alignment score to guide the fine-tuning, \acrshort{HN} uses a contrastive learning approach based on an ad-hoc dataset with real positive (good alignment) and negative (poor alignment) prompt-image pairs, and 
\acrshort{GORS} composes a fine-tuning set generating images from the diffusion model and selecting them based on \acrshort{BLIP}. While \acrshort{GORS} is very close in spirit to \acrshort{MITUNE}, its performance is ``biased'' -- the filtering criteria overlaps with the final evaluation strategy -- as explicitly acknowledged by its authors~\citep{huang2023t2icompbench}. Overall, while both \acrshort{DPOK} and \acrshort{GORS} still require external assistance, \acrshort{MITUNE} generates images \emph{and} selects them using the target model itself, i.e., it is the first fully self-supervised model for T2I alignment to the best of our knowledge. 

\vspace{-7pt}
\begin{minipage}[t]{0.32\textwidth}
\centering
\phantom{XXXXXX} %%% DO NOT REMOVE THIS
\vspace{-5pt}
\includegraphics[width=0.96\textwidth]{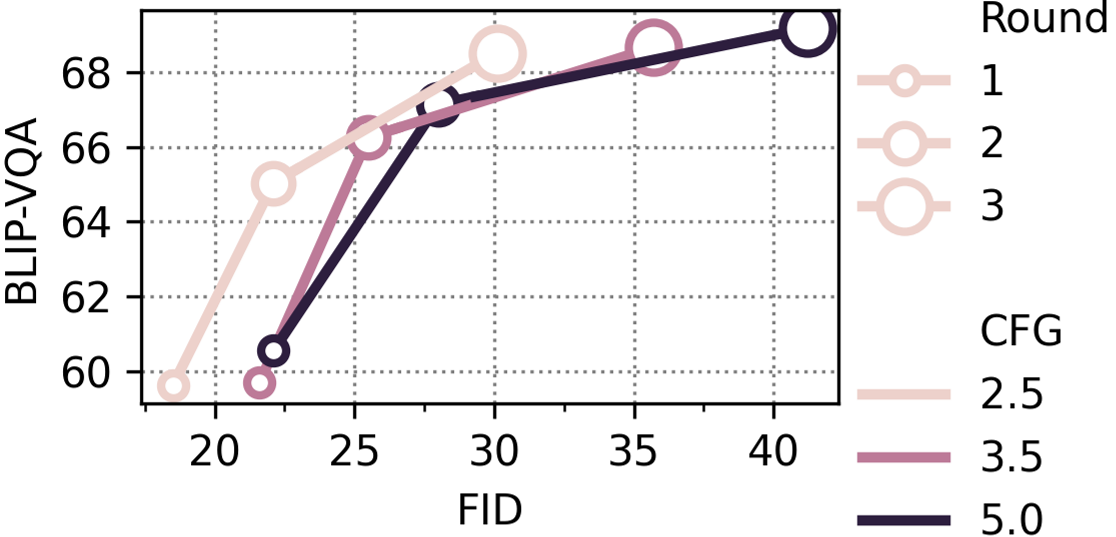}
\captionof{figure}{Hyper-params search. 
}
\label{fig:grid-search}
\end{minipage}
%%%%%%%
\begin{minipage}[t]{0.67\textwidth}
\small
\centering
\captionof{table}{Comparing image quality/variety scores.}
\label{tab:results_fid_r2_cfg2d5}
\vspace{-5pt}
\scalebox{0.73}{
\begin{tabular}{
@{}r
@{$\:$}r
%%%%%%%%%%
@{$\:\:$}r
@{$\:$}r
@{$\:$}r
@{$\:$}r
@{$\:$}r
@{$\:$}r
@{$\:$}r
%%%%%%%%%%%
@{$\:$}r
@{$\:$}r
@{$\:$}r
@{}
}
\toprule
&&\multicolumn{7}{c}{\acrshort{MITUNE} ($R$=2, \textit{CFG}=2.5)} &
\multirow{2}{*}{\TINY{\acrshort{DALLE}$^\ddagger$}} & 
\multirow{2}{*}{\TINY{\acrshort{IMAGEN}$^\ddagger$}} &
\multirow{2}{*}{\TINY{\acrshort{SDXL}$^\ddagger$}}
\\
\cmidrule(r){3-9}
\TINY{Metric} &
\TINY{\acrshort{SDBASE}} &
\TINY{Color} & 
\TINY{Shape} & 
\TINY{Texture} & 
\TINY{Spatial} & 
\TINY{Non-sp.} & 
\TINY{Comp.} &
\TINY{(\textit{avg})} &
& & 
\\
\cmidrule(r){1-1}
\cmidrule(r){2-2}
\cmidrule(r){3-9}
\cmidrule(r){10-12} 
\acrshort{FID}($\downarrow$)      
    &17.1       
    &22.1   &16.8   &17.3       &18.8       &16.8      &20.6    & (\textit{18.7})
    &20.1   &17.2   &13.2   
\\ 
\acrshort{DINO}($\downarrow$)  
    &229.1   
    &279.0   &236.9     &250.4   &251.7     &231.9      &255.6  & (\textit{250.9})
    &284.4   &213.9     &185.6          
\\
\acrshort{CMMD}($\downarrow$)  
    &0.641     
    &0.681   &0.634     &0.694   &0.669     &0.709      &0.671  & (\textit{0.680})
    &0.894   &0.854     &0.898          
\\ 
\bottomrule
\end{tabular}
}
\\
\fontsize{7}{7}\selectfont\it 
Results from 30k samples of \textsc{ms-coco-2014} validation set; 
$\ddagger$ results \textit{from} \citep{imagenteamgoogle2024imagen3}
\end{minipage}
%%%%%%%%%%

\noindent \textbf{Alignment/image quality-variety trade-offs.} 
\acrshort{MITUNE} results in \Cref{tab:results_blip} are obtained from a grid search across multiple fine-tuning rounds $R$ and \acrshort{CFG} values. In fact, we observe different trade-offs between alignment and image quality across different configurations.
We exemplify this in \Cref{fig:grid-search}, for the Color category.
The figure highlights two opposite dynamics:
T2I alignment benefits from multiple fine-tuning rounds (higher \acrshort{BLIP})
but can introduce image artifacts and reduce measured diversity
(higher \acrshort{FID}). While this trade-off is neither mentioned nor quantified in the literature of the considered methods, it is to be expected -- strictly abiding to a prompt 
impacts the ``generative pathways'' at sampling time. 
Interestingly, lowering \acrshort{CFG} (typically set to 7.5) counterbalances
these dynamics and enables a ``sweet spot'' -- as the model better aligns to a category thanks to fine-tuning, one can alleviate the guidance scale dependency at generation.
\Cref{tab:results_fid_r2_cfg2d5} complements this analysis by showing \acrshort{FID}, \acrshort{DINO} and \acrshort{CMMD} scores for all categories, as well for \acrshort{SDBASE} and three state of the art models -- while
all metrics indeed suggest a possible reduction in image variety considering \acrshort{SDBASE},
\acrshort{MITUNE} scores are comparable with other state-of-the-art models (see \Cref{fig:qualitative_sdbase} for example images).

%%%%%%%%%%%%%%%%%%%%%%%%%%%%
%%%%%%%%%%%%%%%%%%%%%%%%%%%%
%%%%%%%%%%%%%%%%%%%%%%%%%%%%
\vspace{2pt}
\begin{minipage}[t]{0.235\textwidth}
\vspace{-7pt}
\centering
\captionof{table}{
FT set selection.
\label{tab:results_finetuning_composition}
}
\vspace{-8pt}
\scalebox{0.58}{
\begin{tabular}{
@{}
l
@{$\:\;$}c
@{$\:$}c
%%%%%
@{$\:\:$}c
@{$\:$}c
@{}}
\toprule
    & \multicolumn{2}{c}{\bf\acrshort{BLIP}} 
    & \multicolumn{2}{c}{\bf\acrshort{HPS}}
\\
\cmidrule(r){2-3}
\cmidrule{4-5}
Strategy 
    & \TINY{Color} & \TINY{Shape} 
    & \TINY{Color} & \TINY{Shape}
\\
\cmidrule(r){1-1}
\cmidrule(r){2-3}
\cmidrule{4-5}
MI only             
    % &61.57 &48.40 % there are @R=1
    &65.04      &50.08
    &29.13      &25.57
\\
\cmidrule(r){1-1}
\cmidrule(r){2-3}
\cmidrule{4-5}
HPS only            
    &59.43              &46.87 
    &\TINY{n.a.}   &\TINY{n.a.}
\\
\cmidrule(r){1-1}
\cmidrule(r){2-3}
\cmidrule{4-5}
MI+Real(0.25)
    &61.34      &48.47
    &29.16         &25.87
\\
MI+Real(0.5)   
    &61.63      &49.50
    &29.38          &25.92
\\
MI+Real(0.9)    
    &59.83      &48.92
    &28.60          &25.60
\\
\bottomrule
\end{tabular}
}
%%%%%%%%%%%%%%%%
%%%%%%%%%%%%%%%%
%%%%%%%%%%%%%%%%
\end{minipage}
\begin{minipage}[t]{0.56\textwidth}
\vspace{-7pt}
\captionof{table}{
Alignment (\%) using \acrshort{SDXL}.
\label{tab:sdxl}
}
\vspace{-8pt}
\scalebox{0.6}{
\begin{tabular}{
@{}r
@{$\:\:$}r
@{$\:$}r
@{$\:$}r
@{$\:$}r
@{$\:$}r
@{$\:$}r
%%%%%%%%%
@{$\:\:$}r
@{$\:$}r
@{$\:$}r
@{$\:$}r
@{$\:$}r
@{$\:$}r
@{}
}
\toprule
& \multicolumn{6}{c}{\bf\acrshort{BLIP}}
& \multicolumn{6}{c}{\bf\acrshort{HPS}}
\\
\cmidrule(r){2-7}
\cmidrule{8-13}
Method               
    &\TINY{Color}     &\TINY{Shape}      &\TINY{Texture}    &\TINY{2D-Sp.} &\TINY{Non-Sp.} &\TINY{Comp.}
    &\TINY{Color}     &\TINY{Shape}      &\TINY{Texture}    &\TINY{2D-Sp.} &\TINY{Non-Sp.} &\TINY{Comp.}
\\
\cmidrule(r){1-1}
\cmidrule(r){2-7}
\cmidrule{8-13}
\textit{(ref)} \acrshort{SDXL}      
    &60.78     &49.70      &55.78   &21.02      &68.16     &52.68
    &28.47     &24.99      &25.85   &28.50     &26.64    &25.90   
\\
\TINY{\acrshort{SDBASE}}
    &49.65     &42.71      &49.99   &15.77  &66.23 &50.53 
    &27.64     &24.56      &24.99   &27.50  &26.66 &25.70
\\
\cmidrule(r){1-1}
\cmidrule(r){2-7}
\cmidrule{8-13}
\acrshort{MITUNE}    
    &69.66     &55.86      &66.74   &22.18      &72.17     &57.74
    &29.03     &25.90      &27.15   &29.57      &27.56     &26.70 
\\
\cmidrule(r){1-1}
\cmidrule(r){2-7}
\cmidrule{8-13}
{\fontsize{8}{8}\selectfont\acrshort{MITUNE}$\,\boxminus\,$\it (ref)}
	 &8.88      &6.16      &10.96     &1.16         &4.01         &5.06      
	 &0.56      &0.91      &1.30      &1.07         &0.92         &0.80      
\\
{\fontsize{8}{8}\selectfont\acrshort{MITUNE}$\,\%\,$\it (ref)}
	 &14.61     &12.39     &19.65     &5.52         &5.88         &9.61      
	 &1.97      &3.64      &5.03      &3.75         &3.45         &3.09      
\\
\bottomrule
\end{tabular}
}
\end{minipage}
%%%%%%%%%%%%%%%%
%%%%%%%%%%%%%%%%
%%%%%%%%%%%%%%%%
\begin{minipage}[t]{0.193\textwidth}
\vspace{-7pt}
\centering
\captionof{table}{
DiffusionDB.
}
\label{tab:results_diffusionDB}
\vspace{-8pt}
\scalebox{0.64}{
\begin{tabular}{
@{}
l 
@{$\:\:\:\:$}
l
@{}}
\toprule
\bf Model                             & \bf \acrshort{HPS} \\
\midrule
\acrshort{SDBASE}                     & 23.99    \\
DiffusionDB                           & 24.35    \\
\midrule
\acrshort{MITUNE}                     & 25.32    
\\
\midrule

{\fontsize{9}{9}\selectfont\acrshort{MITUNE}$\,\boxminus\,$\it base}
	 &1.33   
\\
{\fontsize{9}{9}\selectfont\acrshort{MITUNE}$\,\boxminus\,$\it DiffusionDB}
	 &0.97
\\
\bottomrule
\end{tabular}
}
\end{minipage}

\noindent\textbf{Fine-tuning set composition.}
The strategy to select prompt-image pairs for the fine-tuning set
has a large design space beyond the use of \gls{MI}. In \Cref{tab:results_finetuning_composition}, 
we report (for two categories for brevity) alignment performance using two alternative
strategies. Specifically, using \acrshort{HPS} rather than \gls{MI} degrades performance.\footnote{We compute only \acrshort{BLIP} to avoid evaluation bias~\citep{huang2023t2icompbench}.}
Results when composing the fine-tuning set by mixing \gls{MI}-selected and real images 
selected from the $\textsc{cc2m}$ dataset \citep{changpinyo2021conceptual} are instead inconsistent
(\acrshort{BLIP} steadily degrades but \acrshort{HPS} signals an 
improvement in some scenarios). 

\noindent\textbf{\acrshort{SDXL} and \small DiffusionDB.}
We complete our evaluation by presenting results obtained applying \acrshort{MITUNE} on \acrshort{SDXL} in \Cref{tab:sdxl}, and considering
an alternative scenario closer to real user application using DiffusionDB in \Cref{tab:results_diffusionDB} to complement the synthetic nature of \acrshort{T2ICOMPBENCH}. 
As expected, ``vanilla'' \acrshort{SDXL} 
significantly outperforms \acrshort{SDBASE}, yet \acrshort{MITUNE}
enables sizable improvements on \acrshort{SDXL} alignment (see \Cref{fig:qualitative_sdxl}). 
For the realistic alignment use case in \Cref{tab:results_diffusionDB}, we select prompt-images pairs from DiffusionDB
and we contrast alignment when fine-tuning using the images already paired with prompts
against \gls{MI}-selected ones.
We use \acrshort{SDBASE} as base model and report only \acrshort{HPS} 
scores\footnote{The higher prompt complexity 
does not well suit \acrshort{BLIP} text decomposition   
(see \Cref{app:diffusion_db_results}).} in \Cref{tab:results_diffusionDB}. 
Overall, fine-tuning with DiffusionDB images
improves the base model, yet \acrshort{MITUNE} enables superior performance
(see \Cref{fig:qualitative_diffusiondb}).

\vspace{-0.7em}
\begin{minipage}[t]{0.5\textwidth}
\fontsize{7}{7}\selectfont
\begin{tabular}{
@{}p{27pt}
@{}p{27pt}
@{}p{27pt}
@{}p{27pt}
@{}p{26pt}
@{}p{26pt}
@{}p{26pt}
@{}p{26pt}
@{}}
\fontsize{4.9}{4.9}\selectfont{\acrshort{SDBASE}} &
\hfil\fontsize{5.8}{5.8}\selectfont{\acrshort{DPOK}} &
\hfil\fontsize{5.8}{5.8}\selectfont{\acrshort{GORS}} &
\hfil\fontsize{5.8}{5.8}\selectfont{\acrshort{HN}} &
\hfil\fontsize{5.8}{5.8}\selectfont{\acrshort{AE}} &
\hfil\fontsize{5.8}{5.8}\selectfont{\acrshort{SDG}} &
\hfil\fontsize{5.8}{5.8}\selectfont{\acrshort{SCG}} &
\hfil\fontsize{5.8}{5.8}\selectfont{\acrshort{MITUNE}}
\\
\end{tabular}

\vspace{3pt}
%%%%%%%%%%%%%%%%%%%%%%%%%%%5
% COLOR
%%%%%%%%%%%%%%%%%%%%%%%%%%%5
\centering
\includegraphics[width=\linewidth]{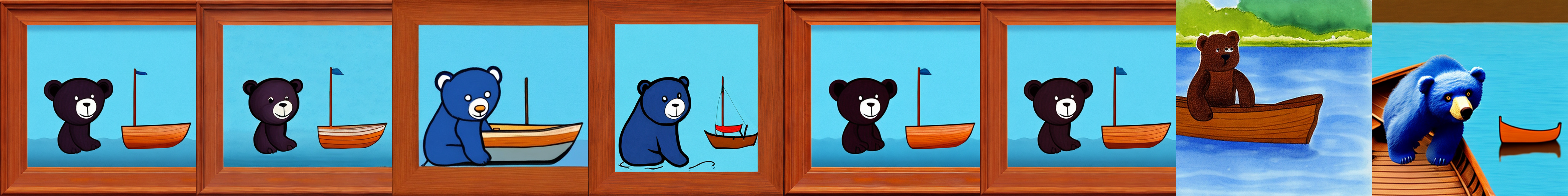}
(Color) ``\textit{a blue bear and a brown boat}''

%%%%%%%%%%%%%%%%%%%%%%%%%%%5
% SHAPE
%%%%%%%%%%%%%%%%%%%%%%%%%%%5
\phantom{XXXXX}
\includegraphics[width=\linewidth]{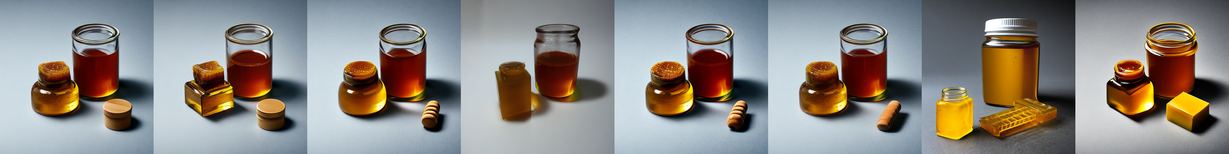}
(Shape) ``\textit{a cubic block and a cylindrical jar of honey}''

%%%%%%%%%%%%%%%%%%%%%%%%%%%5
% TEXTURE
%%%%%%%%%%%%%%%%%%%%%%%%%%%5
\phantom{XXXXX}
\includegraphics[width=\linewidth]{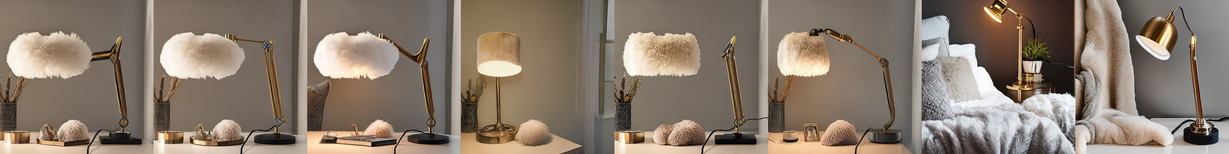}
\hfill
(Texture) ``\textit{a metallic desk lamp and a fluffy blanket}''

%%%%%%%%%%%%%%%%%%%%%%%%%%%5
% 2D-SPATIAL
%%%%%%%%%%%%%%%%%%%%%%%%%%%5
\phantom{XXXXX}
\includegraphics[width=\linewidth]{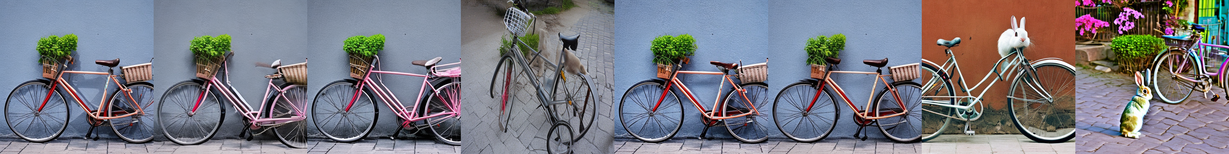}
\hfill
(2D-Spatial) ``\textit{a rabbit near a bicycle}''

%%%%%%%%%%%%%%%%%%%%%%%%%%%5
% NON-SPATIAL
%%%%%%%%%%%%%%%%%%%%%%%%%%%5
\phantom{XXXXX}
\includegraphics[width=\linewidth]{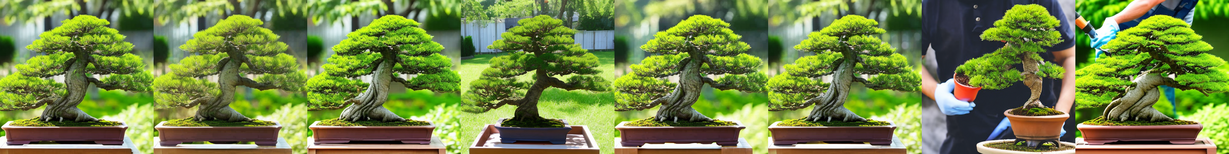}
\hfill
(Non-spatial) ``\textit{A gardener is pruning a beautiful bonsai tree.}''

%%%%%%%%%%%%%%%%%%%%%%%%%%%5
% COMPLEX
%%%%%%%%%%%%%%%%%%%%%%%%%%%5
\phantom{XXXXX}
\includegraphics[width=\linewidth]{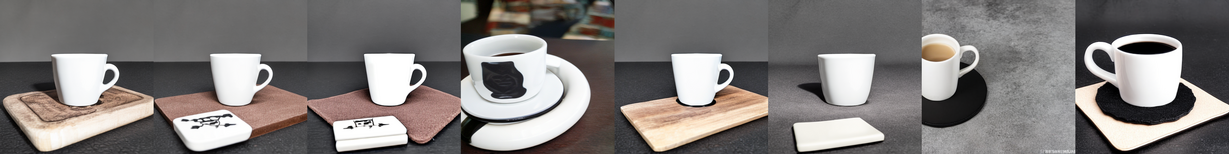}
(Complex) ``\textit{The white mug is on top of the black coaster.}''

%%%%%%%%%%%%%%%%%%%%
\centering
\captionof{figure}{
Qualitative examples from \Cref{tab:results_blip} (same seed used for a given prompt).
More examples in \Cref{app:qualitative_sdbase}.
}
\label{fig:qualitative_sdbase}
\end{minipage}
%%%%%%%%%%%%%%%%%%%%%%%%%%%%%%%%%
%%%%%%%%%%%%%%%%%%%%%%%%%%%%%%%%%
\hspace{3pt}
%%%%%%%%%%%%%%%%%%%%%%%%%%%%%%%%%
%%%%%%%%%%%%%%%%%%%%%%%%%%%%%%%%%
\begin{minipage}[t]{0.48\textwidth}
\fontsize{7}{7}\selectfont
\begin{tabular}{
@{}
p{48pt}@{}
p{48pt}@{}
p{2pt}@{}
p{48pt}@{}
p{48pt}@{}
}
\hfil\fontsize{5.8}{5.8}\selectfont{\acrshort{SDXL}} &
\hfil\fontsize{5.8}{5.8}\selectfont{\acrshort{MITUNE}} &
&
\hfil\fontsize{5.8}{5.8}\selectfont{\acrshort{SDXL}} &
\hfil\fontsize{5.8}{5.8}\selectfont{\acrshort{MITUNE}}
\\
\end{tabular}

\includegraphics[width=0.46\linewidth]{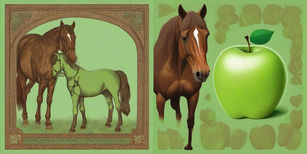}
\hspace{2pt}
\includegraphics[width=0.46\linewidth]{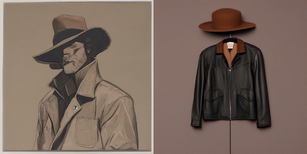}
\\
\fontsize{5.5}{5.5}\selectfont
(Color) ``A green apple and a brown horse'' \hspace{3pt} ``A black jacked and a brown hat''

\vspace{3pt}
\includegraphics[width=0.46\linewidth]{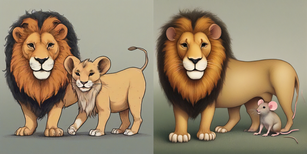}
\hspace{2pt}
\includegraphics[width=0.46\linewidth]{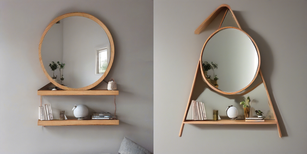}
\\
\fontsize{5.5}{5.5}\selectfont
(Shape) ``A big lion and a small mouse'' \hspace{6pt} ``A circular mirror and a triangular shelf unit''

\vspace{-3pt}
\centering
\captionof{figure}{
Qualitative examples from \Cref{tab:sdxl} (same seed used for a given prompt).
More examples in \Cref{app:qualitative_sdxl}.
\label{fig:qualitative_sdxl}
}

%%%%%%%%%%%%%%%%%%%%%%%%%%%%%%%%%
%%%%%%%%%%%%%%%%%%%%%%%%%%%%%%%%%
% \hspace{3pt}
%%%%%%%%%%%%%%%%%%%%%%%%%%%%%%%%%
%%%%%%%%%%%%%%%%%%%%%%%%%%%%%%%%%
\vspace{5pt}
\centering

\begin{tabular}{
@{}
p{52pt}@{}
p{52pt}@{}
p{52pt}@{}
}
\hfil\fontsize{5.8}{5.8}\selectfont{\acrshort{SDBASE}} &
\hfil\fontsize{5.8}{5.8}\selectfont{Fine-tuned using \newline DiffusionDB images} &
\hfil\fontsize{5.8}{5.8}\selectfont{\acrshort{MITUNE}}
\\
\end{tabular}

\includegraphics[width=0.8\linewidth]{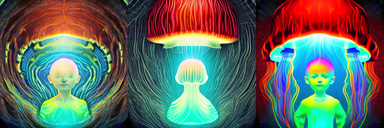}
\\
(Human prompt) ``Child's body with a radioactive jellyfish as a head, realistic illustration, backlit, intricate, indie studio, fantasy, rim lighting, vibrant colors, emotional''

\vspace{3pt}
\includegraphics[width=0.8\linewidth]{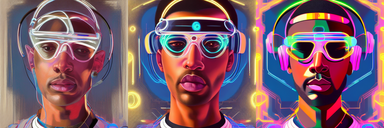}
\\
(Human Prompt) ``Digital neon cyberpunk male with geordi eye visor and headphones portrait painting by donato giancola, kilian eng, john berkey, j. c. leyendecker, alphonse mucha''
\centering
\captionof{figure}{
Qualitative examples from \Cref{tab:results_diffusionDB} (same seed used for a given prompt). 
More examples in \Cref{app:qualitative_diffusiondb}.
\label{fig:qualitative_diffusiondb}
}
\end{minipage}

\vspace{-1em}
\section{Conclusion}\label{sec:conclusion}
\gls{T2I} alignment emerged as an important endeavor to steer image generation to follow the semantics and user intent expressed through a natural text prompt, as it can save considerable manual effort.
In this work, we presented a novel approach to improve model alignment, that uses point-wise \gls{MI} between prompt-image pairs as a meaningful signal to evaluate the amount of information ``flowing'' between natural text and images. We demonstrated, both qualitatively and quantitatively, that point-wise \gls{MI} is coherent with existing alignment measures that either use auxiliary \gls{VQA} models or elicit human intervention.

We presented \acrshort{MITUNE}, a lightweight, self-supervised fine-tuning method that uses a pre-trained \gls{T2I} model such as \gls{SD} to estimate \gls{MI}, and to generate a synthetic set of aligned prompt-image pairs, which is then used in a parameter-efficient fine-tuning stage, to align the \gls{T2I} model.
Our approach does not require human annotation, auxiliary \gls{VQA} models, nor costly inference-time techniques, and achieves a new state-of-the-art across all categories/metrics explored in the literature, often by a sizable margin. These results carry on in more complex tasks, and for various base models, illustrating the flexibility of \acrshort{MITUNE}.

\clearpage

\newpage
{\small
\bibliographystyle{iclr2025_conference}
\bibliography{main}

\begin{thebibliography}{106}
\providecommand{\natexlab}[1]{#1}
\providecommand{\url}[1]{\texttt{#1}}
\expandafter\ifx\csname urlstyle\endcsname\relax
  \providecommand{\doi}[1]{doi: #1}\else
  \providecommand{\doi}{doi: \begingroup \urlstyle{rm}\Url}\fi

\bibitem[Agarwal et~al.(2023)Agarwal, Karanam, Joseph, Saxena, Goswami, and
  Srinivasan]{agarwal2023}
Aishwarya Agarwal, Srikrishna Karanam, K~J Joseph, Apoorv Saxena, Koustava
  Goswami, and Balaji~Vasan Srinivasan.
\newblock A-star: Test-time attention segregation and retention for
  text-to-image synthesis.
\newblock In \emph{Proceedings of the IEEE/CVF International Conference on
  Computer Vision (ICCV)}, pp.\  2283--2293, October 2023.

\bibitem[Alemi \& Fischer(2018)Alemi and Fischer]{alemi2019gilbo}
Alexander~A Alemi and Ian Fischer.
\newblock Gilbo: One metric to measure them all.
\newblock \emph{Advances in Neural Information Processing Systems}, 31, 2018.

\bibitem[Alemi et~al.(2016)Alemi, Fischer, Dillon, and Murphy]{alemi2019deep}
Alexander~A Alemi, Ian Fischer, Joshua~V Dillon, and Kevin Murphy.
\newblock Deep variational information bottleneck.
\newblock In \emph{International Conference on Learning Representations}, 2016.

\bibitem[Balaji et~al.(2022{\natexlab{a}})Balaji, Nah, Huang, Vahdat, Song,
  Kreis, Aittala, Aila, Laine, Catanzaro, Karras, and Liu]{balaji2022ediff}
Yogesh Balaji, Seungjun Nah, Xun Huang, Arash Vahdat, Jiaming Song, Karsten
  Kreis, Miika Aittala, Timo Aila, Samuli Laine, Bryan Catanzaro, Tero Karras,
  and Ming-Yu Liu.
\newblock ediff-i: Text-to-image diffusion models with ensemble of expert
  denoisers.
\newblock \emph{arXiv preprint arXiv:2211.01324}, 2022{\natexlab{a}}.

\bibitem[Balaji et~al.(2022{\natexlab{b}})Balaji, Nah, Huang, Vahdat, Song,
  Zhang, Kreis, Aittala, Aila, Laine, Catanzaro, Karras, and Liu]{balaji2022}
Yogesh Balaji, Seungjun Nah, Xun Huang, Arash Vahdat, Jiaming Song, Qinsheng
  Zhang, Karsten Kreis, Miika Aittala, Timo Aila, Samuli Laine, Bryan
  Catanzaro, Tero Karras, and Ming-Yu Liu.
\newblock ediff-i: Text-to-image diffusion models with ensemble of expert
  denoisers.
\newblock \emph{arXiv preprint arXiv:2211.01324}, 2022{\natexlab{b}}.

\bibitem[Belghazi et~al.(2018)Belghazi, Baratin, Rajeshwar, Ozair, Bengio,
  Courville, and Hjelm]{belghazi2018mine}
Mohamed~Ishmael Belghazi, Aristide Baratin, Sai Rajeshwar, Sherjil Ozair,
  Yoshua Bengio, Aaron Courville, and Devon Hjelm.
\newblock Mutual information neural estimation.
\newblock In \emph{Proceedings of the 35th International Conference on Machine
  Learning}, 2018.

\bibitem[Bell \& Sejnowski(1995)Bell and Sejnowski]{bell1995information}
Anthony~J Bell and Terrence~J Sejnowski.
\newblock An information-maximization approach to blind separation and blind
  deconvolution.
\newblock \emph{Neural computation}, 7\penalty0 (6):\penalty0 1129--1159, 1995.

\bibitem[Brekelmans et~al.(2022)Brekelmans, Huang, Ghassemi, Steeg, Grosse, and
  Makhzani]{brekelmans2023improving}
Rob Brekelmans, Sicong Huang, Marzyeh Ghassemi, Greg~Ver Steeg, Roger~Baker
  Grosse, and Alireza Makhzani.
\newblock Improving mutual information estimation with annealed and
  energy-based bounds.
\newblock In \emph{International Conference on Learning Representations}, 2022.

\bibitem[Changpinyo et~al.(2021)Changpinyo, Sharma, Ding, and
  Soricut]{changpinyo2021conceptual}
Soravit Changpinyo, Piyush Sharma, Nan Ding, and Radu Soricut.
\newblock Conceptual 12m: Pushing web-scale image-text pre-training to
  recognize long-tail visual concepts.
\newblock In \emph{Proceedings of the IEEE/CVF conference on computer vision
  and pattern recognition}, pp.\  3558--3568, 2021.

\bibitem[Chatterjee et~al.(2024)Chatterjee, Stan, Aflalo, Paul, Ghosh, Gokhale,
  Schmidt, Hajishirzi, Lal, Baral, and Yang]{chatterjee2024SpRight}
Agneet Chatterjee, Gabriela Ben~Melech Stan, Estelle Aflalo, Sayak Paul, Dhruba
  Ghosh, Tejas Gokhale, Ludwig Schmidt, Hannaneh Hajishirzi, Vasudev Lal,
  Chitta Baral, and Yezhou Yang.
\newblock Getting it right: Improving spatial consistency in text-to-image
  models, 2024.

\bibitem[Chefer et~al.(2023{\natexlab{a}})Chefer, Alaluf, Vinker, Wolf, and
  Cohen-Or]{chefer2023}
Hila Chefer, Yuval Alaluf, Yael Vinker, Lior Wolf, and Daniel Cohen-Or.
\newblock Attend-and-excite: Attention-based semantic guidance for
  text-to-image diffusion models.
\newblock \emph{ACM Trans. Graph.}, 42\penalty0 (4), 2023{\natexlab{a}}.

\bibitem[Chefer et~al.(2023{\natexlab{b}})Chefer, Alaluf, Vinker, Wolf, and
  Cohen-Or]{chefer2023attendandexcite}
Hila Chefer, Yuval Alaluf, Yael Vinker, Lior Wolf, and Daniel Cohen-Or.
\newblock Attend-and-excite: Attention-based semantic guidance for
  text-to-image diffusion models, 2023{\natexlab{b}}.

\bibitem[Chen et~al.(2016)Chen, Duan, Houthooft, Schulman, Sutskever, and
  Abbeel]{chen2016infogan}
Xi~Chen, Yan Duan, Rein Houthooft, John Schulman, Ilya Sutskever, and Pieter
  Abbeel.
\newblock Infogan: Interpretable representation learning by information
  maximizing generative adversarial nets.
\newblock \emph{Advances in neural information processing systems}, 29, 2016.

\bibitem[Clark et~al.(2024)Clark, Vicol, Swersky, and Fleet]{clark2024directly}
Kevin Clark, Paul Vicol, Kevin Swersky, and David~J. Fleet.
\newblock Directly fine-tuning diffusion models on differentiable rewards.
\newblock In \emph{The Twelfth International Conference on Learning
  Representations}, 2024.
\newblock URL \url{https://openreview.net/forum?id=1vmSEVL19f}.

\bibitem[Conwell \& Ullman(2022)Conwell and Ullman]{conwell2022}
Colin Conwell and Tomer Ullman.
\newblock Testing relational understanding in text-guided image generation,
  2022.

\bibitem[Dahary et~al.(2024)Dahary, Patashnik, Aberman, and
  Cohen-Or]{dahary2024}
Omer Dahary, Or~Patashnik, Kfir Aberman, and Daniel Cohen-Or.
\newblock Be yourself: Bounded attention for multi-subject text-to-image
  generation, 2024.

\bibitem[Dhariwal \& Nichol(2021)Dhariwal and Nichol]{dhariwal2021}
Prafulla Dhariwal and Alexander Nichol.
\newblock Diffusion models beat gans on image synthesis.
\newblock In M.~Ranzato, A.~Beygelzimer, Y.~Dauphin, P.S. Liang, and J.~Wortman
  Vaughan (eds.), \emph{Advances in Neural Information Processing Systems},
  volume~34, pp.\  8780--8794. Curran Associates, Inc., 2021.

\bibitem[Fan et~al.(2023)Fan, Watkins, Du, Liu, Ryu, Boutilier, Abbeel,
  Ghavamzadeh, Lee, and Lee]{fan2023dpok}
Ying Fan, Olivia Watkins, Yuqing Du, Hao Liu, Moonkyung Ryu, Craig Boutilier,
  Pieter Abbeel, Mohammad Ghavamzadeh, Kangwook Lee, and Kimin Lee.
\newblock Reinforcement learning for fine-tuning text-to-image diffusion
  models.
\newblock In \emph{Thirty-seventh Conference on Neural Information Processing
  Systems}, 2023.
\newblock URL \url{https://openreview.net/forum?id=8OTPepXzeh}.

\bibitem[Feng et~al.(2023{\natexlab{a}})Feng, He, Fu, Jampani, Akula, Narayana,
  Basu, Wang, and Wang]{feng2023}
Weixi Feng, Xuehai He, Tsu-Jui Fu, Varun Jampani, Arjun~Reddy Akula, Pradyumna
  Narayana, Sugato Basu, Xin~Eric Wang, and William~Yang Wang.
\newblock Training-free structured diffusion guidance for compositional
  text-to-image synthesis.
\newblock In \emph{The Eleventh International Conference on Learning
  Representations}, 2023{\natexlab{a}}.
\newblock URL \url{https://openreview.net/forum?id=PUIqjT4rzq7}.

\bibitem[Feng et~al.(2023{\natexlab{b}})Feng, He, Fu, Jampani, Akula, Narayana,
  Basu, Wang, and Wang]{feng2023Structured}
Weixi Feng, Xuehai He, Tsu-Jui Fu, Varun Jampani, Arjun~Reddy Akula, Pradyumna
  Narayana, Sugato Basu, Xin~Eric Wang, and William~Yang Wang.
\newblock Training-free structured diffusion guidance for compositional
  text-to-image synthesis.
\newblock In \emph{The Eleventh International Conference on Learning
  Representations}, 2023{\natexlab{b}}.
\newblock URL \url{https://openreview.net/forum?id=PUIqjT4rzq7}.

\bibitem[Franzese et~al.(2024)Franzese, Bounoua, and
  Michiardi]{franzese2024minde}
Giulio Franzese, Mustapha Bounoua, and Pietro Michiardi.
\newblock {MINDE}: Mutual information neural diffusion estimation.
\newblock In \emph{The Twelfth International Conference on Learning
  Representations}, 2024.
\newblock URL \url{https://openreview.net/forum?id=0kWd8SJq8d}.

\bibitem[Gafni et~al.(2022)Gafni, Polyak, Ashual, Sheynin, Parikh, and
  Taigman]{gafni2022}
Oran Gafni, Adam Polyak, Oron Ashual, Shelly Sheynin, Devi Parikh, and Yaniv
  Taigman.
\newblock Make-a-scene: Scene-based text-to-image generation with human priors,
  2022.

\bibitem[Gordon et~al.(2023)Gordon, Bitton, Shafir, Garg, Chen, Lischinski,
  Cohen-Or, and Szpektor]{gordon2023mismatch}
Brian Gordon, Yonatan Bitton, Yonatan Shafir, Roopal Garg, Xi~Chen, Dani
  Lischinski, Daniel Cohen-Or, and Idan Szpektor.
\newblock Mismatch quest: Visual and textual feedback for image-text
  misalignment, 2023.

\bibitem[Grimal et~al.(2024)Grimal, Le~Borgne, Ferret, and
  Tourille]{grimal2024tiam}
Paul Grimal, Herv\'e Le~Borgne, Olivier Ferret, and Julien Tourille.
\newblock Tiam - a metric for evaluating alignment in text-to-image generation.
\newblock In \emph{Proceedings of the IEEE/CVF Winter Conference on
  Applications of Computer Vision (WACV)}, pp.\  2890--2899, January 2024.

\bibitem[Hessel et~al.(2021)Hessel, Holtzman, Forbes, Le~Bras, and
  Choi]{hessel2021-clipscore}
Jack Hessel, Ari Holtzman, Maxwell Forbes, Ronan Le~Bras, and Yejin Choi.
\newblock {CLIPS}core: A reference-free evaluation metric for image captioning.
\newblock In \emph{Proceedings of the 2021 Conference on Empirical Methods in
  Natural Language Processing}, 2021.

\bibitem[Heusel et~al.(2017)Heusel, Ramsauer, Unterthiner, Nessler, and
  Hochreiter]{fid}
Martin Heusel, Hubert Ramsauer, Thomas Unterthiner, Bernhard Nessler, and Sepp
  Hochreiter.
\newblock Gans trained by a two time-scale update rule converge to a local nash
  equilibrium.
\newblock In I.~Guyon, U.~Von Luxburg, S.~Bengio, H.~Wallach, R.~Fergus,
  S.~Vishwanathan, and R.~Garnett (eds.), \emph{Advances in Neural Information
  Processing Systems}, volume~30. Curran Associates, Inc., 2017.
\newblock URL
  \url{https://proceedings.neurips.cc/paper_files/paper/2017/file/8a1d694707eb0fefe65871369074926d-Paper.pdf}.

\bibitem[Hjelm et~al.(2019)Hjelm, Fedorov, Lavoie-Marchildon, Grewal, Bachman,
  Trischler, and Bengio]{hjelm2019learning}
R~Devon Hjelm, Alex Fedorov, Samuel Lavoie-Marchildon, Karan Grewal, Phil
  Bachman, Adam Trischler, and Yoshua Bengio.
\newblock Learning deep representations by mutual information estimation and
  maximization.
\newblock In \emph{International Conference on Learning Representations}, 2019.

\bibitem[Ho \& Salimans(2021)Ho and Salimans]{ho2021classifierfree}
Jonathan Ho and Tim Salimans.
\newblock Classifier-free diffusion guidance.
\newblock In \emph{NeurIPS 2021 Workshop on Deep Generative Models and
  Downstream Applications}, 2021.
\newblock URL \url{https://openreview.net/forum?id=qw8AKxfYbI}.

\bibitem[Ho \& Salimans(2022)Ho and Salimans]{ho2022classifier}
Jonathan Ho and Tim Salimans.
\newblock Classifier-free diffusion guidance.
\newblock \emph{arXiv preprint arXiv:2207.12598}, 2022.

\bibitem[Ho et~al.(2020)Ho, Jain, and Abbeel]{ho2020}
Jonathan Ho, Ajay Jain, and Pieter Abbeel.
\newblock Denoising diffusion probabilistic models.
\newblock In \emph{Advances in Neural Information Processing Systems},
  volume~33, pp.\  6840--6851, 2020.

\bibitem[Hu et~al.(2021)Hu, Shen, Wallis, Allen-Zhu, Li, Wang, Wang, and
  Chen]{hu2021lora}
Edward~J. Hu, Yelong Shen, Phillip Wallis, Zeyuan Allen-Zhu, Yuanzhi Li, Shean
  Wang, Lu~Wang, and Weizhu Chen.
\newblock Lora: Low-rank adaptation of large language models, 2021.

\bibitem[Hu et~al.(2023)Hu, Liu, Kasai, Wang, Ostendorf, Krishna, and
  Smith]{hu2023tifa}
Yushi Hu, Benlin Liu, Jungo Kasai, Yizhong Wang, Mari Ostendorf, Ranjay
  Krishna, and Noah~A. Smith.
\newblock Tifa: Accurate and interpretable text-to-image faithfulness
  evaluation with question answering.
\newblock In \emph{Proceedings of the IEEE/CVF International Conference on
  Computer Vision (ICCV)}, pp.\  20406--20417, October 2023.

\bibitem[Huang et~al.(2023)Huang, Sun, Xie, Li, and Liu]{huang2023t2icompbench}
Kaiyi Huang, Kaiyue Sun, Enze Xie, Zhenguo Li, and Xihui Liu.
\newblock T2i-compbench: A comprehensive benchmark for open-world compositional
  text-to-image generation.
\newblock In \emph{Thirty-seventh Conference on Neural Information Processing
  Systems Datasets and Benchmarks Track}, 2023.
\newblock URL \url{https://openreview.net/forum?id=weHBzTLXpH}.

\bibitem[Huang et~al.(2020)Huang, Makhzani, Cao, and
  Grosse]{huang2020evaluating}
Sicong Huang, Alireza Makhzani, Yanshuai Cao, and Roger Grosse.
\newblock Evaluating lossy compression rates of deep generative models.
\newblock In \emph{International Conference on Machine Learning}. PMLR, 2020.

\bibitem[HuggingFace(2022)]{hfnegative}
HuggingFace.
\newblock Negative prompts.
\newblock
  \url{https://huggingface.co/spaces/stabilityai/stable-diffusion/discussions/7857},
  2022.

\bibitem[Imagen-Team et~al.(2024)]{imagenteamgoogle2024imagen3}
Imagen-Team et~al.
\newblock Imagen 3, 2024.

\bibitem[Jayasumana et~al.(2024)Jayasumana, Ramalingam, Veit, Glasner,
  Chakrabarti, and Kumar]{cmmd}
Sadeep Jayasumana, Srikumar Ramalingam, Andreas Veit, Daniel Glasner, Ayan
  Chakrabarti, and Sanjiv Kumar.
\newblock Rethinking fid: Towards a better evaluation metric for image
  generation.
\newblock In \emph{Proceedings of the IEEE/CVF Conference on Computer Vision
  and Pattern Recognition (CVPR)}, pp.\  9307--9315, June 2024.

\bibitem[Jiang et~al.(2024)Jiang, Song, Wu, Zhang, Shen, Zong, Liu, and
  Li]{jiang2024CoMat}
Dongzhi Jiang, Guanglu Song, Xiaoshi Wu, Renrui Zhang, Dazhong Shen, Zhuofan
  Zong, Yu~Liu, and Hongsheng Li.
\newblock Comat: Aligning text-to-image diffusion model with image-to-text
  concept matching, 2024.
\newblock URL \url{https://arxiv.org/abs/2404.03653}.

\bibitem[Kang et~al.(2023)Kang, Galim, and Koo]{kang2023counting}
Wonjun Kang, Kevin Galim, and Hyung~Il Koo.
\newblock Counting guidance for high fidelity text-to-image synthesis, 2023.

\bibitem[Karthik et~al.(2023)Karthik, Roth, Mancini, and Akata]{karthik2023if}
Shyamgopal Karthik, Karsten Roth, Massimiliano Mancini, and Zeynep Akata.
\newblock If at first you don't succeed, try, try again: Faithful
  diffusion-based text-to-image generation by selection.
\newblock \emph{arXiv preprint arXiv:2305.13308}, 2023.

\bibitem[Kendall(1938)]{kendall38}
M.~G. Kendall.
\newblock {A Ner Measure of Rank Correlation}.
\newblock \emph{Biometrika}, 30\penalty0 (1-2):\penalty0 81--93, 06 1938.

\bibitem[Kim et~al.(2023)Kim, Lee, Kim, Ha, and Zhu]{kim2023}
Yunji Kim, Jiyoung Lee, Jin-Hwa Kim, Jung-Woo Ha, and Jun-Yan Zhu.
\newblock Dense text-to-image generation with attention modulation.
\newblock In \emph{ICCV}, 2023.

\bibitem[Kingma et~al.(2021)Kingma, Salimans, Poole, and Ho]{kingma2021}
Diederik~P Kingma, Tim Salimans, Ben Poole, and Jonathan Ho.
\newblock Variational diffusion models.
\newblock In A.~Beygelzimer, Y.~Dauphin, P.~Liang, and J.~Wortman Vaughan
  (eds.), \emph{Advances in Neural Information Processing Systems}, 2021.
\newblock URL \url{https://openreview.net/forum?id=2LdBqxc1Yv}.

\bibitem[Kirillov et~al.(2023)Kirillov, Mintun, Ravi, Mao, Rolland, Gustafson,
  Xiao, Whitehead, Berg, Lo, Dollar, and Girshick]{kirillov2023ICCV}
Alexander Kirillov, Eric Mintun, Nikhila Ravi, Hanzi Mao, Chloe Rolland, Laura
  Gustafson, Tete Xiao, Spencer Whitehead, Alexander~C. Berg, Wan-Yen Lo, Piotr
  Dollar, and Ross Girshick.
\newblock Segment anything.
\newblock In \emph{Proceedings of the IEEE/CVF International Conference on
  Computer Vision (ICCV)}, pp.\  4015--4026, October 2023.

\bibitem[Kong et~al.(2024)Kong, Liu, Li, Yogatama, and
  Steeg]{kong2024interpretable}
Xianghao Kong, Ollie Liu, Han Li, Dani Yogatama, and Greg~Ver Steeg.
\newblock Interpretable diffusion via information decomposition.
\newblock In \emph{The Twelfth International Conference on Learning
  Representations}, 2024.
\newblock URL \url{https://openreview.net/forum?id=X6tNkN6ate}.

\bibitem[Krojer et~al.(2023)Krojer, Poole-Dayan, Voleti, Pal, and
  Reddy]{krojer2023HNitm}
Benno Krojer, Elinor Poole-Dayan, Vikram Voleti, Christopher Pal, and Siva
  Reddy.
\newblock Are diffusion models vision-and-language reasoners?
\newblock \emph{arXiv preprint arXiv:2305.16397}, 2023.

\bibitem[Lee et~al.(2023)Lee, Liu, Ryu, Watkins, Du, Boutilier, Abbeel,
  Ghavamzadeh, and Gu]{lee2023aligning}
Kimin Lee, Hao Liu, Moonkyung Ryu, Olivia Watkins, Yuqing Du, Craig Boutilier,
  Pieter Abbeel, Mohammad Ghavamzadeh, and Shixiang~Shane Gu.
\newblock Aligning text-to-image models using human feedback, 2023.

\bibitem[Letizia \& Tonello(2022)Letizia and Tonello]{letizia2022copula}
Nunzio~A Letizia and Andrea~M Tonello.
\newblock Copula density neural estimation.
\newblock \emph{arXiv preprint arXiv:2211.15353}, 2022.

\bibitem[Li et~al.(2023{\natexlab{a}})Li, Li, Savarese, and Hoi]{li2023blip2}
Junnan Li, Dongxu Li, Silvio Savarese, and Steven Hoi.
\newblock Blip-2: bootstrapping language-image pre-training with frozen image
  encoders and large language models.
\newblock In \emph{Proceedings of the 40th International Conference on Machine
  Learning}, 2023{\natexlab{a}}.

\bibitem[Li et~al.(2023{\natexlab{b}})Li, Keuper, Zhang, and Khoreva]{li2023}
Yumeng Li, Margret Keuper, Dan Zhang, and Anna Khoreva.
\newblock Divide \& bind your attention for improved generative semantic
  nursing.
\newblock In \emph{34th British Machine Vision Conference 2023, {BMVC} 2023},
  2023{\natexlab{b}}.

\bibitem[Lin et~al.(2015)Lin, Maire, Belongie, Bourdev, Girshick, Hays, Perona,
  Ramanan, Zitnick, and Dollár]{lin2015microsoftcococommonobjects}
Tsung-Yi Lin, Michael Maire, Serge Belongie, Lubomir Bourdev, Ross Girshick,
  James Hays, Pietro Perona, Deva Ramanan, C.~Lawrence Zitnick, and Piotr
  Dollár.
\newblock Microsoft coco: Common objects in context, 2015.
\newblock URL \url{https://arxiv.org/abs/1405.0312}.

\bibitem[Liu et~al.(2022)Liu, Li, Du, Torralba, and Tenenbaum]{liu2022}
Nan Liu, Shuang Li, Yilun Du, Antonio Torralba, and Joshua~B. Tenenbaum.
\newblock Compositional visual generation with composable diffusion models.
\newblock In \emph{Computer Vision – ECCV 2022: 17th European Conference, Tel
  Aviv, Israel, October 23–27, 2022, Proceedings, Part XVII}, pp.\
  423–439, 2022.

\bibitem[Liu et~al.(2024)Liu, Wang, Yin, Molchanov, Wang, Cheng, and
  Chen]{liu2024dora}
Shih-Yang Liu, Chien-Yi Wang, Hongxu Yin, Pavlo Molchanov, Yu-Chiang~Frank
  Wang, Kwang-Ting Cheng, and Min-Hung Chen.
\newblock Dora: Weight-decomposed low-rank adaptation, 2024.

\bibitem[Liu \& Chilton(2022)Liu and Chilton]{liu2022design}
Vivian Liu and Lydia~B Chilton.
\newblock Design guidelines for prompt engineering text-to-image generative
  models.
\newblock In \emph{Proceedings of the 2022 CHI Conference on Human Factors in
  Computing Systems}, 2022.

\bibitem[Ma et~al.(2023)Ma, Liang, Chen, and Lu]{ma2023subjectdiffusionopen}
Jian Ma, Junhao Liang, Chen Chen, and Haonan Lu.
\newblock Subject-diffusion:open domain personalized text-to-image generation
  without test-time fine-tuning, 2023.

\bibitem[MacKay(2003)]{mackay2003information}
David~JC MacKay.
\newblock \emph{Information theory, inference and learning algorithms}.
\newblock Cambridge university press, 2003.

\bibitem[Mahajan et~al.(2023)Mahajan, Rahman, Yi, and
  Sigal]{mahajan2023prompting}
Shweta Mahajan, Tanzila Rahman, Kwang~Moo Yi, and Leonid Sigal.
\newblock Prompting hard or hardly prompting: Prompt inversion for
  text-to-image diffusion models, 2023.

\bibitem[Mañas et~al.(2024)Mañas, Astolfi, Hall, Ross, Urbanek, Williams,
  Agrawal, Romero-Soriano, and Drozdzal]{manas2024OPT2I}
Oscar Mañas, Pietro Astolfi, Melissa Hall, Candace Ross, Jack Urbanek, Adina
  Williams, Aishwarya Agrawal, Adriana Romero-Soriano, and Michal Drozdzal.
\newblock Improving text-to-image consistency via automatic prompt
  optimization, 2024.

\bibitem[McAllester \& Stratos(2020)McAllester and
  Stratos]{mcallester2020formal}
David McAllester and Karl Stratos.
\newblock Formal limitations on the measurement of mutual information.
\newblock In \emph{International Conference on Artificial Intelligence and
  Statistics}, 2020.

\bibitem[Meral et~al.(2024)Meral, Simsar, Tombari, and Yanardag]{meral2024}
Tuna Han~Salih Meral, Enis Simsar, Federico Tombari, and Pinar Yanardag.
\newblock Conform: Contrast is all you need for high-fidelity text-to-image
  diffusion models.
\newblock In \emph{Proceedings of the IEEE/CVF Conference on Computer Vision
  and Pattern Recognition}, 2024.

\bibitem[Nichol et~al.(2022)Nichol, Dhariwal, Ramesh, Shyam, Mishkin, McGrew,
  Sutskever, and Chen]{nichol2022glide}
Alex Nichol, Prafulla Dhariwal, Aditya Ramesh, Pranav Shyam, Pamela Mishkin,
  Bob McGrew, Ilya Sutskever, and Mark Chen.
\newblock Glide: Towards photorealistic image generation and editing with
  text-guided diffusion models, 2022.

\bibitem[Ogezi \& Shi(2024)Ogezi and Shi]{ogezi2024optimizing}
Michael Ogezi and Ning Shi.
\newblock Optimizing negative prompts for enhanced aesthetics and fidelity in
  text-to-image generation, 2024.

\bibitem[{\O}ksendal(2003)]{oksendal2003stochastic}
Bernt {\O}ksendal.
\newblock \emph{Stochastic differential equations}.
\newblock Springer, 2003.

\bibitem[Oord et~al.(2018)Oord, Li, and Vinyals]{oord2019representation}
Aaron van~den Oord, Yazhe Li, and Oriol Vinyals.
\newblock Representation learning with contrastive predictive coding.
\newblock \emph{Advances in neural information processing systems}, 2018.

\bibitem[OpenAI(2024)]{chatgpt}
OpenAI.
\newblock Chatgpt (gpt-4) [large language model], 2024.
\newblock URL \url{https://chat.openai.com/chat}.

\bibitem[Oquab et~al.(2024)Oquab, Darcet, Moutakanni, Vo, Szafraniec, Khalidov,
  Fernandez, HAZIZA, Massa, El-Nouby, Assran, Ballas, Galuba, Howes, Huang, Li,
  Misra, Rabbat, Sharma, Synnaeve, Xu, Jegou, Mairal, Labatut, Joulin, and
  Bojanowski]{dino}
Maxime Oquab, Timoth{\'e}e Darcet, Th{\'e}o Moutakanni, Huy~V. Vo, Marc
  Szafraniec, Vasil Khalidov, Pierre Fernandez, Daniel HAZIZA, Francisco Massa,
  Alaaeldin El-Nouby, Mido Assran, Nicolas Ballas, Wojciech Galuba, Russell
  Howes, Po-Yao Huang, Shang-Wen Li, Ishan Misra, Michael Rabbat, Vasu Sharma,
  Gabriel Synnaeve, Hu~Xu, Herve Jegou, Julien Mairal, Patrick Labatut, Armand
  Joulin, and Piotr Bojanowski.
\newblock {DINO}v2: Learning robust visual features without supervision.
\newblock \emph{Transactions on Machine Learning Research}, 2024.
\newblock ISSN 2835-8856.
\newblock URL \url{https://openreview.net/forum?id=a68SUt6zFt}.

\bibitem[Paninski(2003)]{paninski2003estimation}
Liam Paninski.
\newblock Estimation of entropy and mutual information.
\newblock \emph{Neural computation}, 15\penalty0 (6):\penalty0 1191--1253,
  2003.

\bibitem[Papamakarios et~al.(2017)Papamakarios, Pavlakou, and
  Murray]{papamakarios2017masked}
George Papamakarios, Theo Pavlakou, and Iain Murray.
\newblock Masked autoregressive flow for density estimation.
\newblock \emph{Advances in neural information processing systems}, 30, 2017.

\bibitem[Podell et~al.(2024)Podell, English, Lacey, Blattmann, Dockhorn,
  M{\"u}ller, Penna, and Rombach]{podell2024}
Dustin Podell, Zion English, Kyle Lacey, Andreas Blattmann, Tim Dockhorn, Jonas
  M{\"u}ller, Joe Penna, and Robin Rombach.
\newblock {SDXL}: Improving latent diffusion models for high-resolution image
  synthesis.
\newblock In \emph{The Twelfth International Conference on Learning
  Representations}, 2024.
\newblock URL \url{https://openreview.net/forum?id=di52zR8xgf}.

\bibitem[Radford et~al.(2021)Radford, Kim, Hallacy, Ramesh, Goh, Agarwal,
  Sastry, Askell, Mishkin, Clark, Krueger, and Sutskever]{radford2021}
Alec Radford, Jong~Wook Kim, Chris Hallacy, Aditya Ramesh, Gabriel Goh,
  Sandhini Agarwal, Girish Sastry, Amanda Askell, Pamela Mishkin, Jack Clark,
  Gretchen Krueger, and Ilya Sutskever.
\newblock Learning transferable visual models from natural language
  supervision.
\newblock In \emph{Proceedings of the 38th International Conference on Machine
  Learning}, volume 139, pp.\  8748--8763, 2021.

\bibitem[Ramesh et~al.(2022)Ramesh, Dhariwal, Nichol, Chu, and
  Chen]{ramesh2022hierarchical}
Aditya Ramesh, Prafulla Dhariwal, Alex Nichol, Casey Chu, and Mark Chen.
\newblock Hierarchical text-conditional image generation with clip latents,
  2022.

\bibitem[Rassin et~al.(2023)Rassin, Hirsch, Glickman, Ravfogel, Goldberg, and
  Chechik]{rassin2023}
Royi Rassin, Eran Hirsch, Daniel Glickman, Shauli Ravfogel, Yoav Goldberg, and
  Gal Chechik.
\newblock Linguistic binding in diffusion models: Enhancing attribute
  correspondence through attention map alignment.
\newblock In \emph{Thirty-seventh Conference on Neural Information Processing
  Systems}, 2023.
\newblock URL \url{https://openreview.net/forum?id=AOKU4nRw1W}.

\bibitem[Rhodes et~al.(2020)Rhodes, Xu, and Gutmann]{rhodes2020telescoping}
Benjamin Rhodes, Kai Xu, and Michael~U Gutmann.
\newblock Telescoping density-ratio estimation.
\newblock \emph{Advances in neural information processing systems}, 2020.

\bibitem[Rombach et~al.(2022)Rombach, Blattmann, Lorenz, Esser, and
  Ommer]{rombach2022}
Robin Rombach, Andreas Blattmann, Dominik Lorenz, Patrick Esser, and Bj\"orn
  Ommer.
\newblock High-resolution image synthesis with latent diffusion models.
\newblock In \emph{Proceedings of the IEEE/CVF Conference on Computer Vision
  and Pattern Recognition (CVPR)}, pp.\  10684--10695, June 2022.

\bibitem[Ronneberger et~al.(2015)Ronneberger, Fischer, and
  Brox]{ronneberger2015}
Olaf Ronneberger, Philipp Fischer, and Thomas Brox.
\newblock U-net: Convolutional networks for biomedical image segmentation.
\newblock In \emph{Medical Image Computing and Computer-Assisted Intervention
  -- MICCAI 2015}, pp.\  234--241, 2015.

\bibitem[Saharia et~al.(2022)Saharia, Chan, Saxena, Li, Whang, Denton,
  Ghasemipour, Gontijo~Lopes, Karagol~Ayan, Salimans, Ho, Fleet, and
  Norouzi]{saharia2022}
Chitwan Saharia, William Chan, Saurabh Saxena, Lala Li, Jay Whang, Emily~L
  Denton, Kamyar Ghasemipour, Raphael Gontijo~Lopes, Burcu Karagol~Ayan, Tim
  Salimans, Jonathan Ho, David~J Fleet, and Mohammad Norouzi.
\newblock Photorealistic text-to-image diffusion models with deep language
  understanding.
\newblock In \emph{Advances in Neural Information Processing Systems},
  volume~35, pp.\  36479--36494, 2022.

\bibitem[Salimans \& Ho(2022)Salimans and Ho]{salimans2022distillation}
Tim Salimans and Jonathan Ho.
\newblock Progressive distillation for fast sampling of diffusion models, 2022.

\bibitem[Samuel et~al.(2024)Samuel, Ben-Ari, Raviv, Darshan, and
  Chechik]{Samuel2023SeedSelect}
Dvir Samuel, Rami Ben-Ari, Simon Raviv, Nir Darshan, and Gal Chechik.
\newblock Generating images of rare concepts using pre-trained diffusion
  models.
\newblock In \emph{AAAI}, 2024.

\bibitem[Shannon(1948)]{shannon1948mathematical}
Claude~Elwood Shannon.
\newblock A mathematical theory of communication.
\newblock \emph{The Bell system technical journal}, 27\penalty0 (3):\penalty0
  379--423, 1948.

\bibitem[Shen et~al.(2024)Shen, Song, Xue, Wang, and Liu]{shen2024SCFG}
Dazhong Shen, Guanglu Song, Zeyue Xue, Fu-Yun Wang, and Yu~Liu.
\newblock Rethinking the spatial inconsistency in classifier-free diffusion
  guidance, 2024.

\bibitem[Sohl-Dickstein et~al.(2015)Sohl-Dickstein, Weiss, Maheswaranathan, and
  Ganguli]{sohl-dickstein15}
Jascha Sohl-Dickstein, Eric Weiss, Niru Maheswaranathan, and Surya Ganguli.
\newblock Deep unsupervised learning using nonequilibrium thermodynamics.
\newblock In \emph{Proceedings of the 32nd International Conference on Machine
  Learning}, volume~37, pp.\  2256--2265, 2015.

\bibitem[Song \& Ermon(2019{\natexlab{a}})Song and
  Ermon]{song2019understanding}
Jiaming Song and Stefano Ermon.
\newblock Understanding the limitations of variational mutual information
  estimators.
\newblock In \emph{International Conference on Learning Representations},
  2019{\natexlab{a}}.

\bibitem[Song \& Ermon(2019{\natexlab{b}})Song and Ermon]{song2019}
Yang Song and Stefano Ermon.
\newblock Generative modeling by estimating gradients of the data distribution.
\newblock In H.~Wallach, H.~Larochelle, A.~Beygelzimer, F.~d'~Alch\'{e}-Buc,
  E.~Fox, and R.~Garnett (eds.), \emph{Advances in Neural Information
  Processing Systems}, volume~32. Curran Associates, Inc., 2019{\natexlab{b}}.

\bibitem[Song \& Ermon(2020)Song and Ermon]{song2020}
Yang Song and Stefano Ermon.
\newblock Improved techniques for training score-based generative models.
\newblock In H.~Larochelle, M.~Ranzato, R.~Hadsell, M.F. Balcan, and H.~Lin
  (eds.), \emph{Advances in Neural Information Processing Systems}, volume~33,
  pp.\  12438--12448. Curran Associates, Inc., 2020.

\bibitem[Song et~al.(2021)Song, Sohl-Dickstein, Kingma, Kumar, Ermon, and
  Poole]{song2021a}
Yang Song, Jascha Sohl-Dickstein, Diederik~P Kingma, Abhishek Kumar, Stefano
  Ermon, and Ben Poole.
\newblock Score-based generative modeling through stochastic differential
  equations.
\newblock In \emph{International Conference on Learning Representations}, 2021.

\bibitem[Stratos(2019)]{stratos2018mutual}
Karl Stratos.
\newblock Mutual information maximization for simple and accurate
  part-of-speech induction.
\newblock In \emph{Proceedings of the 2019 Conference of the North American
  Chapter of the Association for Computational Linguistics: Human Language
  Technologies, Volume 1 (Long and Short Papers)}, 2019.

\bibitem[Sun et~al.(2023)Sun, Fu, Hu, Wang, Rassin, Juan, Alon, Herrmann, van
  Steenkiste, Krishna, and Rashtchian]{sun2023dreamsync}
Jiao Sun, Deqing Fu, Yushi Hu, Su~Wang, Royi Rassin, Da-Cheng Juan, Dana Alon,
  Charles Herrmann, Sjoerd van Steenkiste, Ranjay Krishna, and Cyrus
  Rashtchian.
\newblock Dreamsync: Aligning text-to-image generation with image understanding
  feedback, 2023.

\bibitem[Tang et~al.(2023)Tang, Liu, Pandey, Jiang, Yang, Kumar, Stenetorp,
  Lin, and Ture]{tang2023}
Raphael Tang, Linqing Liu, Akshat Pandey, Zhiying Jiang, Gefei Yang, Karun
  Kumar, Pontus Stenetorp, Jimmy Lin, and Ferhan Ture.
\newblock What the {DAAM}: Interpreting stable diffusion using cross attention.
\newblock In \emph{Proceedings of the 61st Annual Meeting of the Association
  for Computational Linguistics (Volume 1: Long Papers)}, pp.\  5644--5659,
  July 2023.

\bibitem[Wallace et~al.(2023)Wallace, Dang, Rafailov, Zhou, Lou, Purushwalkam,
  Ermon, Xiong, Joty, and Naik]{wallace2023diffusion}
Bram Wallace, Meihua Dang, Rafael Rafailov, Linqi Zhou, Aaron Lou, Senthil
  Purushwalkam, Stefano Ermon, Caiming Xiong, Shafiq Joty, and Nikhil Naik.
\newblock Diffusion model alignment using direct preference optimization, 2023.

\bibitem[Wang et~al.(2022)Wang, Montoya, Munechika, Yang, Hoover, and
  Chau]{diffusiondb}
Zijie~J. Wang, Evan Montoya, David Munechika, Haoyang Yang, Benjamin Hoover,
  and Duen~Horng Chau.
\newblock {{DiffusionDB}}: {{A}} large-scale prompt gallery dataset for
  text-to-image generative models.
\newblock \emph{arXiv:2210.14896 [cs]}, 2022.
\newblock URL \url{https://arxiv.org/abs/2210.14896}.

\bibitem[Wang et~al.(2023{\natexlab{a}})Wang, Montoya, Munechika, Yang, Hoover,
  and Chau]{wang2023}
Zijie~J. Wang, Evan Montoya, David Munechika, Haoyang Yang, Benjamin Hoover,
  and Duen~Horng Chau.
\newblock {D}iffusion{DB}: A large-scale prompt gallery dataset for
  text-to-image generative models.
\newblock In \emph{Proceedings of the 61st Annual Meeting of the Association
  for Computational Linguistics (Volume 1: Long Papers)}, 2023{\natexlab{a}}.

\bibitem[Wang et~al.(2023{\natexlab{b}})Wang, Sha, Ding, Wang, and
  Tu]{wang2023tokencompose}
Zirui Wang, Zhizhou Sha, Zheng Ding, Yilin Wang, and Zhuowen Tu.
\newblock Tokencompose: Grounding diffusion with token-level supervision,
  2023{\natexlab{b}}.

\bibitem[Witteveen \& Andrews(2022)Witteveen and
  Andrews]{witteveen2022investigating}
Sam Witteveen and Martin Andrews.
\newblock Investigating prompt engineering in diffusion models, 2022.

\bibitem[Wu et~al.(2023{\natexlab{a}})Wu, Liu, Zhao, Bui, Lin, Zhang, and
  Chang]{wu2023}
Qiucheng Wu, Yujian Liu, Handong Zhao, Trung Bui, Zhe Lin, Yang Zhang, and
  Shiyu Chang.
\newblock Harnessing the spatial-temporal attention of diffusion models for
  high-fidelity text-to-image synthesis.
\newblock In \emph{Proceedings of the IEEE/CVF International Conference on
  Computer Vision (ICCV)}, pp.\  7766--7776, October 2023{\natexlab{a}}.

\bibitem[Wu et~al.(2023{\natexlab{b}})Wu, Hao, Sun, Chen, Zhu, Zhao, and
  Li]{wu2023human}
Xiaoshi Wu, Yiming Hao, Keqiang Sun, Yixiong Chen, Feng Zhu, Rui Zhao, and
  Hongsheng Li.
\newblock Human preference score v2: A solid benchmark for evaluating human
  preferences of text-to-image synthesis, 2023{\natexlab{b}}.

\bibitem[Wu et~al.(2023{\natexlab{c}})Wu, Sun, Zhu, Zhao, and Li]{wu2023HPS}
Xiaoshi Wu, Keqiang Sun, Feng Zhu, Rui Zhao, and Hongsheng Li.
\newblock Human preference score: Better aligning text-to-image models with
  human preference, 2023{\natexlab{c}}.

\bibitem[Wu et~al.(2023{\natexlab{d}})Wu, Sun, Zhu, Zhao, and Li]{wu2023better}
Xiaoshi Wu, Keqiang Sun, Feng Zhu, Rui Zhao, and Hongsheng Li.
\newblock Human preference score: Better aligning text-to-image models with
  human preference, 2023{\natexlab{d}}.

\bibitem[Wu et~al.(2024)Wu, Yu, Huang, Russakovsky, and
  Arora]{wu2024conceptmixcompositionalimagegeneration}
Xindi Wu, Dingli Yu, Yangsibo Huang, Olga Russakovsky, and Sanjeev Arora.
\newblock Conceptmix: A compositional image generation benchmark with
  controllable difficulty, 2024.
\newblock URL \url{https://arxiv.org/abs/2408.14339}.

\bibitem[Xu et~al.(2023)Xu, Liu, Wu, Tong, Li, Ding, Tang, and
  Dong]{xu2023imagereward}
Jiazheng Xu, Xiao Liu, Yuchen Wu, Yuxuan Tong, Qinkai Li, Ming Ding, Jie Tang,
  and Yuxiao Dong.
\newblock Imagereward: Learning and evaluating human preferences for
  text-to-image generation.
\newblock In \emph{Thirty-seventh Conference on Neural Information Processing
  Systems}, 2023.
\newblock URL \url{https://openreview.net/forum?id=JVzeOYEx6d}.

\bibitem[Yuan et~al.(2024)Yuan, Chen, Ji, and Gu]{yuan2024selfplay}
Huizhuo Yuan, Zixiang Chen, Kaixuan Ji, and Quanquan Gu.
\newblock Self-play fine-tuning of diffusion models for text-to-image
  generation, 2024.

\bibitem[Zhang et~al.(2024{\natexlab{a}})Zhang, Yang, Cai, Yu, Wang, Xie, Tian,
  Xu, Tang, Yang, and Cui]{zhang2024realcompo}
Xinchen Zhang, Ling Yang, Yaqi Cai, Zhaochen Yu, Kai-Ni Wang, Jiake Xie,
  Ye~Tian, Minkai Xu, Yong Tang, Yujiu Yang, and Bin Cui.
\newblock Realcompo: Balancing realism and compositionality improves
  text-to-image diffusion models, 2024{\natexlab{a}}.
\newblock URL \url{https://arxiv.org/abs/2402.12908}.

\bibitem[Zhang et~al.(2024{\natexlab{b}})Zhang, Zhang, Nie, Li, Chen, Hao,
  Zhang, Liu, and Li]{zhang2024spdiffusion}
Yang Zhang, Rui Zhang, Xuecheng Nie, Haochen Li, Jikun Chen, Yifan Hao, Xin
  Zhang, Luoqi Liu, and Ling Li.
\newblock Spdiffusion: Semantic protection diffusion for multi-concept
  text-to-image generation, 2024{\natexlab{b}}.
\newblock URL \url{https://arxiv.org/abs/2409.01327}.

\bibitem[Zhao et~al.(2024)Zhao, Yang, Zhang, Shao, Zhang, Qiao, Luo, and
  Ji]{Zhao_2024_CVPR}
Lirui Zhao, Yue Yang, Kaipeng Zhang, Wenqi Shao, Yuxin Zhang, Yu~Qiao, Ping
  Luo, and Rongrong Ji.
\newblock Diffagent: Fast and accurate text-to-image api selection with large
  language model.
\newblock In \emph{Proceedings of the IEEE/CVF Conference on Computer Vision
  and Pattern Recognition (CVPR)}, pp.\  6390--6399, June 2024.

\bibitem[Zhao et~al.(2018)Zhao, Song, and Ermon]{zhao2018information}
Shengjia Zhao, Jiaming Song, and Stefano Ermon.
\newblock A lagrangian perspective on latent variable generative models.
\newblock In \emph{Proc. 34th Conference on Uncertainty in Artificial
  Intelligence}, 2018.

\bibitem[Zhou et~al.(2022)Zhou, Koltun, and Kr{\"a}henb{\"u}hl]{zhou2021simple}
Xingyi Zhou, Vladlen Koltun, and Philipp Kr{\"a}henb{\"u}hl.
\newblock Simple multi-dataset detection.
\newblock In \emph{CVPR}, 2022.

\bibitem[Zhu et~al.(2024)Zhu, Chen, Shen, Li, and Elhoseiny]{zhu2024minigpt}
Deyao Zhu, Jun Chen, Xiaoqian Shen, Xiang Li, and Mohamed Elhoseiny.
\newblock Mini{GPT}-4: Enhancing vision-language understanding with advanced
  large language models.
\newblock In \emph{The Twelfth International Conference on Learning
  Representations}, 2024.
\newblock URL \url{https://openreview.net/forum?id=1tZbq88f27}.

\end{thebibliography}
}

%%%%%%%%%%%%%%%%%%%%%%%%%%%%%%%%%%%%%%%%%%%%%%%%%%%%%%%%%%%
\newpage
\appendix
\appendixpage
{
\fontsize{8.7}{8.7}\selectfont
\startcontents[sections]
\printcontents[sections]{l}{1}{\setcounter{tocdepth}{2}}
}
\newpage
\section[
Details on \gls{MI} estimation
\newline
{\fontsize{8}{8}\selectfont
\textcolor{blue!50}{
\it Proof of \Cref{eq:mi} $\cdot$ Referenced in \Cref{sec:preliminaries}
}}
]{Details on \gls{MI} estimation}\label{app:minde}
In this Section, we provide the proof for \Cref{eq:mi}.
We start by recalling the definition of the forward and backward processes for a discrete-time diffusion model. For the forward process, we use the following Markov chain
\begin{equation*}
    q(\mbz_{0:T},\mbp)=q(\mbz_0,\mbp)\prod\limits_{t=1}^T q(\mbz_t\g \mbz_{t-1}),\quad q(\mbz_t\g \mbz_{t-1})=\mathcal{N}(\mbz_t; \sqrt{1-\beta_t} \mbz_{t-1},\beta_t I)
\end{equation*}

The backward process (with or without a conditioning signal $\mbp$) evolves according to
\begin{equation*}
    p_{\mbtheta}(\mbz_{0:T})=p(\mbz_T)\prod\limits_{t=1}^T p_{\mbtheta}(\mbz_{t-1}\g \mbz_t),\quad p_{\mbtheta}(\mbz_{0:T} \g \mbp)=p(\mbz_T)\prod\limits_{t=1}^T p_{\mbtheta}(\mbz_{t-1}\g \mbz_t, \mbp)
\end{equation*}

where $p_{\mbtheta}(\mbz_{t-1}\g \mbz_t)=\mathcal{N}(\mbz_{t-1};\mu_\mbtheta(\mbz_t),\beta_t I)$, with $ \mu_\mbtheta(\mbz_t)=\frac{1}{\sqrt{\alpha_t}}
                \left(
                    \mbz_t - \frac{\beta_t}{\sqrt{1-\bar{\alpha}_t}} {\mbepsilon}_\mbtheta(\mbz_t, t) 
                \right)$. Similar expressions can be obtained for the conditional version. 

Our goal here is to show that the following equality holds
\begin{equation*}
    \E_{\mbp}[\KL{q(\mbz_0\g \mbp)}{q(\mbz_0)}]=\E_{\mbz,\mbp}[\textsc{I}(\mbz,\mbp)],
\end{equation*}
which is the condition that $\textsc{I}(\mbz,\mbp)$ of \Cref{eq:mi} should satisfy to be a valid point-wise \gls{MI} estimator. In particular, we will show that
\begin{equation*}
      \E_{\mbp}[\KL{q(\mbz_0\g \mbp)}{q(\mbz_0)}]= \E_{t,\mbp,\mbz,\mbepsilon}
    \left[
    \kappa_t
    ||
        \mbepsilon_\mbtheta(\mbz_t, \mbp, t) - \mbepsilon_\mbtheta(\mbz_t, \emptyset, t)
    ||^2
    \right],\,\,\, \kappa_t = \frac{\beta_t T}{2\alpha_t (1- \bar{\alpha}_t)}.
\end{equation*}

To simplify our proof strategy, we consider the ideal case of perfect training, i.e., $p_{\mbtheta}(\mbz_{0:T},\mbp)=q(\mbz_{0:T},\mbp)$. Moreover, since $q(\mbz_{t}\g \mbz_{t-1},\mbp)=q(\mbz_{t}\g \mbz_{t-1})$, we can rewrite the $\KL{q(\mbz_0\g \mbp)}{q(\mbz_0)}$ term as follows

\vspace{-5pt}
\begin{flalign*}
    &\KL{q(\mbz_0\g \mbp)}{q(\mbz_0)}=\KL{q(\mbz_{0:T}\g \mbp)}{q(\mbz_{0:T})}=\KL{p_\mbtheta(\mbz_{0:T}\g \mbp)}{p_{\mbtheta}(\mbz_{0:T})}=\\&
    \int p_\mbtheta(\mbz_{0:T}\g \mbp) \log\frac{p_\mbtheta(\mbz_{0:T}\g \mbp)}{p_{\mbtheta}(\mbz_{0:T})}\dd \mbz_{0:T}=\int p_\mbtheta(\mbz_{0:T}\g \mbp) \sum\limits_{t=1}^T\log\frac{p_\mbtheta(\mbz_{t-1}\g \mbz_t, \mbp)}{p_{\mbtheta}(\mbz_{t-1}\g \mbz_t)}\dd \mbz_{0:T}=\\&
    \sum\limits_{t=1}^T \int p_\mbtheta(\mbz_{0:t-2,t:T}\g \mbp) \left(\int p_\mbtheta(\mbz_{t-1}\g \mbz_t, \mbp) \log\frac{p_\mbtheta(\mbz_{t-1}\g \mbz_t, \mbp)}{p_{\mbtheta}(\mbz_{t-1}\g \mbz_t)}\dd \mbz_{t-1}\right) \dd \mbz_{0:t-2,t:T}=\\&
    \sum\limits_{t=1}^T \int p_\mbtheta(\mbz_{t}\g \mbp) \KL{p_\mbtheta(\mbz_{t-1}\g \mbz_t, \mbp)}{p_\mbtheta(\mbz_{t-1}\g \mbz_t)} \dd \mbz_{t}=\\&
    \sum\limits_{t=1}^T \frac{1}{2\beta_t}\int p_\mbtheta(\mbz_{t}\g \mbp) \norm{\mu_{\mbtheta}(\mbz_t)-\mu_{\mbtheta}(\mbz_t,\mbp)}^2 \dd \mbz_{t}=\\&
    \sum\limits_{t=1}^T \frac{1}{2\beta_t}\frac{\beta_t^2}{\alpha_t (1-\bar{\alpha}_t)}\int p_\mbtheta(\mbz_{t}\g \mbp) \norm{\epsilon_{\mbtheta}(\mbz_t)-\epsilon_{\mbtheta}(\mbz_t,\mbp)}^2 \dd \mbz_{t}=\\&
    \E_{t,\mbz_t}\left[\kappa_t\norm{\epsilon_{\mbtheta}(\mbz_t,\emptyset, t)-\epsilon_{\mbtheta}(\mbz_t,\mbp,t)}^2\right]=\\&
    \E_{t,\mbz,\epsilon}\left[\kappa_t\norm{\epsilon_{\mbtheta}(\mbz_t,\emptyset, t)-\epsilon_{\mbtheta}(\mbz_t,\mbp,t)}^2\right],\quad \kappa_t = \frac{\beta_t T}{2\alpha_t (1- \bar{\alpha}_t)}
\end{flalign*}

which allows to prove that the quantity in \Cref{eq:mi} is indeed a valid point-wise \gls{MI} estimator.%, where it is useful to remember that $\E_{t,\mbz_t}=\E_{t,\mbz,\epsilon}$.

\section{Details on user study}\label{app:user_study}
Our user studies are based on small focus groups participants only, with (lightly) guided discussions led by a moderator. In those campaigns we elicited feedback from users regarding the comparison of different alignment metrics, aiming to understand if MI is a \emph{plausible} choice. Although launching large-scale survey campaigns would be desirable, this would require a completely different organization and implementation with respect to what what we adopted for this work.

\noindent \textbf{The survey web app.}
Beside \emph{punctually} comparing alignment metrics~\Cref{sec:method} and methods~\Cref{sec:results}, we designed a web app to collect \emph{subjective} feedback, in the form of mini surveys, from real users. Each survey is composed of multiple tests, each showing
a prompt and a set of images generated from it. 
Under the hood, the web app corresponds to a jupyter notebook with \texttt{ipywidgets}\footnote{\url{https://ipywidgets.readthedocs.io/en/stable/}} for UI controls, rendered via the \texttt{voila}\footnote{\url{https://voila.readthedocs.io/en/stable/using.html}} framework and deployed live via a docker-ized HuggingFace space. Via the web app we run campaigns to \emph{compare alignment metrics} and to \emph{compare alignment methods}.

\begin{figure}[t]
\includegraphics[width=\linewidth]{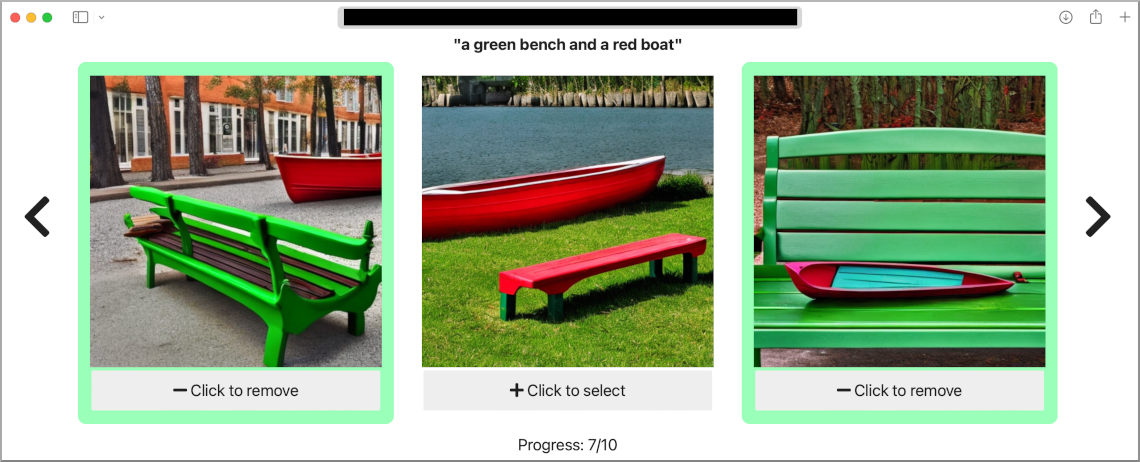}
\caption{Web app screenshot example of the alignment metric comparison survey.}
\label{fig:survey-example}
\end{figure}

\Cref{fig:survey-example} shows an example screenshot of the alignment metric comparison survey~\Cref{sec:method}. As from the example, users are free to select from 0 to up to 3 images for each prompt. However, to stress users subjectivity, we intentionally did not provide guidelines on how to handle ``odd'' cases (e.g., if the prompt asks for a picture of ``an apple'', but the picture show more than one apple). Last, each survey is saved as a separate CSV with the timestamp of its creation which also serves as unique identifier of the survey, i.e., neither a user identifier nor cookies are required by the web app logic, so users' privacy and anonymity is preserved.

\begin{table}[h]
\centering
\caption{User study about comparing alignment metrics.
\label{tab:survey}
}
\fontsize{8}{8}\selectfont
\begin{tabular}{
@{}
c
@{$\:\:\:$}c
@{$\:\:\:$}c
@{$\:\:\:$}r
@{$\:\:\:$}r
@{$\:\:\:$}r
@{$\:\:\:$}r
@{}}
\toprule
\multicolumn{3}{c}{\bf Metric}&
\multicolumn{4}{c}{\bf Campaign answers (\%)}
\\
\cmidrule(r){1-3}
\cmidrule(r){4-7}
\gls{MI} &
\acrshort{BLIP} &
\gls{HPS} &
Academic users &
Random users &
Students &
\it avg
\\
\cmidrule(r){1-3}
\cmidrule(r){4-6}
\cmidrule(r){7-7}
 \Circle  &  \Circle    &  \Circle      &     14.7   &   16.9   &   25.0   & \it 18.9\\
 \Circle  &  \Circle    &  \CIRCLE      &      1.8   &   14.0   &    2.7   & \it  6.2\\
 \Circle  &  \CIRCLE    &  \Circle      &     10.4   &   22.0   &    4.1   & \it 12.2\\
 \Circle  &  \CIRCLE    &  \CIRCLE      &      4.0   &    7.4   &    3.6   & \it  5.0\\
 \CIRCLE  &  \Circle    &  \Circle      &      4.0   &   10.6   &    0.9   & \it  5.2\\
 \CIRCLE  &  \Circle    &  \CIRCLE      &      6.0   &    2.5   &    2.3   & \it  3.6\\
 \CIRCLE  &  \CIRCLE    &  \Circle      &     18.7   &   10.6   &   16.8   & \it 15.4\\
 \CIRCLE  &  \CIRCLE    &  \CIRCLE      &     40.4   &   16.0   &   44.6   & \it 33.7\\
\cmidrule(r){1-3}
\cmidrule(r){4-6}
\cmidrule(r){7-7}
\CIRCLE      &  \textcolor{lightgray}{\RIGHTcircle} &  \textcolor{lightgray}{\RIGHTcircle} &     69.1   &   39.7   &   64.6   & \it 57.8\\
\textcolor{lightgray}{\RIGHTcircle} &  \CIRCLE      &  \textcolor{lightgray}{\RIGHTcircle} &     73.5   &   56.0   &   69.1   & \it 66.2\\
\textcolor{lightgray}{\RIGHTcircle} &  \textcolor{lightgray}{\RIGHTcircle} & \CIRCLE       &     52.2   &   39.9   &   53.2   & \it 48.2\\
\bottomrule
\end{tabular}
\\
\CIRCLE (selected) $\quad$ \Circle (not selected) $\quad$ \textcolor{lightgray}{\RIGHTcircle} (indifferent to the selection) 
\end{table}

\subsection[
Comparing alignment metrics
\newline
{\fontsize{8}{8}\selectfont
\textcolor{blue!50}{
\it Preliminary analysis to understand if \acrshort{MI} is a meaningful alignment signal 
$\cdot$ Referenced in \Cref{sec:method}
}}
]{Comparing alignment metrics\label{app:survey-metrics}}

In the first surveys campaign we aimed to understand how users perceive images pre-selected by \acrshort{BLIP}, \gls{HPS} and \gls{MI}. Specifically, we run surveys composed of 10 tests, each showing a prompt and the related best image among 50 generations (using \acrshort{SDBASE}) as ranked according to each metric separately~(\Cref{sec:method}). Each of the 10 prompts is randomly selected from a pool of 700 prompts
for the T2I-Combench color category, and at each test
the order in which the 3 pictures is shown is also randomized.

We run surveys across three user groups: \emph{Academic users} (5 members) are representative of highly informed and tech savvy users, who are familiar with how generative models work; \emph{Random users} (25 members) are representative of illiterate users who are not familiar with computer-based image generation; \emph{Students} (16 members) are representative of masters' level students who are familiar with image generation tools, and who have attended introductory-level machine learning classes.

Overall, we collected 102 surveys (45, 35 and 22 surveys across 3 days for Academics, Random users and Students respectively) which we detail in 
\Cref{tab:survey}. The top part of the table breaks down all possible answers combinations. The results, although with some differences between user groups, clearly highlight that the three alignment metrics we consider in this work are roughly equivalent, with \gls{MI} and \acrshort{BLIP} being preferred over \gls{HPS}. For the Academics and Students groups, all the three images are considered sufficiently aligned with the prompt in almost half of the cases ($40.4\%$ and $44.6\%$ respectively). Interestingly, random users select only one of the three images about 10$\times$ much more frequently than the other two groups (on average 14.2\% for real users while 5.4\% and 3\% for Academics and Students respectively). We hypothesise that being previously exposed (or not) to the technical problems of image generation from the alignment perspective, or simply being literate (or not) about machine learning can influence the selection among the three pictures. 

The bottom part of the table summarizes the answers for each individual metric. 
Despite the general preference for \acrshort{BLIP}, the results corroborate once more that MI provides a meaningful alignment signal (possibly compatible with aesthetics too).

Finally, we recall that our goal in this section is to study whether \textbf{\gls{MI} is a \emph{plausible} alignment measure}, rather than electing the ``best'' alignment metric. Indeed, this analysis does not indicate the final performance of alignment methods, which instead we report in \Cref{tab:results_blip}.

\begin{table}[t]
\caption{Users study comparing alignment methods. \textbf{Bold} shows best performance; \DIA shows the best method per-family.
\label{tab:survey-methods}
}
\centering
\fontsize{8}{8}\selectfont
\begin{tabular}{
    @{$\:\:\:$}
    c
    @{$\:\:\:$}
    r
    @{$\:\:\:\:\:$}
    r
    @{$\:\:\:\:\:$}
    r
    @{$\:\:\:\:\:$}
    r
    @{$\:\:\:\:\:$}
    r
    @{$\:\:\:\:\:$}
    r
    @{$\:\:\:\:\:$}
    r
}

% c r r r r r r r}
\toprule
Alignment & & \multicolumn{6}{c}{Category (\%)}  
\\
\cmidrule(r){3-8}
Methodology & Model   & Color  & Shape   & Texture  & 2D-Spatial  & Non-spatial  & Complex
\\
\cmidrule(r){1-2}
\cmidrule{3-8}
\it none   
    & \gls{SD}-2.1-base   & 29.76                       & 11.90                      & 40.48             & 35.71                      & 66.67                       & 29.76 \\
\cmidrule(r){1-2}
\cmidrule{3-8}
\multirow{3}{*}{Inference-time}
    & \gls{AE}            & \DIA31.95                   & \DIA15.48                  & \DIA52.38         & 32.14                      & 65.48                       & 30.95 \\
    & \gls{SDG}           & 26.19                       & \DIA15.48                  & 38.10             & 38.10                      & 61.90                       & 29.76 \\
    & \gls{SCG}           & 20.24                       & 11.90                      & 33.33             & \DIA40.48                  & \DIA69.05                   & \DIA39.29 \\
\cmidrule(r){1-2}
\cmidrule{3-8}
\multirow{4}{*}{Fine-tuning}    
    & \gls{DPOK}          & 23.81                       & 16.67                       & \DIA47.62        & 34.52                      & \DIA70.24                   & \DIA38.10 \\
    & \gls{GORS}          & \DIA34.52                   & 14.29                       & 48.81            & \DIA36.90                  & 65.48                       & 30.95 \\
    & \gls{HN}            & 23.81                       & \DIA19.05                   & 30.95            & 20.24                      & 47.62                       & 23.81 \\
\cmidrule(r){2-2}
\cmidrule{3-8}
    & \acrshort{MITUNE}   & \bf46.43                    & \bf25.01                    & \bf{53.19}       & \bf45.24                   & \bf73.81                    & \bf46.43 \\
\bottomrule
\end{tabular}
\end{table}

\subsection[
Comparing alignment methods
\newline
{\fontsize{8}{8}\selectfont
\textcolor{blue!50}{
\it Actual benchmark of \acrshort{MITUNE} against alternative methods
$\cdot$ Referenced in \Cref{sec:benchmark-and-metrics}
}}
]{Comparing alignment methods\label{app:survey-methods}}
In this second survey campaign we aimed to understand how users perceive images generated by the 8 methods we considered
in our study, i.e., ``vanilla'' \gls{SD}, \gls{AE}~\citep{chefer2023attendandexcite}, \gls{SDG}~\citep{feng2023Structured} and \gls{SCG}~\citep{salimans2022distillation} \gls{DPOK}~\citep{fan2023dpok}, \gls{GORS}~\citep{huang2023t2icompbench}, \gls{HN}~\citep{krojer2023HNitm} and
our method \gls{MITUNE} (when used with a single round of fine-tuning).

To do so, we run surveys composed of 2 tests %rounds 
for each T2I-Combench category (12 rounds in total).
Each test shows a prompt and the 8 pictures generated using a different method.
For each category, we randomly selected 100 prompts from T2I-Combench test set to pre-generate the pictures.
At run time, the web app randomly selects 2 prompts for each category, and also randomly selects 
images from the related pool. Last, it randomly arranges both the tests 
(so that categories 
are shuffled) and the methods (so that pictures of a method are not visualized in the same position in the visualized grid).

\Cref{tab:survey-methods} collects the results of a campaign with 42 surveys.
Specifically, the table shows the percentage of answers where the picture of a given method was selected (no matter if other methods were
also selected) -- 
theses results are integrated in right side of~\Cref{tab:results_blip} and are duplicated here for completeness.

\section[
Experimental protocol details
\newline
{\fontsize{8}{8}\selectfont
\textcolor{blue!50}{
\it List of parameters used and computation costs considerations. $\cdot$ Referenced in \Cref{sec:mitune-finetuning}
}}
]{Experimental protocol details}\label{app:exp_protocol}
We report in \Cref{tab:hyperparameters} all the hyper-parameters we used for our experiments.

\begin{table}[H]
\small
\centering
\caption{Training hyperparameters.}
\fontsize{7}{7}\selectfont
\label{tab:hyperparameters}
\begin{tabular}{
    @{$\:\:\:\:\:$} 
    l
    @{$\:\:\:\:\:$}
    r
   % @{$\:\:\:\:\:\:$}
   % r
}
\toprule
{\bf Name}  & {\bf Value}  \\ %  &  {\bf Candidates}  
\midrule
Trainable model & \acrshort{UNET} \\
\midrule
Trainable timesteps & $t\sim U(500, 1000)$ \\
\midrule
%Regularizer formula   &     $-{||\epsilon_{\text {pre}} - \epsilon||}_2^2$    \\
%Regularizer weight $\lambda$   &    $0.1$     \\
%\midrule
PEFT & DoRA~\citep{liu2024dora} \\ % parameter-efficient finetuning
Rank   &    $32$     \\
$\alpha$    &    $32$       \\  % We then scale $\Delta W x$ by $\frac{\alpha}{r}$,
\midrule
Learning rate (LR)  &  $1e-4$      \\ %  1e-4 5e-5
Gradient norm clipping    &     $1.0$    \\
\midrule
LR scheduler   &    Constant      \\
LR warmup steps  &      $0$   \\
\midrule
Optimizer  &      AdamW  \\
AdamW - $\beta_1$   &  $0.9$      \\
AdamW - $\beta_2$   &     $0.999$        \\
AdamW - weight decay & $1e-2$   \\
AdamW - $\epsilon$  & $1e-8$   \\
\midrule
Resolution & $512 \times 512$         \\
Classifier-free guidance scale &  7.5         \\
Denoising steps & $50$ \\
\midrule
Batch size  & 400 \\
Training iterations & 300 \\
\midrule
GPUs for Training & 1 $\times$ NVIDIA A100       \\
% Training Time    &          \\
\bottomrule
\end{tabular}
\end{table}

Next, we provide additional details on the computational cost of \gls{MITUNE}. In our approach, there are two distinct phases that require computational effort:

\noindent \textbf{Generation}: The first is the construction of the fine-tuning set $\mathcal{S}$ based on point-wise \gls{MI}. As a reminder, for this phase, we use a pre-trained \gls{SD} model (namely \acrshort{SDBASE} at a resolution $512 \times 512$) and, given a prompt, conditionally generate 50 images, while at the same time computing point-wise \gls{MI} between the prompt and each image. This is done for all the prompts in the set $\mathcal{P}$. Specifically for T2I-Combench, each category training set has 700 prompts, and for each prompt we generate 50 images from which we select the one with highest MI. The generation of the 700$\times$50 fine-tuning set  requires roughly \emph{24 hours, i.e., about 2min per-prompt} on a single A100-80GB GPU -- the 50 images are generated together (as they roughly require 50GB of the 80GB available VRAM), while each prompt is processed sequentially.

\noindent\textbf{Fine-Tuning}: The second is the parameter efficient fine-tuning of the pre-trained model. Using the configuration discussed above, \gls{MITUNE} requires \emph{8 hours} when using a single A100-80GB GPU.

Note that ($i$) there is no overhead at image generation time: once a pre-trained model has been fine-tuned with \gls{MITUNE}, conditional sampling takes the same amount of time of ``vanilla'' \gls{SD} and ($ii$) while we report computational costs considering a single GPU, this is a extreme scenario and the time to process the workloads scales down (almost linearly) with the number of GPUs used according to our observations.

\section[
\acrshort{HPS} scores range
\newline
{\fontsize{8}{8}\selectfont
\textcolor{blue!50}{
\it Discussion about the natural small values range provided by HPS $\cdot$ Referenced in \Cref{sec:results}
}}
]{\acrshort{HPS} scores range}\label{app:hps_range}
\citet{wu2023} report a detailed benchmark of their metrics across 20+ models in the 
\acrshort{HPS}-v2 GitHub repository \texttt{\hyperlink{https://github.com/tgxs002/HPSv2}{https://github.com/tgxs002/HPSv2}}.
These details are hidden by default when loading the repository home page
and need to be explicitly ``opened'' expanding collapsed menus (e.g., $\blacktriangleright$ \texttt{v2 benchmark}).
To ease discussion, in \Cref{tab:appendix_hps_reference} we report an extract
of these benchmarks focusing on StableDiffusion as other models
are out scope for our study.

\begin{table}[H]
\centering
\caption{
HPS benchmark across multiple Stable Diffusion models extracted for HPS-v2 GitHub repo.
\label{tab:appendix_hps_reference}
}
\footnotesize
\begin{tabular}{
    @{}
    c
    @{$\:\:\:$}l 
    @{$\:\:\:$}c
    @{$\:\:\:$}c 
    @{$\:\:\:$}c 
    @{$\:\:\:$}c 
    @{$\:\:\:$}c
    @{}
}
\toprule
    Benchmark & 
    Model & 
    Animation & 
    Concept-Art &
    Painting &
    Photo &
    (\textit{avg}) 
\\
\midrule
\multirow{4}{*}{v2}
& SDXL Refiner (0.9)    &28.45  &27.66  &27.67  &27.46 	&(\textit{27.80}) \\
& SDXL Base (0.9) 	    &28.42 	&27.63 	&27.60 	&27.29 	&(\textit{27.73}) \\
\cmidrule{2-7}
& SD (2.0) 	            &27.48 	&26.89 	&26.86 	&27.46 	&(\textit{27.17}) \\
& SD (1.4) 	            &27.26 	&26.61 	&26.66 	&27.27 	&(\textit{26.95}) \\
\midrule
\multirow{4}{*}{v2.1}
&SDXL Refiner (0.9) 	&33.26 	&32.07 	&31.63 	&28.38 	&(\textit{31.34}) \\
&SDXL Base (0.9) 	    &32.84  &31.36 	&30.86 	&27.48 	&(\textit{30.63}) \\
\cmidrule{2-7}
&SD (2.0) 	            &27.09 	&26.02 	&25.68 	&26.73 	&(\textit{26.38}) \\
&SD (1.4) 	            &26.03 	&24.87 	&24.80 	&25.70 	&(\textit{25.35}) \\
\bottomrule
\end{tabular}
\end{table}

Results refer to two benchmark and are visually split between \acrshort{SD}
and \acrshort{SDXL}. The columns Animation, Concept-Art, Painting and Photo
are different images style, while (\textit{avg}) reflects average by row.

Both versions of the benchmark present similar takeaways which 
we can summarize in two main observations. Specifically, ($i$) 
different versions of the same model present $<0.5$ differences
and ($ii$) \acrshort{SDXL} outperforms \acrshort{SD} of
about +1 point -- the variation of \acrshort{HPS} scores is 
extremely contained even if these models are different generations apart.

Our \acrshort{HPS} scores in \Cref{tab:results_blip} present similar properties, but other literature
(e.g., Table 2 in \cite{Zhao_2024_CVPR}) present similar evidence.

\section{Additional results and ablations}\label{app:full_results}
\subsection[
Ablation: Fine-tuning set selection strategies
\newline
\fontsize{8}{8}\selectfont
{
\it
\textcolor{blue!50}{
Discussing alternative strategies to \acrshort{MI} for composing the fine-tuning set $\cdot$ Related to \Cref{tab:results_finetuning_composition}
}
}
]{Ablation: Fine-tuning set selection strategies}\label{app:abl_ft_set_strategies}

\noindent\textbf{Fine-tuning set selection strategy.}
It is important to stress that creating a fine-tuning dataset using the very same metric used for the final evaluation can artificially introduce a bias as stated in \citep{huang2023t2icompbench}: “calculating the rewards for GORS with the automatic evaluation metrics can lead to biased results”.

The selection strategy to compose the fine-tuning dataset is directly related to alignment scores and different fine-tuning methods opt for different choices. Specifically: HN-ITM uses an ad-hoc dataset with real positive and negative pairs;
GORS uses a synthetic dataset with no selection, but the fine-tuning loss of each sample is weighted by \acrshort{BLIP}
DPOK synthesizes new images at each training iteration since it is an online RL fine-tuning approach, and uses a pre-trained human preference model for reward.
\Cref{tab:results_blip}, in the main paper, shows alternative fine-tuning strategies based on synthetic generated data using a variety of selection scores: GORS and DPOK are the closest methods to MI-TUNE from this point of view, yet generally underperforming compared to it. 

%%%%%%%%%%%%%%%%%%%%%%%%%%%%
%%%%%%%%%%%%%%%%%%%%%%%%%%%%
%%%%%%%%%%%%%%%%%%%%%%%%%%%%
\begin{minipage}[t]{0.4\textwidth}
% \begin{table}[htbp]
\vspace{-7pt}
\centering
\captionof{table}{
FT set selection.
\label{tab:appendix_results_finetuning_composition}
}
\vspace{-8pt}
\scalebox{0.88}{
\begin{tabular}{
@{}
l
@{$\:\;$}c
@{$\:$}c
%%%%%
@{$\:\:$}c
@{$\:$}c
@{}}
\toprule
    & \multicolumn{2}{c}{\bf\acrshort{BLIP}} 
    & \multicolumn{2}{c}{\bf\acrshort{HPS}}
\\
\cmidrule(r){2-3}
\cmidrule{4-5}
Strategy 
    & \TINY{Color} & \TINY{Shape} 
    & \TINY{Color} & \TINY{Shape}
\\
\cmidrule(r){1-1}
\cmidrule(r){2-3}
\cmidrule{4-5}
MI only             
    &65.04      &50.08
    &29.13      &25.57
\\
\cmidrule(r){1-1}
\cmidrule(r){2-3}
\cmidrule{4-5}
HPS only            
    &59.43              &46.87 
    &\TINY{\it n.a.}   &\TINY{\it n.a.}
\\
\cmidrule(r){1-1}
\cmidrule(r){2-3}
\cmidrule{4-5}
MI+Real(0.25)
    &61.34      &48.47
    &29.16         &25.87
\\
MI+Real(0.5)   
    &61.63      &49.50
    &29.38          &25.92
\\
MI+Real(0.9)    
    &59.83      &48.92
    &28.60          &25.60
\\
\bottomrule
\end{tabular}
}
\end{minipage}
\begin{minipage}[t]{0.58\textwidth}
For completeness, we perform an experiment where we fine-tune based on a dataset selected via \gls{HPS} scores. Results in \Cref{tab:appendix_results_finetuning_composition} (same as \Cref{tab:results_finetuning_composition}, but duplicated here for simplicity) show that selecting fine-tuning samples based on \gls{MI} outperforms such an alternative strategy, using \acrshort{BLIP}.

\vspace{3pt}
Next, another natural question to ask is whether the self-supervised fine-tuning method we suggest in this work is a valid strategy. Indeed, instead of using synthetic image data for fine-tuning the base model, it is also possible to use real-life, \end{minipage}

captioned image data. Then, we present an ablation on the use of real samples, along with synthetic images, in the fine-tuning 
procedure. In \Cref{tab:appendix_results_finetuning_composition}(bottom) we report the experimental results obtained by composing the fine-tuning dataset by 
imposing the ratio of images generated by the SD model to $x$, and the ratio of real images taken from the  CC12M dataset \citep{changpinyo2021conceptual} to $(1-x)$,
where in both cases we select the candidate images to be used in the fine-tuning set $\mathcal{S}$ using \gls{MI}. So, for example, \acrshort{MI}+Real(0.25) indicates that we use 25\% of real images. 
Interestingly, we observe the following trend. Complementing the synthetically generated samples with few real ones does not benefit alignment (lower \acrshort{BLIP}) but might have a positive effect for aesthetics (higher \acrshort{HPS}).

\noindent\textbf{Fine-tuning set size.} We continue by reporting an ablation on the fine-tuning set $\mathcal{S}$ size. 

Specifically, based on Algorithm~\ref{algo:mitune}, two parameters determine both the quality and the associated computational cost related to the fine-tuning set $\mathcal{S}$: the number of candidate images $M$, and how many $k$ are selected to be included in $\mathcal{S}$. 

\begin{table}[H]
\small
\centering
\captionof{table}{\acrshort{BLIP} alignment results on \gls{T2I}-CompBench's Color and Shape categories varying size and composition of fine-tuning set. Results obtained using $R$=1.}
\label{tab:appendix_ablation_M_k}
\scalebox{0.9}{
\begin{tabular}{
    @{$\:\:\:$}
    r
    @{$\:\:$}r
    @{$\quad\quad$}
    r
    @{$\quad\quad$}
    r
}
\toprule
    \multicolumn{2}{c}{Hyper-params} & 
    \multicolumn{2}{c}{Category}  
\\
\cmidrule(r){1-2}
\cmidrule{3-4}
    $M$ & $k$ & Color & Shape
\\
\cmidrule(r){1-2}
\cmidrule{3-4}
    30      &1 &  
    58.12   &47.48
\\
\cmidrule(r){1-2}
\cmidrule{3-4}
    50     &7 & 
    59.31  &47.26
\\
    50      &1 &
    61.57   &48.40
\\
\cmidrule(r){1-2}
\cmidrule{3-4}
    100     &1 &
    60.12   &47.80
\\
    500     &1     &
    59.28   &46.79
    \\
\bottomrule
\end{tabular}
}
\end{table}

\Cref{tab:appendix_ablation_M_k} shows that the best performance is obtained selecting $2\%$ images (1 image out of 50).
We repeated the finetuning experiments on the categories Color and Shape 
by varying the selection ratio in the ranges $\{7/50,1/30,1/100,1/500\}$. Results indicate that the best selection ratio is the middle-range corresponding to the baseline \gls{MITUNE}. 
We hypothesise that higher selection ratios pollute the fine-tuning set with lower quality images, while a more selective threshold favours images which have the highest alignment but possibly lower realism. Additionally, we remark that the number $M$ of candidate images has a negligible impact, above $M=50$, whereas fewer candidate images induce degraded performance. Hence, the value $M=50$ is, in our experiments, a sweet-spot that produces a valid candidate set, while not imposing a large computational burden.

\subsection[
Ablation: Fine-tuning model adapters and modalities
\newline
{\fontsize{8}{8}\selectfont
\textcolor{blue!50}{
\it Investigating LoRA, DoRA and fine-tuning or not CLIP $\cdot$ Referenced in \Cref{sec:mitune-finetuning}
}}
]
{Ablation: Fine-tuning model adapters and modalities}\label{app:abl_model}

In this Section, we provide additional results (\Cref{tab:results_ablation_model}) on \gls{MITUNE}, concerning which part of the pre-trained \gls{SD} model to fine-tune. In particular, we tried to fine-tune the denoising \acrshort{UNET} network alone and both the denoising and the text encoding (\acrshort{CLIP}) networks. The baseline results are obtained, as described in the main paper, with Do-RA \citep{liu2024dora} adapters. Switching to Lo-RA layers \cite{hu2021lora} incurs in a performance degradation, a trend observed also for other tasks in the literature \citep{liu2024dora}. Interestingly, joint fine-tuning of the \acrshort{UNET} backbone together with the text encoder layers degrades performance as well, which has also been observed in the literature \cite{huang2023t2icompbench}. Even if, in principle, a joint fine-tuning strategy should provide better results, as the amount of information transferred from the prompt to the image is bottle-necked by the text encoder architecture, we observed empirically more unstable training dynamics than the variant where only the score network backbone is fine-tuned, resulting in degraded performance.

\begin{table}[H]
\small
\centering
\caption{\acrshort{BLIP} alignment results on \gls{T2I}-CompBench's Color and Shape categories finetuning different portions of the model.
}
\label{tab:results_ablation_model}
\begin{tabular}{
    @{$\:\:\:$} %p{2.8cm}
    r
    @{$\:\:\:\:\:$}
    r
    @{$\:\:\:\:\:$}
    r
}
\toprule
\multirow{2}{*}{\bf Model} & 
\multicolumn{2}{c}{\bf Category}  \\
\cmidrule(r){2-3}
 & Color & Shape \\
\midrule
\gls{MITUNE} DoRA  &
  61.57 &    48.40    \\
\midrule
\gls{MITUNE} LoRA  & 
58.25      &  48.27  \\
\midrule
\gls{MITUNE} UNet+Text(joint)  & 
 57.88     &  47.79  \\
\bottomrule
\end{tabular}
\end{table}

\subsection[
Ablation: Combining categories into a single model
\newline
{
\fontsize{8}{8}\selectfont
\textcolor{blue!50}{
\it Investigating policies to create a monolithic model merging multiple categories $\cdot$ Reference in \Cref{sec:mitune-finetuning}
}}
]{Ablation: Combining categories into a single model\label{app:abl_Merge}}

The design space for T2I alignment improvement has
many options and this should call not only to investigate
alignment performance but also operational and computational costs.
For instance, fine-tuning methods require to create ad-hoc
models while one can argue that a single/multi-purpose
model might be a more lean and general solution.

This calls for investigating if/how different
task-specific fine-tuned models can be combined into a single 
model to address the different tasks at once. For the 
T2I-Combench, we considered two design options:
\begin{enumerate}
    \item \textbf{Weights merging}: 
    the DoRA weights of the 6 distinct per-category models are ``merged'' doing
    their arithmethic means forming a new ``meta'' model.

    \item \textbf{Joint optimization}:
    we create a new ``meta'' model by running a single fine-tuning process but using the union of the category-specific fine-tuning set.
\end{enumerate}

\begin{figure}[H]
\centering
\includegraphics[width=0.8\linewidth]{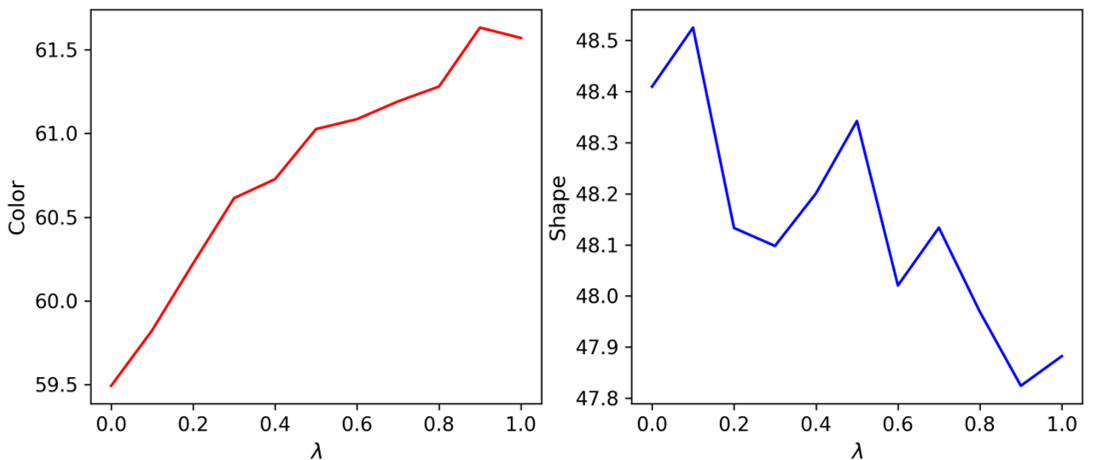}
\caption{
Weights merging: $\lambda \times$ Color + $(1.0-\lambda) \times$ Shape.
}
\label{fig:weights_merging}
\end{figure}

To start from a reference example, Fig.~\ref{fig:weights_merging} reports the BLIP-VQA obtained when testing on the color and shape test sets on the merged model obtained of the two task-specific models. The hyper-parameter $\lambda$ is used to balance the merging. For instance, at $\lambda=0$, the performance on color (left plot) are obtained using the shape-only model. Overall, the results show that these two categories are (partially) conflicting across all $\lambda$ values. Yet, a performance trade off might be sufficient in some scenarios.

\begin{table}[H]
\small
\centering
\caption{
Benchmarking strategies for combining models.
\label{tab:results_ablation_merge}
}
\begin{tabular}{
l
@{$\:\:\:$}l
@{$\:\:\:$}l
@{$\:\:\:$}l
@{$\:\:\:$}l
@{$\:\:\:$}l
@{$\:\:\:$}l
}
\toprule
\gls{MITUNE}  & Color & Shape & Texture & 2D-Spatial & Non-spatial & Complex \\
\it variants & 
\tinytiny{(\acrshort{BLIP})} & 
\tinytiny{(\acrshort{BLIP})} & 
\tinytiny{(\acrshort{BLIP})} & 
\tinytiny{(\acrshort{UNIDET})} & 
\tinytiny{(\acrshort{BLIP})} & 
\tinytiny{(\acrshort{BLIP})}  
\\
\midrule
% \gls{MITUNE} or \gls{MITUNE}$\star$  
{\it from} \cref{tab:results_blip}
& \bf61.57  &    \bf48.40   & \bf58.27   &  \bf18.51  &  67.77  & 53.54  \\
\midrule
% \gls{MITUNE}-
Model weighting  
&  58.50 & 48.23 & 58.22  &  16.72   &  68.28   &   54.35    \\
% \gls{MITUNE}-
Joint optimization
& 60.35  &  47.73  &  57.96  &  18.44  &  \bf69.68 &  \bf54.88     \\
\bottomrule
\end{tabular}
\end{table}

We then extended the analysis across all categories using a simple arithmetic mean for model merging, i.e., all models have the same weight.
Results are reported in Table~\ref{tab:results_ablation_merge} using \gls{MITUNE} as reference.
Overall, for most categories, the single ``meta'' model has degraded performance and 
neither weights merging nor joint optimization are the best alternative across all categories.

\clearpage
\section{Qualitative examples for \acrshort{T2ICOMPBENCH} using \acrshort{SDBASE}}\label{app:qualitative_sdbase}
%%%%%%%%%%%%%%%%%%%%%%%%%%%%%%%%%%%%%%%%%%%%%%%
%%% SHAPE
%%%%%%%%%%%%%%%%%%%%%%%%%%%%%%%%%%%%%%%%%%%%%%%
\vspace{-1cm}
\subsection{Color prompts}
\vspace{-2cm}
\begin{figure}[H]

\begin{tabular}{
@{}p{49pt}
@{}p{49pt}
@{}p{49pt}
@{}p{49pt}
@{}p{48pt}
@{}p{48pt}
@{}p{48pt}
@{}p{48pt}
@{}}
\hfil\fontsize{8.5}{8.5}\selectfont\acrshort{SDBASE} &
\hfil\acrshort{DPOK} &
\hfil\acrshort{GORS} &
\hfil\acrshort{HN} &
\hfil\acrshort{AE} &
\hfil\acrshort{SDG} &
\hfil\acrshort{SCG} &
\hfil\acrshort{MITUNE}
\\
\end{tabular}

\minipage{0.9\textwidth}
\includegraphics[width=\linewidth]{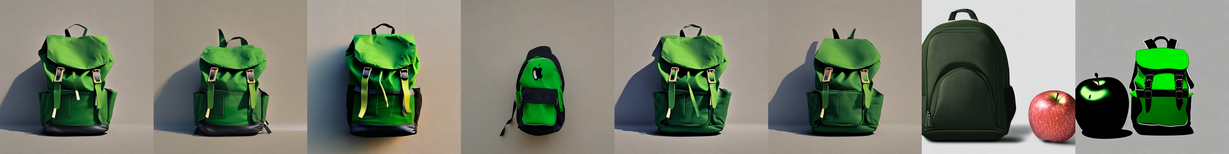}
\vspace{-0.7cm}
\caption*{a black apple and a green backpack}
\endminipage
\hfill
\vspace{0.1cm}
\minipage{0.9\textwidth}
\includegraphics[width=\linewidth]{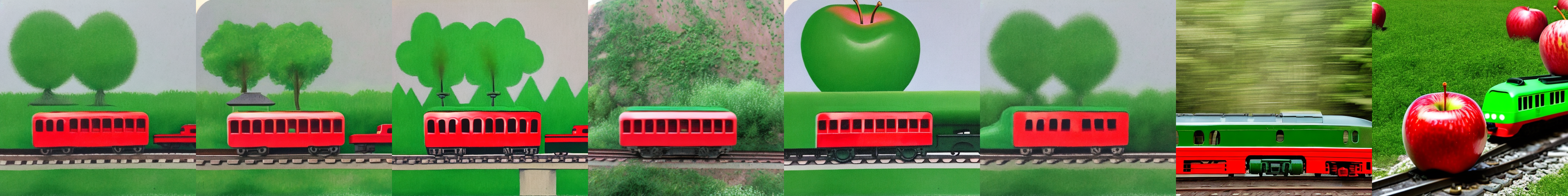}
\vspace{-0.7cm}
\caption*{a red apple and a green train}
\endminipage
\hfill
\minipage{0.9\textwidth}
\includegraphics[width=\linewidth]{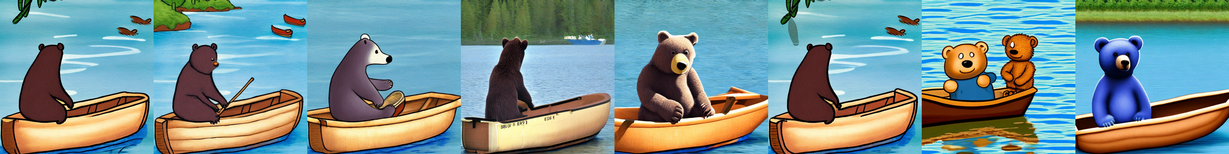}
\vspace{-0.7cm}
\caption*{a blue bear and a brown boat}
\endminipage
\hfill
\minipage{0.9\textwidth}
\includegraphics[width=\linewidth]{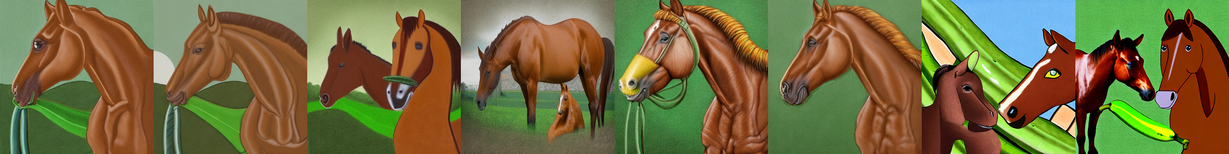}
\vspace{-0.7cm}
\caption*{a green banana and a brown horse}
\endminipage
\hfill
\minipage{0.9\textwidth}
\includegraphics[width=\linewidth]{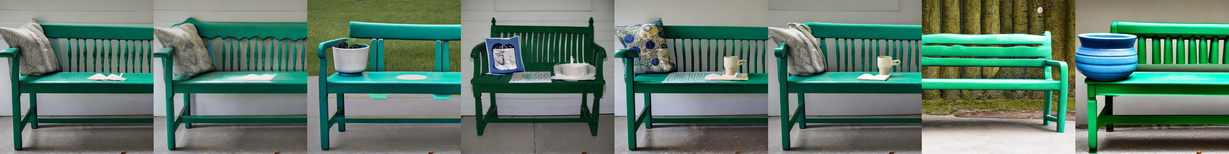}
\vspace{-0.7cm}
\caption*{a green bench and a blue bowl}
\endminipage
\hfill
\minipage{0.9\textwidth}
\includegraphics[width=\linewidth]{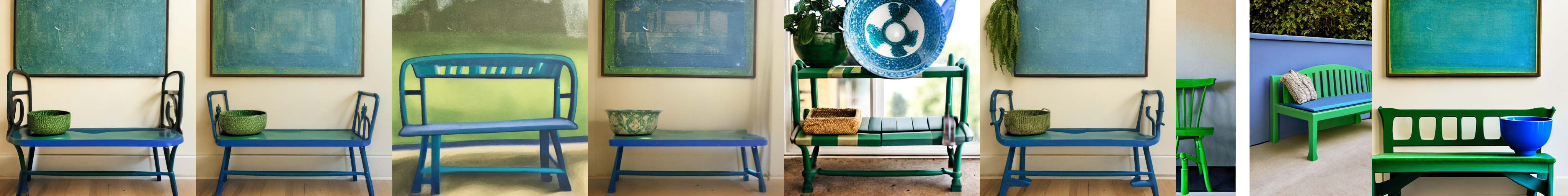}
\vspace{-0.7cm}
\caption*{a green bench and a blue bowl}
\endminipage
\hfill
\minipage{0.9\textwidth}
\includegraphics[width=\linewidth]{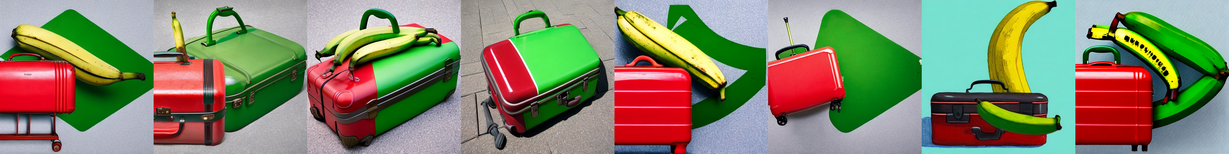}
\vspace{-0.7cm}
\caption*{a green banana and a red suitcase}
\endminipage
\hfill
\minipage{0.9\textwidth}
\includegraphics[width=\linewidth]{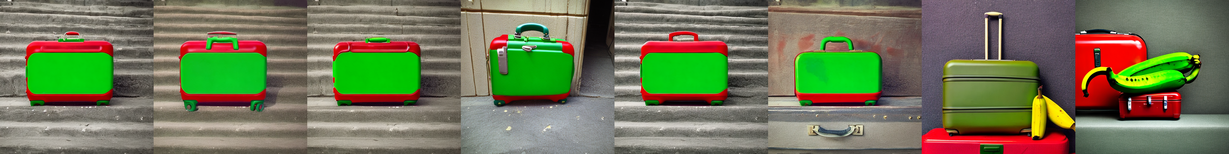}
\vspace{-0.7cm}
\caption*{a green banana and a red suitcase}
\endminipage
\hfill
\caption{Qualitative examples for Color from \Cref{tab:results_blip} (same seed used for a given prompt).}
\label{fig:benchmark-example-color1}
\end{figure}

%%%%%%%%%%%%%%%%%%%%%%%%%%%%%%%%%%%%%%%%%%%%%%%
%%% SHAPE
%%%%%%%%%%%%%%%%%%%%%%%%%%%%%%%%%%%%%%%%%%%%%%%
\vspace{-1cm}
\subsection{Shape prompts}

\begin{figure}[H]
\begin{tabular}{
@{}p{49pt}
@{}p{49pt}
@{}p{49pt}
@{}p{49pt}
@{}p{48pt}
@{}p{48pt}
@{}p{48pt}
@{}p{48pt}
@{}}
\hfil\fontsize{8.5}{8.5}\selectfont\acrshort{SDBASE} &
\hfil\acrshort{DPOK} &
\hfil\acrshort{GORS} &
\hfil\acrshort{HN} &
\hfil\acrshort{AE} &
\hfil\acrshort{SDG} &
\hfil\acrshort{SCG} &
\hfil\acrshort{MITUNE}
\\
\end{tabular}

\minipage{0.9\textwidth}
\includegraphics[width=\linewidth]{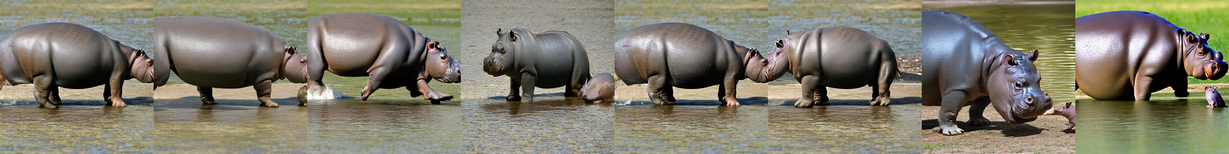}
\vspace{-0.5cm}
\caption*{a big hippopotamus and a small mouse}
\endminipage
\hfill
\minipage{0.9\textwidth}
\includegraphics[width=\linewidth]{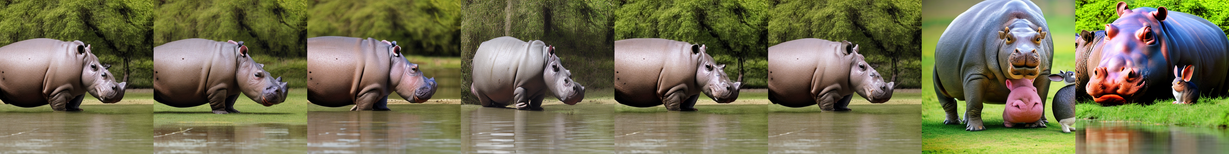}
\vspace{-0.7cm}
\caption*{a big hippopotamus and a small rabbit}
\endminipage
\hfill
\hfill
\minipage{0.9\textwidth}
\includegraphics[width=\linewidth]{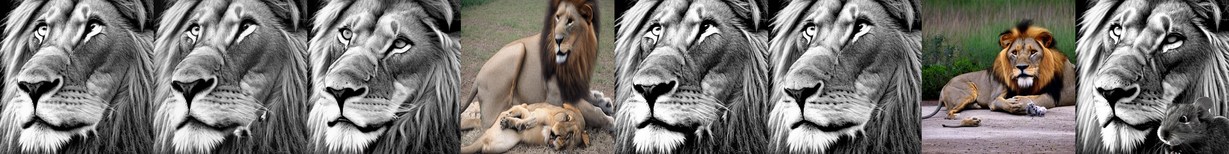}
\vspace{-0.7cm}
\caption*{a big lion and a small mouse}
\endminipage
\hfill
\minipage{0.9\textwidth}
\includegraphics[width=\linewidth]{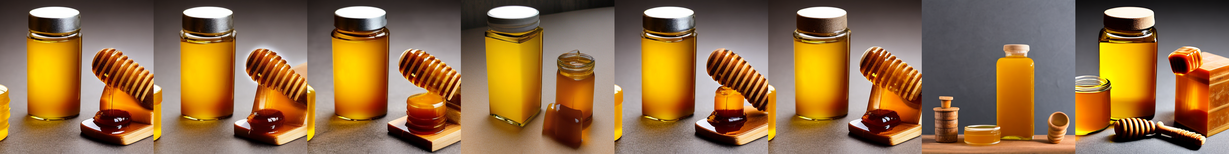}
\vspace{-0.7cm}
\caption*{a cubic block and a cylindrical jar of honey}
\endminipage
\hfill
\minipage{0.9\textwidth}
\includegraphics[width=\linewidth]{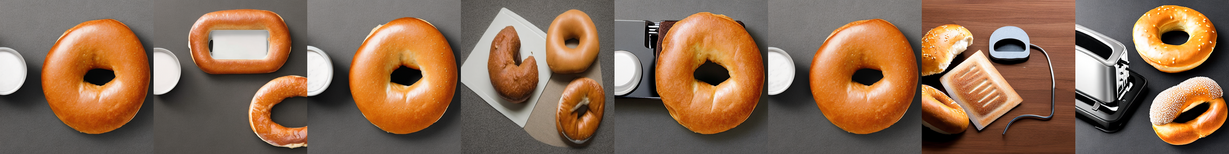}
\vspace{-0.7cm}
\caption*{a round bagel and a rectangular toaster}
\endminipage
\hfill
\minipage{0.9\textwidth}
\includegraphics[width=\linewidth]{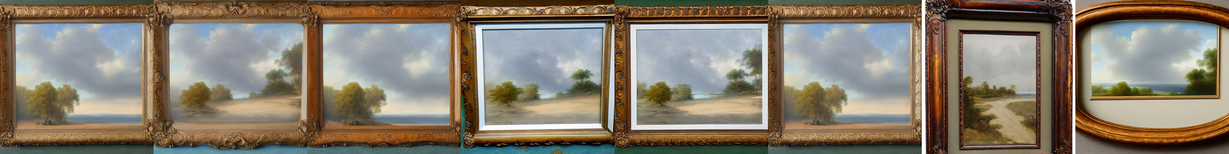}
\vspace{-0.7cm}
\caption*{an oval picture frame and a rectangular painting}
\endminipage
\hfill
\minipage{0.9\textwidth}
\includegraphics[width=\linewidth]{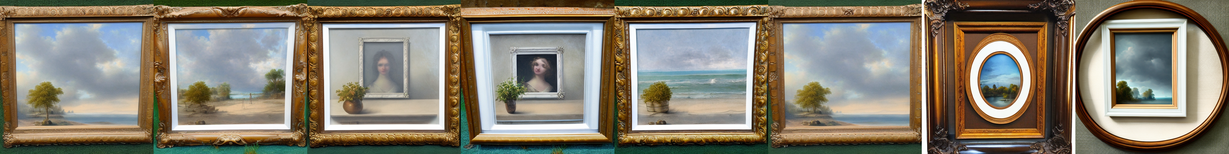}
\vspace{-0.7cm}
\caption*{an oval picture frame and a square painting}
\endminipage
\hfill
\minipage{0.9\textwidth}
\includegraphics[width=\linewidth]{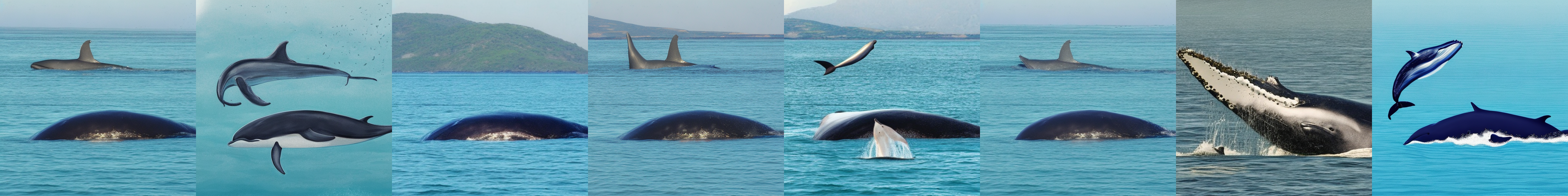}
\vspace{-0.7cm}
\caption*{a big whale and a small dolphin}
\endminipage
\hfill
\caption{Qualitative examples of the Shape category from \Cref{tab:results_blip} (same seed used for a given prompt).}
\label{fig:benchmark-example-shape1}
\end{figure}

%%%%%%%%%%%%%%%%%%%%%%%%%%%%%%%%%%%%%%%%%%%%%%%
%%% TEXTURE
%%%%%%%%%%%%%%%%%%%%%%%%%%%%%%%%%%%%%%%%%%%%%%%
\subsection{Texture prompts}
\begin{figure}[H]
\begin{tabular}{
@{}p{49pt}
@{}p{49pt}
@{}p{49pt}
@{}p{49pt}
@{}p{48pt}
@{}p{48pt}
@{}p{48pt}
@{}p{48pt}
@{}}
\hfil\fontsize{8.5}{8.5}\selectfont\acrshort{SDBASE} &
\hfil\acrshort{DPOK} &
\hfil\acrshort{GORS} &
\hfil\acrshort{HN} &
\hfil\acrshort{AE} &
\hfil\acrshort{SDG} &
\hfil\acrshort{SCG} &
\hfil\acrshort{MITUNE}
\\
\end{tabular}

\minipage{0.9\textwidth}
\includegraphics[width=\linewidth]{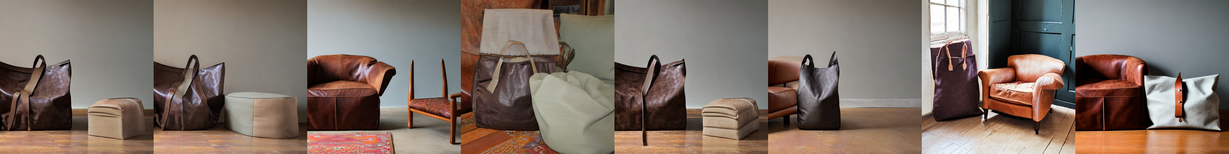}
\vspace{-0.7cm}
\caption*{a fabric bag and a leather chair}
\endminipage
\hfill
\minipage{0.9\textwidth}
\includegraphics[width=\linewidth]{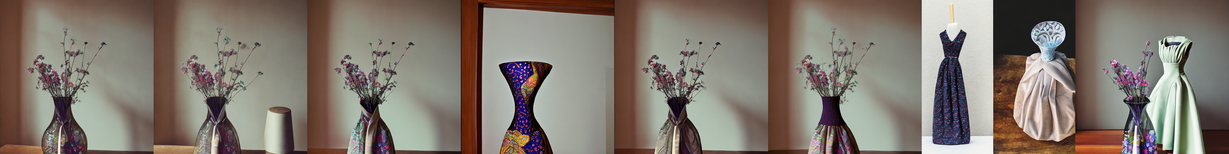}
\vspace{-0.7cm}
\caption*{a fabric dress and a glass vase}
\endminipage
\hfill
\minipage{0.9\textwidth}
\includegraphics[width=\linewidth]{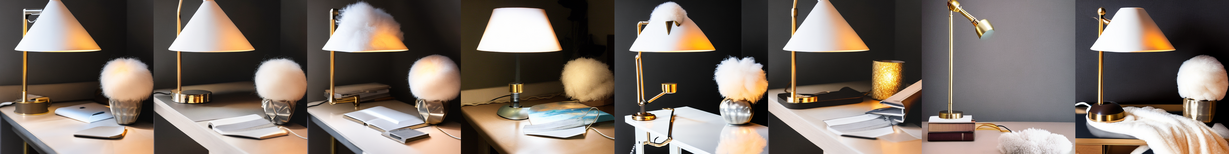}
\vspace{-0.7cm}
\caption*{a metallic desk lamp and a fluffy blanket}
\endminipage
\hfill
\minipage{0.9\textwidth}
\includegraphics[width=\linewidth]{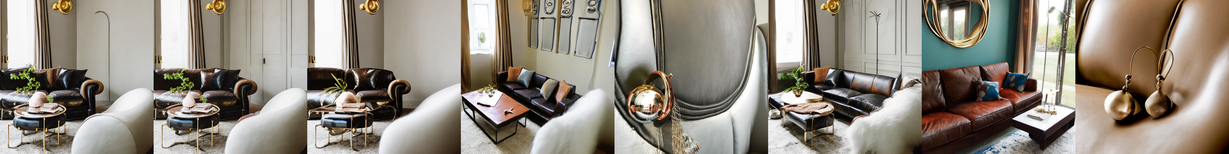}
\vspace{-0.7cm}
\caption*{a metallic earring and a leather sofa}
\endminipage
\hfill
\minipage{0.9\textwidth}
\includegraphics[width=\linewidth]{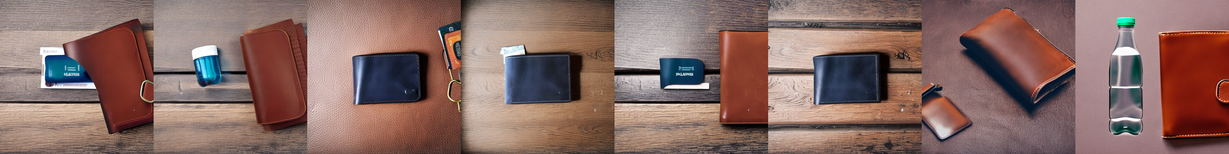}
\vspace{-0.7cm}
\caption*{a plastic bottle and a leather wallet}
\endminipage
\hfill
\minipage{0.9\textwidth}
\includegraphics[width=\linewidth]{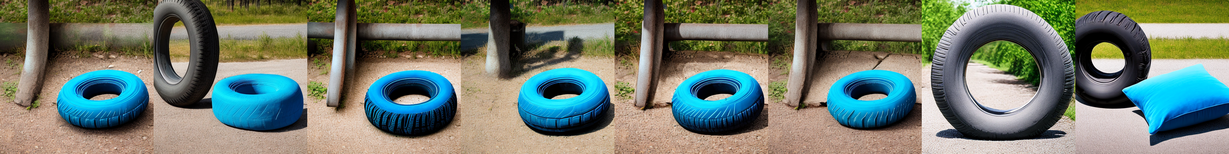}
\vspace{-0.7cm}
\caption*{a rubber tire and a fabric pillow}
\endminipage
\hfill
\minipage{0.9\textwidth}
\includegraphics[width=\linewidth]{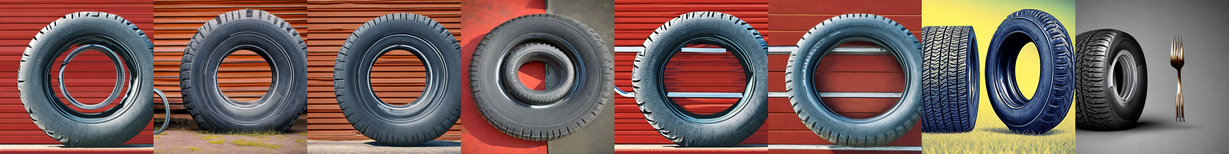}
\vspace{-0.7cm}
\caption*{a rubber tire and a metallic fork}
\endminipage
\hfill
\minipage{0.9\textwidth}
\includegraphics[width=\linewidth]{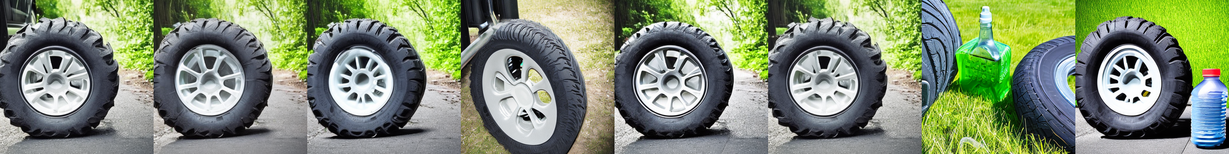}
\vspace{-0.7cm}
\caption*{a rubber tire and a plastic bottle}
\endminipage
\hfill
\caption{Qualitative examples of the Texture category from \Cref{tab:results_blip} (same seed used for a given prompt).}
\label{fig:benchmark-example-texture1}
\end{figure}

%%%%%%%%%%%%%%%%%%%%%%%%%%%%%%%%%%%%%%%%%%%%%%%
%%% 2D-SPATIAL
%%%%%%%%%%%%%%%%%%%%%%%%%%%%%%%%%%%%%%%%%%%%%%

\subsection{2D-Spatial prompts}

\begin{figure}[H]
\begin{tabular}{
@{}p{49pt}
@{}p{49pt}
@{}p{49pt}
@{}p{49pt}
@{}p{48pt}
@{}p{48pt}
@{}p{48pt}
@{}p{48pt}
@{}}
\hfil\fontsize{8.5}{8.5}\selectfont\acrshort{SDBASE} &
\hfil\acrshort{DPOK} &
\hfil\acrshort{GORS} &
\hfil\acrshort{HN} &
\hfil\acrshort{AE} &
\hfil\acrshort{SDG} &
\hfil\acrshort{SCG} &
\hfil\acrshort{MITUNE}
\\
\end{tabular}

\minipage{0.9\textwidth}
\includegraphics[width=\linewidth]{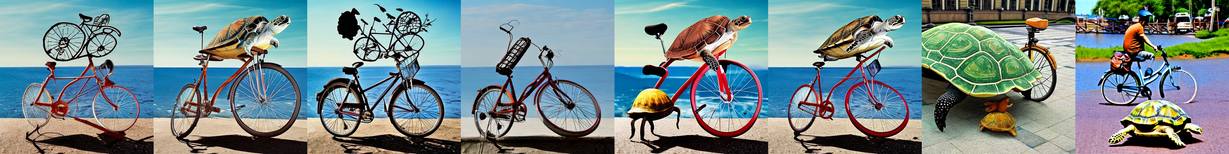}
\vspace{-0.7cm}
\caption*{a bicycle on the top of a turtle}
\endminipage
\hfill
\minipage{0.9\textwidth}
\includegraphics[width=\linewidth]{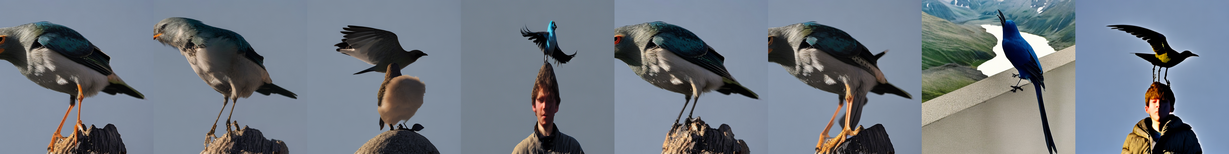}
\vspace{-0.7cm}
\caption*{a bird on the top of a person}
\endminipage
\hfill
\minipage{0.9\textwidth}
\includegraphics[width=\linewidth]{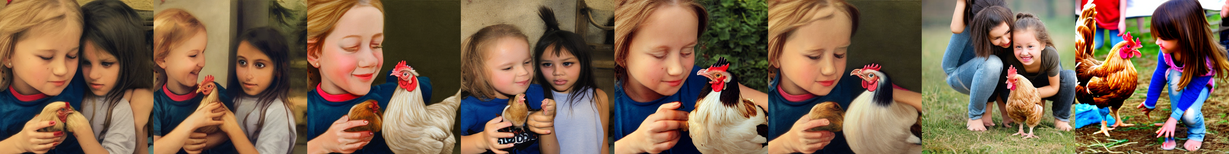}
\vspace{-0.7cm}
\caption*{a chicken on the left of a girl}
\endminipage
\hfill
\minipage{0.9\textwidth}
\includegraphics[width=\linewidth]{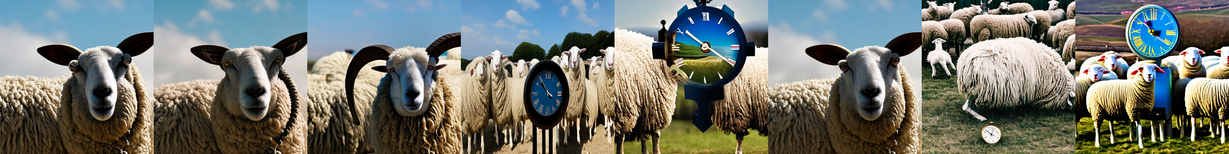}
\vspace{-0.7cm}
\caption*{a clock on the top of a sheep}
\endminipage
\hfill
\minipage{0.9\textwidth}
\includegraphics[width=\linewidth]{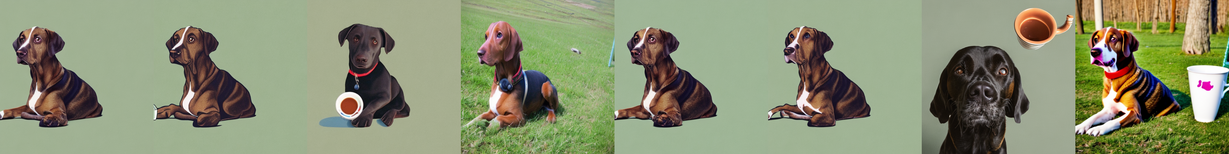}
\vspace{-0.7cm}
\caption*{a cup on the right of a dog}
\endminipage
\hfill
\minipage{0.9\textwidth}
\includegraphics[width=\linewidth]{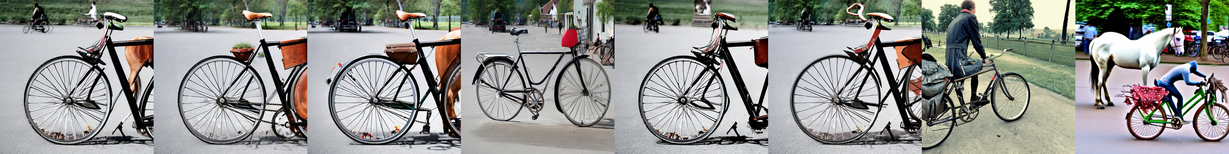}
\vspace{-0.7cm}
\caption*{a horse on side of a bicycle}
\endminipage
\hfill
\minipage{0.9\textwidth}
\includegraphics[width=\linewidth]{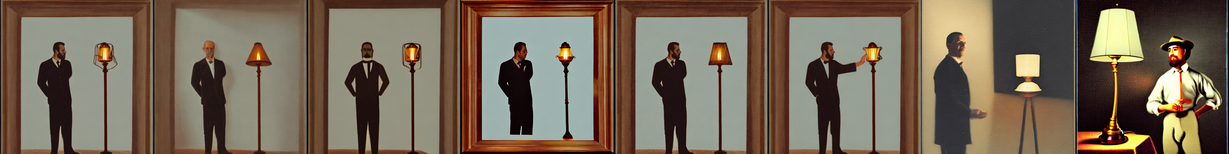}
\vspace{-0.7cm}
\caption*{a man on the right of a lamp}
\endminipage
\hfill
\minipage{0.9\textwidth}
\includegraphics[width=\linewidth]{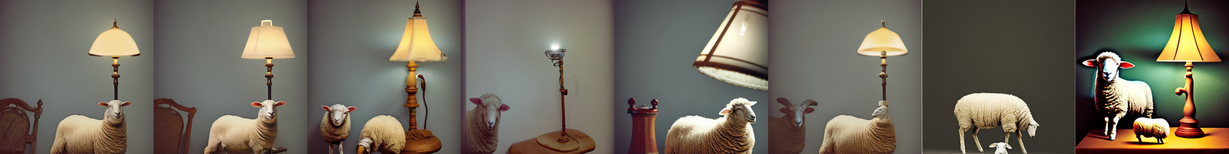}
\vspace{-0.7cm}
\caption*{a sheep on the left of a lamp}
\endminipage
\hfill
\caption{Qualitative examples of the 2D-Spatial category from \Cref{tab:results_blip} (same seed used for a given prompt).}
\label{fig:benchmark-example-spatial}
\end{figure}

%%%%%%%%%%%%%%%%%%%%%%%%%%%%%%%%%%%%%%%%%%%%%%%
%%% NON-SPATIAL
%%%%%%%%%%%%%%%%%%%%%%%%%%%%%%%%%%%%%%%%%%%%%%

\subsection{Non-Spatial prompts}

\begin{figure}[H]
\begin{tabular}{
@{}p{49pt}
@{}p{49pt}
@{}p{49pt}
@{}p{49pt}
@{}p{48pt}
@{}p{48pt}
@{}p{48pt}
@{}p{48pt}
@{}}
\hfil\fontsize{8.5}{8.5}\selectfont\acrshort{SDBASE} &
\hfil\acrshort{DPOK} &
\hfil\acrshort{GORS} &
\hfil\acrshort{HN} &
\hfil\acrshort{AE} &
\hfil\acrshort{SDG} &
\hfil\acrshort{SCG} &
\hfil\acrshort{MITUNE}
\\
\end{tabular}

\minipage{0.9\textwidth}
\includegraphics[width=\linewidth]{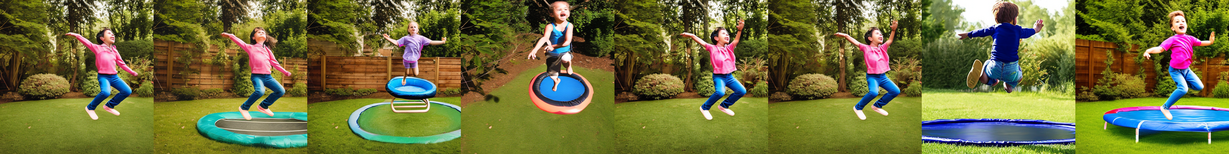}
\vspace{-0.7cm}
\caption*{A child is jumping on a trampoline in their backyard.}
\endminipage
\hfill
\minipage{0.9\textwidth}
\includegraphics[width=\linewidth]{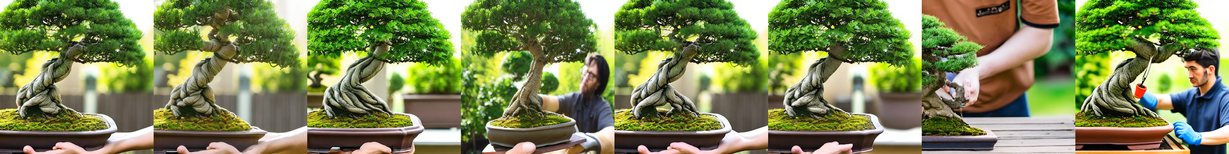}
\vspace{-0.7cm}
\caption*{A gardener is pruning a beautiful bonsai tree.}
\endminipage
\hfill
\minipage{0.9\textwidth}
\includegraphics[width=\linewidth]{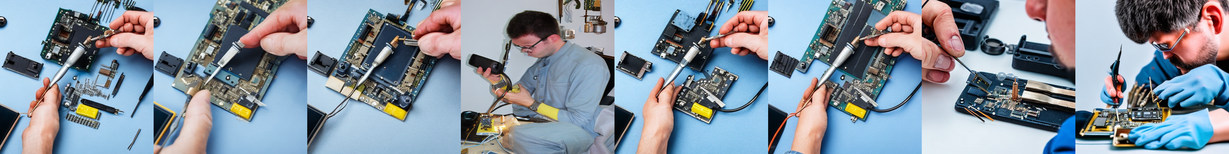}
\vspace{-0.7cm}
\caption*{A man is holding a soldering iron and repairing a broken electronic device.}
\endminipage
\hfill
\minipage{0.9\textwidth}
\includegraphics[width=\linewidth]{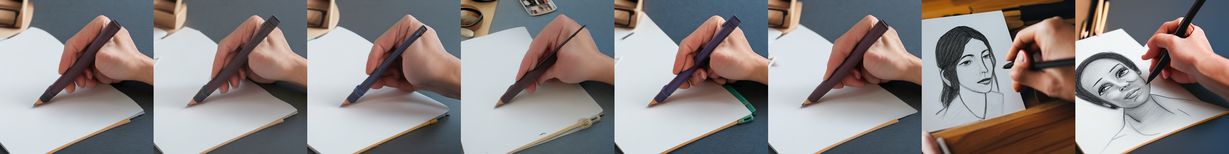}
\vspace{-0.7cm}
\caption*{A person is holding a pencil and sketching a portrait.}
\endminipage
\hfill
\minipage{0.9\textwidth}
\includegraphics[width=\linewidth]{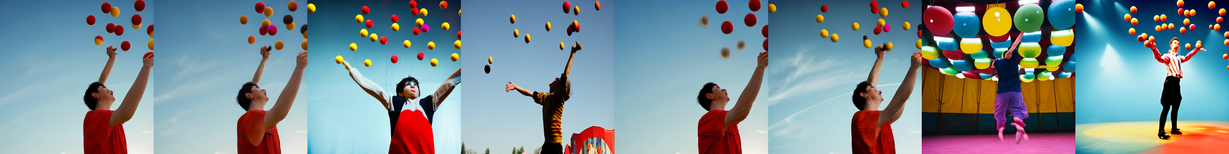}
\vspace{-0.7cm}
\caption*{A person is practicing their juggling skills at the circus.}
\endminipage
\hfill
\minipage{0.9\textwidth}
\includegraphics[width=\linewidth]{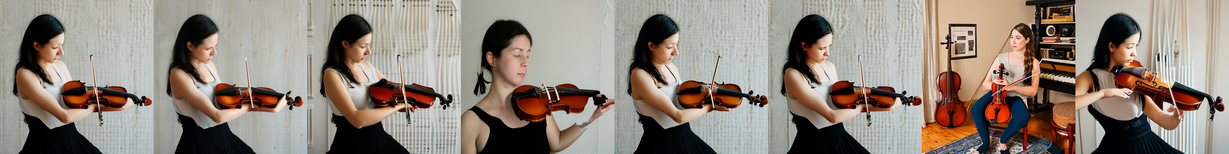}
\vspace{-0.7cm}
\caption*{A woman is practicing her violin in her music room.}
\endminipage
\hfill
\minipage{0.9\textwidth}
\includegraphics[width=\linewidth]{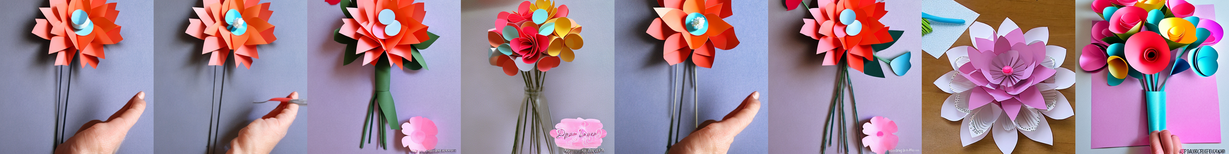}
\vspace{-0.7cm}
\caption*{The paper crafter is making a paper flower bouquet.}
\endminipage
\hfill
\minipage{0.9\textwidth}
\includegraphics[width=\linewidth]{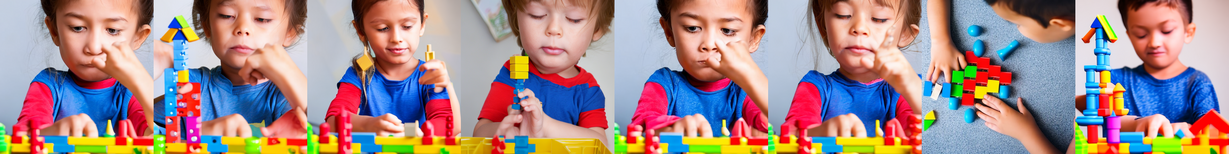}
\vspace{-0.7cm}
\caption*{A child is playing with a toy construction set and building a tower.}
\endminipage
\hfill
\caption{Qualitative examples of the Non-spatial category from \Cref{tab:results_blip} (same seed used for a given prompt).}
\label{fig:benchmark-example-non-spatial1}
\end{figure}

%%%%%%%%%%%%%%%%%%%%%%%%%%%%%%%%%%%%%%%%%%%%%%%
%%% NON-SPATIAL
%%%%%%%%%%%%%%%%%%%%%%%%%%%%%%%%%%%%%%%%%%%%%%

\subsection{Complex prompts}

\begin{figure}[H]
\begin{tabular}{
@{}p{49pt}
@{}p{49pt}
@{}p{49pt}
@{}p{49pt}
@{}p{48pt}
@{}p{48pt}
@{}p{48pt}
@{}p{48pt}
@{}}
\hfil\fontsize{8.5}{8.5}\selectfont\acrshort{SDBASE} &
\hfil\acrshort{DPOK} &
\hfil\acrshort{GORS} &
\hfil\acrshort{HN} &
\hfil\acrshort{AE} &
\hfil\acrshort{SDG} &
\hfil\acrshort{SCG} &
\hfil\acrshort{MITUNE}
\\
\end{tabular}

\minipage{0.9\textwidth}
\includegraphics[width=\linewidth]{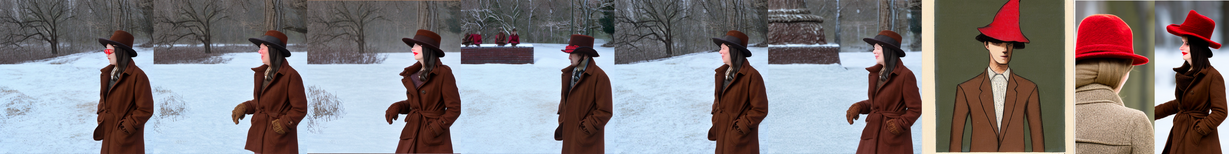}
\vspace{-0.7cm}
\caption*{The red hat was on top of the brown coat.}
\endminipage
\hfill
\minipage{0.9\textwidth}
\includegraphics[width=\linewidth]{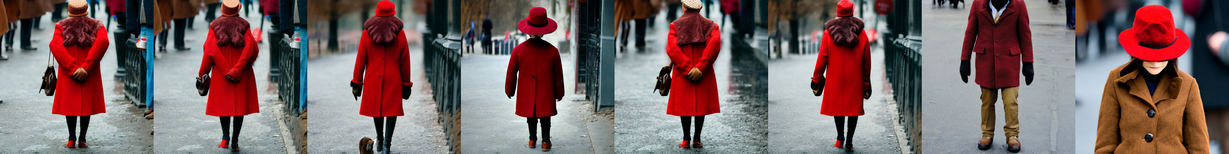}
\vspace{-0.7cm}
\caption*{The red hat was on top of the brown coat.}
\endminipage
\hfill
\minipage{0.9\textwidth}
\includegraphics[width=\linewidth]{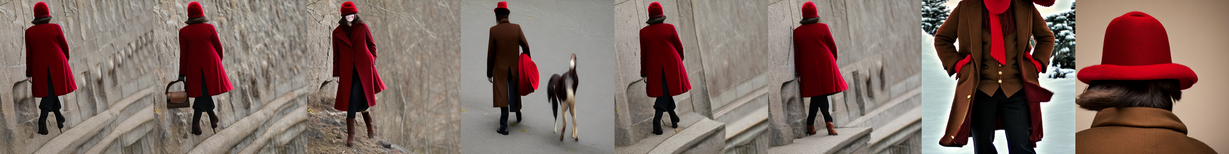}
\vspace{-0.7cm}
\caption*{The red hat was on top of the brown coat.}
\endminipage
\hfill
\minipage{0.9\textwidth}
\includegraphics[width=\linewidth]{arxiv_iclr25/images/benchmark/complex/The_white_mug_is_on_top_of_the_black_coaster._326.png}
\vspace{-0.7cm}
\caption*{The white mug is on top of the black coaster.}
\endminipage
\hfill
\minipage{0.9\textwidth}
\includegraphics[width=\linewidth]{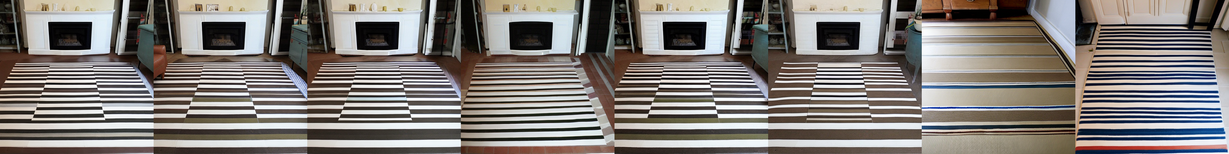}
\vspace{-0.7cm}
\caption*{The striped rug was on top of the tiled floor.}
\endminipage
\hfill
\minipage{0.9\textwidth}
\includegraphics[width=\linewidth]{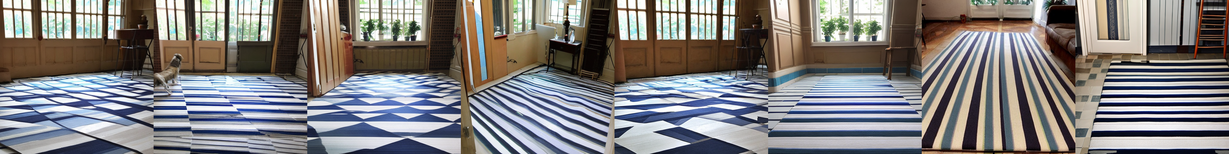}
\vspace{-0.7cm}
\caption*{The striped rug was on top of the tiled floor.}
\endminipage
\hfill
\minipage{0.9\textwidth}
\includegraphics[width=\linewidth]{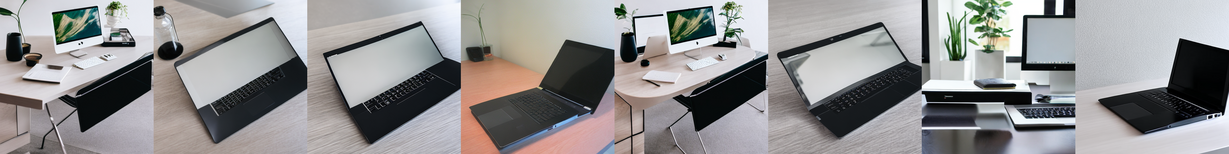}
\vspace{-0.7cm}
\caption*{The sleek black laptop sat on the clean white desk.}
\endminipage
\hfill
\minipage{0.9\textwidth}
\includegraphics[width=\linewidth]{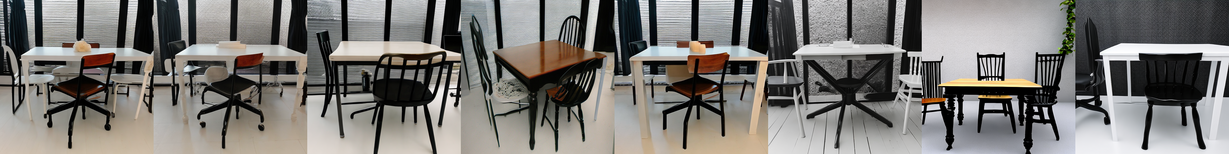}
\vspace{-0.7cm}
\caption*{The black chair was on the left of the white table.}
\endminipage
\caption{Qualitative examples of the Complex category from \Cref{tab:results_blip} (same seed used for a given prompt).}
\label{fig:benchmark-example-complex1}
\end{figure}

\clearpage
\section{Qualitative examples for \acrshort{T2ICOMPBENCH} using \acrshort{SDXL}}\label{app:qualitative_sdxl}

%%%%%%%%%%%%%%%%%%%%%%%%%%%%%%%%%%%%%%%%%
%% COLOR
%%%%%%%%%%%%%%%%%%%%%%%%%%%%%%%%%%%%%%%%%
\begin{minipage}[t]{\textwidth}
\begin{tabular}{
    @{}p{1.3cm}
    @{}p{1.7cm}
    @{}p{1.7cm}
    @{$\,\,$}p{1.7cm}
    @{}p{1.7cm}
    @{$\,\,$}p{1.7cm}
    @{}p{1.7cm}
    @{$\,\,$}p{1.7cm}
    @{}p{1.7cm}
    @{}
}    
&
\hfil\acrshort{SDXL} &
\hfil\acrshort{MITUNE} &
\hfil\acrshort{SDXL} &
\hfil\acrshort{MITUNE} &
\hfil\acrshort{SDXL} &
\hfil\acrshort{MITUNE} &
\hfil\acrshort{SDXL} &
\hfil\acrshort{MITUNE}
\\
%%%%%%%%%%%%%%%%%%%%%%%%%%%%%%%%%%%%%%%%%%%%%%%
%% COLOR
%%%%%%%%%%%%%%%%%%%%%%%%%%%%%%%%%%%%%%%%%%%%%%%
\fontsize{7}{7}\selectfont
    \vspace{-1cm}
    \textbf{Color} 
    \newline 
    prompts.
&
\multicolumn{2}{p{3.4cm}}{
    \includegraphics[width=\linewidth]{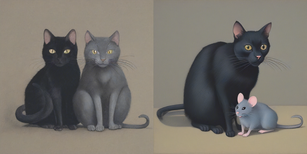}
    \newline
    \centering
    \fontsize{7}{7}\selectfont 
    a black cat and a gray mouse
} 
&
\multicolumn{2}{p{3.4cm}}{
    \includegraphics[width=\linewidth]{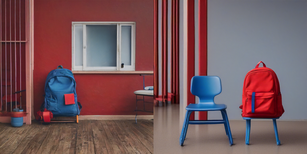}
    \newline
    \fontsize{7}{7}\selectfont
    \centering
    a red backpack and a blue chair
} 
&
\multicolumn{2}{p{3.4cm}}{
    \includegraphics[width=\linewidth]{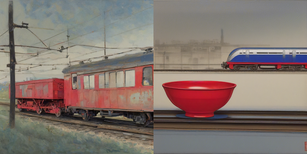}
    \newline
    \fontsize{7}{7}\selectfont
    \centering
    a red bowl and a blue train
}
&
\multicolumn{2}{p{3.4cm}}{
    \includegraphics[width=\linewidth]{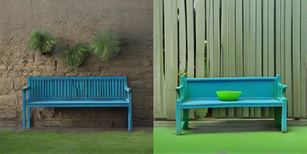}
    \newline
    \fontsize{7}{7}\selectfont
    \centering
    a blue bench and a green bowl
}
\\[2em]
%%%%%%%%%%%%%%%%%%%%%%%%%%%%%%%%%%%%%%%%%%%%%%%
%% SHAPE
%%%%%%%%%%%%%%%%%%%%%%%%%%%%%%%%%%%%%%%%%%%%%%%
\fontsize{7}{7}\selectfont
    \vspace{-1cm}
    \textbf{Shape} 
    \newline 
    prompts. 
&
\multicolumn{2}{p{3.4cm}}{
    \includegraphics[width=\linewidth]{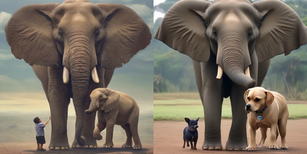}
    \newline
    \fontsize{7}{7}\selectfont
    \centering
    a big elephant and a small dog
}
&
\multicolumn{2}{p{3.4cm}}{
    \includegraphics[width=\linewidth]{arxiv_iclr25/images/sdxl_shape/a_big_lion_and_a_small_mouse_1.png}
    \newline
    \fontsize{7}{7}\selectfont
    \centering
    a big lion and a small mouse
}
&
\multicolumn{2}{p{3.4cm}}{
    \includegraphics[width=\linewidth]{arxiv_iclr25/images/sdxl_shape/a_circular_mirror_and_a_triangular_shelf_unit_6.png}
    \newline
    \fontsize{7}{7}\selectfont
    \centering
    a circular mirror and a triangular shelf unit
}
&
\multicolumn{2}{p{3.4cm}}{
    \includegraphics[width=\linewidth]{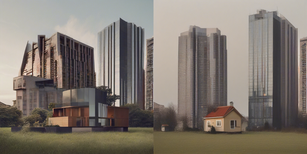}
    \newline
    \fontsize{7}{7}\selectfont
    \centering
    a tall skyscraper and a short cottage
}
\\[2em]
%%%%%%%%%%%%%%%%%%%%%%%%%%%%%%%%%%%%%%%%%%%%%%%
%% TEXTURE
%%%%%%%%%%%%%%%%%%%%%%%%%%%%%%%%%%%%%%%%%%%%%%%
\fontsize{7}{7}\selectfont
    \vspace{-1cm}
    \textbf{Texture} 
    \newline 
    prompts. 
&
\multicolumn{2}{p{3.4cm}}{
    \includegraphics[width=\linewidth]{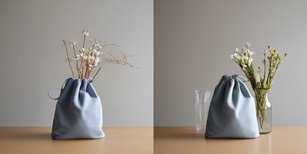}
    \newline
    \centering
    \fontsize{7}{7}\selectfont 
    a fabric bag and a glass vase
} 
&
\multicolumn{2}{p{3.4cm}}{
    \includegraphics[width=\linewidth]{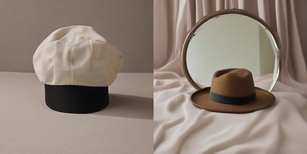}
    \newline
    \centering
    \fontsize{7}{7}\selectfont 
    a fabric hat and a glass mirror
} 
&
\multicolumn{2}{p{3.4cm}}{
    \includegraphics[width=\linewidth]{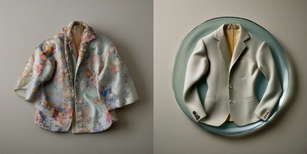}
    \newline
    \centering
    \fontsize{7}{7}\selectfont 
    a fabric jacket and a glass plate
} 
&
\multicolumn{2}{p{3.4cm}}{
    \includegraphics[width=\linewidth]{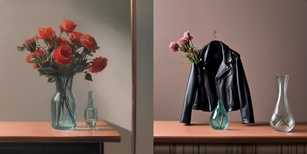}
    \newline
    \centering
    \fontsize{7}{7}\selectfont 
    a leather jacket and a glass vase
} 
\\[2em]
%%%%%%%%%%%%%%%%%%%%%%%%%%%%%%%%%%%%%%%%%%%%%%%
%% 2D-SPATIAL
%%%%%%%%%%%%%%%%%%%%%%%%%%%%%%%%%%%%%%%%%%%%%%%
\fontsize{7}{7}\selectfont
    \vspace{-1cm}
    \textbf{2D-Spatial}
    \newline
    prompts. 
&
\multicolumn{2}{p{3.4cm}}{
    \includegraphics[width=\linewidth]{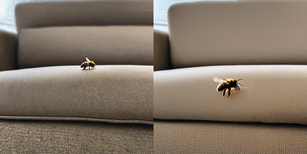}
    \newline
    \centering
    \fontsize{7}{7}\selectfont 
    a bee on side of a couch
} 
&
\multicolumn{2}{p{3.4cm}}{
    \includegraphics[width=\linewidth]{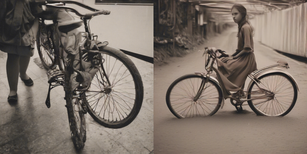}
    \newline
    \centering
    \fontsize{7}{7}\selectfont 
    a bicycle on the bottom of a girl
} 
&
\multicolumn{2}{p{3.4cm}}{
    \includegraphics[width=\linewidth]{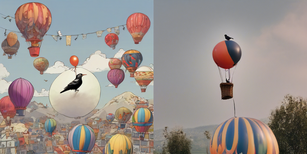}
    \newline
    \centering
    \fontsize{7}{7}\selectfont 
    a bird on the top of a balloon
} 
&
\multicolumn{2}{p{3.4cm}}{
    \includegraphics[width=\linewidth]{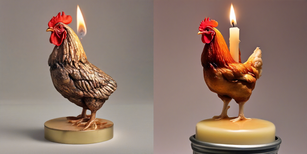}
    \newline
    \centering
    \fontsize{7}{7}\selectfont 
    a candle on the top of a chicken
} 
\\[2em]
%%%%%%%%%%%%%%%%%%%%%%%%%%%%%%%%%%%%%%%%%%%%%%%
%% NON-SPATIAL
%%%%%%%%%%%%%%%%%%%%%%%%%%%%%%%%%%%%%%%%%%%%%%%
\fontsize{7}{7}\selectfont
    \vspace{-1cm}
    \textbf{Non-Spatial}
    \newline
    prompts. 
&
\multicolumn{2}{p{3.4cm}}{
    \includegraphics[width=\linewidth]{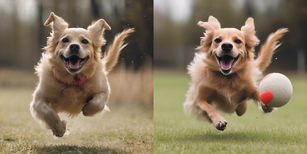}
    \newline
    \centering
    \fontsize{7}{7}\selectfont 
    A dog is chasing after a ball and wagging its tail
} 
&
\multicolumn{2}{p{3.4cm}}{
    \includegraphics[width=\linewidth]{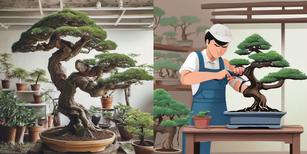}
    \newline
    \centering
    \fontsize{7}{7}\selectfont 
    A gardener is pruning a beautiful bonsai tree
} 
&
\multicolumn{2}{p{3.4cm}}{
    \includegraphics[width=\linewidth]{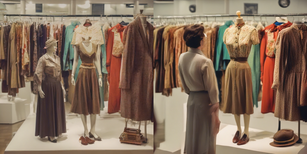}
    \newline
    \centering
    \fontsize{7}{7}\selectfont 
    A person is looking at a display of vintage clothing and admiring the fashion
} 
&
\multicolumn{2}{p{3.4cm}}{
    \includegraphics[width=\linewidth]{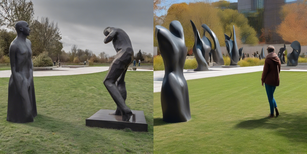}
    \newline
    \centering
    \fontsize{7}{7}\selectfont 
    A person is looking at a sculpture garden and appreciating the artwork
} 
\\[2em]
%%%%%%%%%%%%%%%%%%%%%%%%%%%%%%%%%%%%%%%%%%%%%%%
%% COMPLEX
%%%%%%%%%%%%%%%%%%%%%%%%%%%%%%%%%%%%%%%%%%%%%%%
\fontsize{7}{7}\selectfont
    \vspace{-1cm}
    \textbf{Complex} 
    \newline 
    prompts. 
&
\multicolumn{2}{p{3.4cm}}{
    \includegraphics[width=\linewidth]{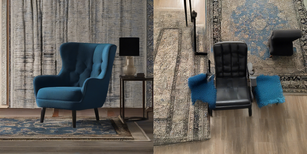}
    \newline
    \centering
    \fontsize{7}{7}\selectfont 
    The black chair is on top of the blue rug
} 
&
\multicolumn{2}{p{3.4cm}}{
    \includegraphics[width=\linewidth]{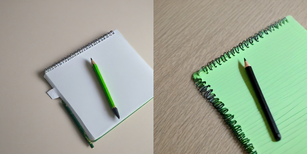}
    \newline
    \centering
    \fontsize{7}{7}\selectfont 
    The black pencil was next to the green notebook
} 
&
\multicolumn{2}{p{3.4cm}}{
    \includegraphics[width=\linewidth]{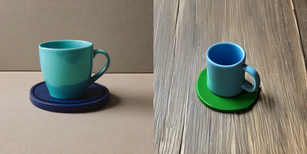}
    \newline
    \centering
    \fontsize{7}{7}\selectfont 
    The blue mug is on top of the green coaster
} 
&
\multicolumn{2}{p{3.4cm}}{
    \includegraphics[width=\linewidth]{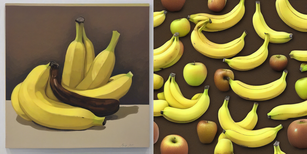}
    \newline
    \centering
    \fontsize{7}{7}\selectfont 
    The bright yellow banana contrasted with the dull brown apple
} 
\\
\end{tabular}
\captionof{figure}{Qualitative examples from \Cref{tab:sdxl} (same seed used for a given prompt).}
\end{minipage}

\clearpage
\section{Fine-tuning with \acrshort{DIFFUSIONDB} dataset}\label{app:qualitative_diffusiondb}
\subsection[
Selecting images and \acrshort{BLIP} prompts decomposition
\newline
{
\fontsize{8}{8}\selectfont
\it
\textcolor{blue!50}{
Discussing \acrshort{BLIP} limitation when handling DiffusionDB prompts $\cdot$ Referenced in \Cref{sec:results}
}}
]{Selecting images and \acrshort{BLIP} prompts decomposition
\label{app:diffusion_db_results}
}
In this section, we provide additional details about using prompts created by real users, i.e., DiffusionDB.

\noindent\textbf{Dataset properties.} DiffusionDB
 was collected scraping the StableDiffusion discord channels
``[...] \emph{We download chat messages from the Stable
Diffusion Discord channels with DiscordChatExporter, saving them as HTML files.
We focus on channels where users can command
a bot to run Stable Diffusion Version 1 to generate
images by typing a prompt, hyperparameters, and
the number of images} [...]''~\citep{diffusiondb}.
The scraped data is then packaged into parquet files (containing metadata such prompt, image filenames and hyperparams) and zip files (containing the actual images in WebP format) and made available on HuggingFace.

\noindent\textbf{Fine-tuning with \acrshort{DIFFUSIONDB}.}
We fine-tune \gls{SD}-2.1-base on 1,250 prompts randomly sampled and compare two different scenarios.
A first dataset is composed using images provided by DiffusionDB itself. As each prompt in DiffusionDB is paired to (about) 4 generated images we obtain a 5,000 prompt-image pairs reference dataset. For the second dataset, we use the 1,250 prompts to generate $M=50$ images for each prompt and selecting the $k=1$ image with the highest MI. We repeat this procedure 4 times to construct a complementary fine-tuning dataset with prompt-image 5,000 pairs.
We fine-tune \gls{SD}-2.1-base on each of the two datasets with pre-trained loss, then test on 500 DiffusionDB prompts (again, randomly selected and disjoint from the training set prompt-image pairs) generating 10 images for each test prompt. 

\begin{minipage}[t]{0.3\textwidth}
\vspace{-7pt}
\captionof{table}{
DiffusionDB.
}
\centering
\label{tab:appendix_results_diffusionDB}
\vspace{-8pt}
\scalebox{0.7}{
\begin{tabular}{
@{}
l 
@{$\:\:\:\:$}
l
@{}}
\toprule
Model                                 & \acrshort{HPS} \\
\midrule
\acrshort{SDBASE}                     & 23.99    \\
DiffusionDB                           & 24.35    \\
\midrule
\acrshort{MITUNE}                     & 25.32    
\\
\midrule

{\fontsize{9}{9}\selectfont\acrshort{MITUNE}$\,\boxminus\,$\it base}
	 &1.33   
\\
{\fontsize{9}{9}\selectfont\acrshort{MITUNE}$\,\boxminus\,$\it DiffusionDB}
	 &0.97
\\
\bottomrule
\end{tabular}
}
\\
\fontsize{6}{6}\selectfont
\textit A $\,\boxminus\,$ B shows the abs. difference between A and B.
% \end{table}
\end{minipage}
\begin{minipage}[t]{0.69\textwidth}
\vspace{5pt}
\Cref{tab:appendix_results_diffusionDB} (which is duplicating here \Cref{tab:results_diffusionDB} for simplicity)
shows the results. Fine-tuning either using the DiffusionDB images or \gls{MITUNE} can improve HPS score alignment
with respect to the \acrshort{SDBASE} baseline. Yet, \gls{MITUNE} improves upon using directly DiffusionDB images, 
i.e., using \gls{MI} is very competitive compared to (expensive) manual labeling.
\end{minipage}
    
\noindent\textbf{\acrshort{BLIP} prompts decomposition.}   
Our evaluation considers only \acrshort{HPS} as we find that the higher prompt complexity does not well suit the \acrshort{BLIP} prompt decomposition. Recall that \acrshort{BLIP} requires to split the prompt into ``noun phrases'', each used to create a VQA for the BLIP model. Specifically, \acrshort{BLIP} uses spaCy's English pipeline \texttt{en\_core\_web\_sm} to extract noun phrases from the prompt which result complex when the prompt is complex. Below we report some examples related to extracting first three noun phrases extracted from human prompts.

Examples of good/easy segmentations:
\begin{itemize}
    \item \textcolor{red}{concept art} of \textcolor{blue}{a silent hill monster}. painted by {\color{BlueGreen}edward hopper}.

    \item \textcolor{red}{anthropomorphic shark}, \textcolor{blue}{digital art}, {\color{BlueGreen}concept art}

    \item \textcolor{red}{geodesic landscape}, \textcolor{blue}{john chamberlain}, {\color{BlueGreen}christopher balaskas}, tadao ando, 4k
\end{itemize}

Examples of segmentations with missing/broad subjects:
\begin{itemize}
    \item \textcolor{red}{a realistic architectural visualization} of \textcolor{blue}{a sustainable mixed - use post - mordern post - growth walkable people} oriented {\color{BlueGreen}urban development}.
    
    \item \textcolor{red}{a realistic wide angle painting} of \textcolor{blue}{a vintage cathode ray tube}, in {\color{BlueGreen}a park}, in and advanced state of decay, psychedelic mushrooms all around, in a post apocalyptic city, ghibli, daytime, dynamic lighting 
    
    \item render of \textcolor{red}{dreamy beautiful landscape}, dreamy, by \textcolor{blue}{herbaceous plants}, {\color{BlueGreen}artger}, large scale, detailed vintage photo hyper realistic ultra realistic photo realistic photography, unreal engine, high detailed, 8 k

\end{itemize}

\clearpage

\subsection{Qualitative examples for \acrshort{DIFFUSIONDB}}

%%%%%%%%%%%%%%%%%%%%%%%%%%%%%%%%%%%%%%%%%
%% COLOR
%%%%%%%%%%%%%%%%%%%%%%%%%%%%%%%%%%%%%%%%%
\begin{minipage}[t]{\textwidth}
\begin{tabular}{
    @{}
    p{2.2cm}
    @{}p{2.2cm}
    @{}p{2.2cm}
    %%%%%%%%%%%%
    @{$\quad$}p{2.2cm}
    @{}p{2.2cm}
    @{}p{2.2cm}
    @{}
}    
\hfil\fontsize{8}{8}\selectfont\acrshort{SDBASE} &
\fontsize{7}{7}\selectfont Fine-tuned using\newline\acrshort{DIFFUSIONDB} images&
\hfil\fontsize{8}{8}\selectfont\acrshort{MITUNE} &
\hfil\fontsize{8}{8}\selectfont\acrshort{SDBASE} &
\fontsize{7}{7}\selectfont Fine-tuned using\newline\acrshort{DIFFUSIONDB} images&
\hfil\fontsize{8}{8}\selectfont\acrshort{MITUNE}
\\
\multicolumn{3}{p{6.6cm}}{
    \includegraphics[width=\linewidth]{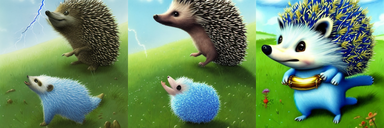}
    \newline
    \centering
    \fontsize{7}{7}\selectfont 
    ( ( ( ( a cute blue hedgehog with big gold ring and blue lightning in green grassland. ) ) ) ), big gold ring!, blue fur, clear sky, extremely detailed, fantasy painting, by jean - baptiste monge!!!!
}
&
\multicolumn{3}{p{6.6cm}}{
    \includegraphics[width=\linewidth]{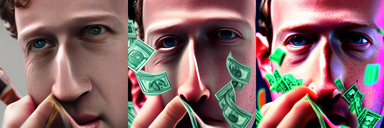}
    \newline
    \fontsize{7}{7}\selectfont 
    a closeup photorealistic photograph of mark zuckerberg eating money. film still, vibrant colors. this 4 k hd image is trending on artstation, featured on behance, well - rendered, extra crisp, features intricate detail, epic compo
}
\\
%%%%%%%%%%%%%%%%%%%%%%%%%%%%%
\multicolumn{3}{p{6.6cm}}{
    \includegraphics[width=\linewidth]{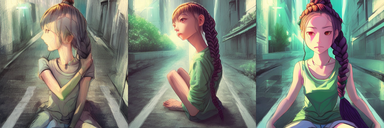}
    \newline
    \fontsize{7}{7}\selectfont 
    concept art, pretty girl sitting on street, braids blue and green, singular, junichi higashi, isamu imakake, intricate, balance, ultra detailed, full far frontal portrait, volumetric lighting, cinematic lighting + masterpiece
}
&
\multicolumn{3}{p{6.6cm}}{
    \includegraphics[width=\linewidth]{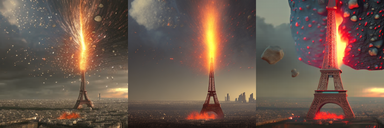}
    \newline
    \fontsize{7}{7}\selectfont
    the eifel tower gets hit by an asteroid, multiple asteroids are in the air, paris in the background is burning, apocalyptic, highly detailed, 4 k, digital paintin, sharp focus, tending on artstation
}
\\
%%%%%%%%%%%%%%%%%%%%%%%%%%%%%%%%%%%%%%%%%%%
\multicolumn{3}{p{6.6cm}}{
    \includegraphics[width=\linewidth]{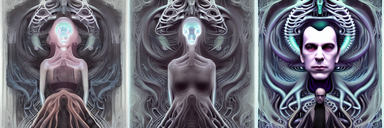}
    \newline
    \fontsize{7}{7}\selectfont
    cosmic lovecraft giger fractal random antihero portrait, pixar style, by tristan eaton stanley artgerm and tom bagshaw.
}
&
\multicolumn{3}{p{6.6cm}}{
    \includegraphics[width=\linewidth]{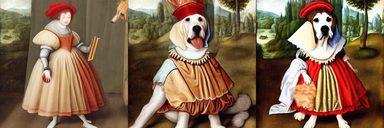}
    \newline
    \fontsize{7}{7}\selectfont
    a dog in a dress during the renaissance
}
\\
%%%%%%%%%%%%%%%%%%%%%%%%%%%%%%%%%%%%%%%%%%%
\multicolumn{3}{p{6.6cm}}{
    \includegraphics[width=\linewidth]{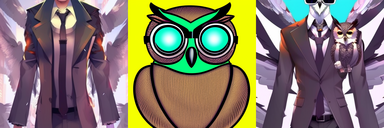}
    \newline
    \fontsize{7}{7}\selectfont
    an anthropomorphic owl, serious looking wearing mechanical sunglasses and grey suit, by kawacy, trending on pixiv, anime, furry art, trending on furaffinity, mafia member.
}
&
\multicolumn{3}{p{6.6cm}}{
    \includegraphics[width=\linewidth]{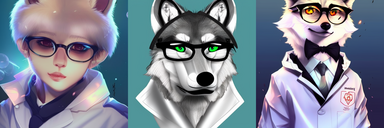}
    \newline
    \fontsize{7}{7}\selectfont
    anthropomorphic wolf with glasses wearing a lab coat, trending on artstation, trending on furaffinity, digital art, by kawacy, anime, furry art, warm light, backlighting, cartoon, concept art.
}
\\
%%%%%%%%%%%%%%%%%%%%%%%%%%%%%%%%%%%%%%%%%%%
\multicolumn{3}{p{6.6cm}}{
    \includegraphics[width=\linewidth]{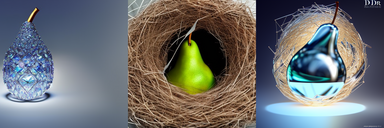}
    \newline
    \fontsize{7}{7}\selectfont
    crystal big pear in a nest, transparent, with light glares, reflections, photo realistic, photography, photorealism, ultra realistic, intricate, detail, rim light, depth of field, unreal engine, dslr, rtx, style swarovski, dior, faberge.
}
&
\multicolumn{3}{p{6.6cm}}{
    \includegraphics[width=\linewidth]{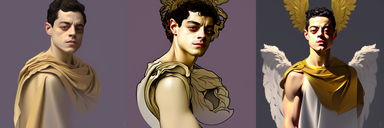}
    \newline
    \fontsize{7}{7}\selectfont
    rami malek as an angel in a golden toga, gray background, alphonse mucha, rhads, ross tran, artstation, artgerm, octane render, 1 6 k.
}
\\
\end{tabular}
\captionof{figure}{Qualitative examples from \Cref{tab:results_diffusionDB} (same seed used for a given prompt).}
\end{minipage}

%%%%%%%%%%%%%%%%%%%%%%%%%%%%%%%%%%%%%%%%%%%%%%%%%%%%%%%%%%
%%%%%%%%%%%%%%%%%%%%%%%%%%%%%%%%%%%%%%%%%%%%%%%%%%%%%%%%%%
%%%%%%%%%%%%%%%%%%%%%%%%%%%%%%%%%%%%%%%%%%%%%%%%%%%%%%%%%%

\clearpage
\section[
Qualitative analysis of \gls{MI} as an alignment measure
\newline
{
\fontsize{8}{8}
\it
\textcolor{blue!50}{
Expanded version of \Cref{fig:MI-qualitative} including all \acrshort{T2ICOMPBENCH} categories 
}
}
]{Qualitative analysis of \gls{MI} as an alignment measure
\label{app:MI-qualitative-all-categories}}
\begin{figure}[!h]
\centering
\hspace{0.01\textwidth}%\hfill
\minipage{0.18\textwidth}
{\small \textbf{Color binding}: \\
``\textit{A blue car and \\
a red horse}''}
\vspace{+1.5cm}
\endminipage
\minipage{0.14\textwidth}
\includegraphics[width=\linewidth]{arxiv_iclr25/images/a_blue_car_and_a_red_horse/a_blue_car_and_a_red_horse_0-onlyimg.png}
\vspace{-0.65cm}
\caption*{
    \hspace{-0.43em}
    \fontsize{6}{6}\selectfont
    \begin{tabular}{@{}r@{$\,=\,$}l@{}}
    \acrshort{BLIP} & 0.93 \\
    \acrshort{HPS} & 0.319 \\
    \gls{MI} & 36.28\\
    \end{tabular}
}
\endminipage
\hspace{0.01\textwidth}%\hfill
\minipage{0.14\textwidth}
\includegraphics[width=\linewidth]{arxiv_iclr25/images/a_blue_car_and_a_red_horse/a_blue_car_and_a_red_horse_1-onlyimg.png}
\vspace{-0.65cm}
\caption*{
    \hspace{-0.43em}
    \fontsize{6}{6}\selectfont
    \begin{tabular}{@{}r@{$\,=\,$}l@{}}
    \acrshort{BLIP} & 0.40 \\
    \acrshort{HPS}  & 0.310 \\
    \gls{MI}        & 24.56 \\
    \end{tabular}
}
\endminipage
\hspace{0.01\textwidth}%\hfill
\minipage{0.14\textwidth}
\includegraphics[width=\linewidth]{arxiv_iclr25/images/a_blue_car_and_a_red_horse/a_blue_car_and_a_red_horse_2-onlyimg.png}
\vspace{-0.65cm}
\caption*{
    \hspace{-0.43em}
    \fontsize{6}{6}\selectfont
    \begin{tabular}{@{}r@{$\,=\,$}l@{}}
    \acrshort{BLIP} & 0.17 \\
    \acrshort{HPS}  & 0.312 \\
    \gls{MI}        & 22.05 \\
    \end{tabular}
}
\endminipage
\hspace{0.01\textwidth}%\hfill
\minipage{0.14\textwidth}
\includegraphics[width=\linewidth]{arxiv_iclr25/images/a_blue_car_and_a_red_horse/a_blue_car_and_a_red_horse_3-onlyimg.png}
\vspace{-0.65cm}
\caption*{
    \hspace{-0.43em}
    \fontsize{6}{6}\selectfont
    \begin{tabular}{@{}r@{$\,=\,$}l@{}}
    \acrshort{BLIP} & 0.06 \\
    \acrshort{HPS}  & 0.263 \\ 
    \gls{MI}        & 15.44 \\
    \end{tabular}
}
\endminipage
\hspace{0.01\textwidth}%\hfill
\minipage{0.14\textwidth}
\includegraphics[width=\linewidth]{arxiv_iclr25/images/a_blue_car_and_a_red_horse/a_blue_car_and_a_red_horse_4-onlyimg.png}
\vspace{-0.65cm}
\caption*{
    \hspace{-0.43em}
    \fontsize{6}{6}\selectfont
    \begin{tabular}{@{}r@{$\,=\,$}l@{}}
    \acrshort{BLIP} & 0.05 \\
    \acrshort{HPS}  & 0.258 \\
    \gls{MI}        & 14.67\\
    \end{tabular}
}
\endminipage

\hspace{0.005\textwidth}%\hfill
\minipage{0.18\textwidth}
{\small \textbf{Texture binding}: \\
``\textit{A fabric dress and \\
a glass table}''}
\vspace{+1.5cm}
\endminipage
\minipage{0.14\textwidth}
\includegraphics[width=\linewidth]{arxiv_iclr25/images/a_fabric_dress_and_a_glass_table/a_fabric_dress_and_a_glass_table_0-onlyimg.png}
\vspace{-0.65cm}
\caption*{
    \hspace{-0.43em}
    \fontsize{6}{6}\selectfont
    \begin{tabular}{@{}r@{$\,=\,$}l@{}}
    \acrshort{BLIP} & 0.90 \\
    \acrshort{HPS}  & 0.257 \\
    \gls{MI}        & 44.6\\
    \end{tabular}
}
\endminipage
\hspace{0.01\textwidth}%\hfill
\minipage{0.14\textwidth}
\includegraphics[width=\linewidth]{arxiv_iclr25/images/a_fabric_dress_and_a_glass_table/a_fabric_dress_and_a_glass_table_1-onlyimg.png}
\vspace{-0.65cm}
\caption*{
    \hspace{-0.43em}
    \fontsize{6}{6}\selectfont
    \begin{tabular}{@{}r@{$\,=\,$}l@{}}
    \acrshort{BLIP} & 0.46 \\
    \acrshort{HPS}  & 0.213 \\
    \gls{MI}        & 28.1 \\
    \end{tabular}
}
\endminipage
\hspace{0.01\textwidth}%\hfill
\minipage{0.14\textwidth}
\includegraphics[width=\linewidth]{arxiv_iclr25/images/a_fabric_dress_and_a_glass_table/a_fabric_dress_and_a_glass_table_2-onlyimg.png}
\vspace{-0.65cm}
\caption*{
    \hspace{-0.43em}
    \fontsize{6}{6}\selectfont
    \begin{tabular}{@{}r@{$\,=\,$}l@{}}
    \acrshort{BLIP} & 0.17 \\
    \acrshort{HPS}  & 0.201 \\
    \gls{MI}        & 19.86\\
    \end{tabular}
}
\endminipage
\hspace{0.01\textwidth}%\hfill
\minipage{0.14\textwidth}
\includegraphics[width=\linewidth]{arxiv_iclr25/images/a_fabric_dress_and_a_glass_table/a_fabric_dress_and_a_glass_table_3-onlyimg.png}
\vspace{-0.65cm}
% \caption*{\acrshort{BLIP}=0.12 \acrshort{HPS}=0.231 \gls{MI}=15.41}
\caption*{
    \hspace{-0.43em}
    \fontsize{6}{6}\selectfont
    \begin{tabular}{@{}r@{$\,=\,$}l@{}}
    \acrshort{BLIP} & 0.12 \\
    \acrshort{HPS}  & 0.231 \\
    \gls{MI}        & 15.41\\
    \end{tabular}
}
\endminipage
\hspace{0.01\textwidth}%\hfill
\minipage{0.14\textwidth}
\includegraphics[width=\linewidth]{arxiv_iclr25/images/a_fabric_dress_and_a_glass_table/a_fabric_dress_and_a_glass_table_4-onlyimg.png}
\vspace{-0.65cm}
\caption*{
    \hspace{-0.43em}
    \fontsize{6}{6}\selectfont
    \begin{tabular}{@{}r@{$\,=\,$}l@{}}
    \acrshort{BLIP} & 0.07 \\
    \acrshort{HPS}  & 0.295 \\
    \gls{MI}        & 9.34
    \end{tabular}
}
\endminipage

\hspace{0.005\textwidth}%\hfill
\minipage{0.18\textwidth}
{\small \textbf{Shape binding}:\\
``\textit{A round bag and \\
a rectangular \\
wallet}''}
\vspace{+1.5cm}
\endminipage
\minipage{0.14\textwidth}
\includegraphics[width=\linewidth]{arxiv_iclr25/images/a_round_bag_and_a_rectangular_wallet/a_round_bag_and_a_rectangular_wallet-1-18.609375.png}
\vspace{-0.65cm}
\caption*{
    \hspace{-0.43em}
    \fontsize{6}{6}\selectfont
    \begin{tabular}{@{}r@{$\,=\,$}l@{}}
    \acrshort{BLIP} & 0.82 \\
    \acrshort{HPS}  & 0.262 \\
    \gls{MI}        & 18.61\\
    \end{tabular}
}
\endminipage
\hspace{0.01\textwidth}%\hfill
\minipage{0.14\textwidth}
\includegraphics[width=\linewidth]{arxiv_iclr25/images/a_round_bag_and_a_rectangular_wallet/a_round_bag_and_a_rectangular_wallet-5-17.15625.png}
\vspace{-0.65cm}
\caption*{
    \hspace{-0.43em}
    \fontsize{6}{6}\selectfont
    \begin{tabular}{@{}r@{$\,=\,$}l@{}}
    \acrshort{BLIP} & 0.64 \\
    \acrshort{HPS}  & 0.247 \\
    \gls{MI}        & 17.16\\
    \end{tabular}
}
\endminipage
\hspace{0.01\textwidth}%\hfill
\minipage{0.14\textwidth}
\includegraphics[width=\linewidth]{arxiv_iclr25/images/a_round_bag_and_a_rectangular_wallet/a_round_bag_and_a_rectangular_wallet-11-14.8359375.png}
\vspace{-0.65cm}
\caption*{
    \hspace{-0.43em}
    \fontsize{6}{6}\selectfont
    \begin{tabular}{@{}r@{$\,=\,$}l@{}}
    \acrshort{BLIP} & 0.27 \\
    \acrshort{HPS}  & 0.262 \\
    \gls{MI}        & 14.84\\
    \end{tabular}
}
\endminipage
\hspace{0.01\textwidth}%\hfill
\minipage{0.14\textwidth}
\includegraphics[width=\linewidth]{arxiv_iclr25/images/a_round_bag_and_a_rectangular_wallet/a_round_bag_and_a_rectangular_wallet-18-12.5.png}
\vspace{-0.65cm}
\caption*{
    \hspace{-0.43em}
    \fontsize{6}{6}\selectfont
    \begin{tabular}{@{}r@{$\,=\,$}l@{}}
    \acrshort{BLIP} & 0.24 \\
    \acrshort{HPS}  & 0.216 \\
    \gls{MI}        & 12.50
    \end{tabular}
}
\endminipage
\hspace{0.005\textwidth}%\hfill
\minipage{0.14\textwidth}
\includegraphics[width=\linewidth]{arxiv_iclr25/images/a_round_bag_and_a_rectangular_wallet/a_round_bag_and_a_rectangular_wallet-22-11.5703125.png}
\vspace{-0.65cm}
\caption*{
    \hspace{-0.43em}
    \fontsize{6}{6}\selectfont
    \begin{tabular}{@{}r@{$\,=\,$}l@{}}
    \acrshort{BLIP} & 0.01 \\
    \acrshort{HPS}  & 0.160 \\
    \gls{MI}        & 11.57    
    \end{tabular}
}
\endminipage % 0.0084

\hspace{0.01\textwidth}%\hfill
\minipage{0.18\textwidth}
{\small \textbf{Spatial relation}: \\
``\textit{a man on \\
the top of \\
a turtle}''}
\vspace{+1.5cm}
\endminipage
\minipage{0.14\textwidth}
\includegraphics[width=\linewidth]{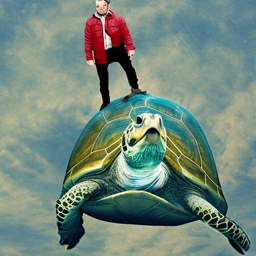}
\vspace{-0.65cm}
\caption*{
    \hspace{-0.43em}
    \fontsize{6}{6}\selectfont
    \begin{tabular}{@{}r@{$\,=\,$}l@{}}
    \acrshort{UNIDET} & 1.00 \\
    \acrshort{HPS}  &   0.301 \\
    \gls{MI}        &   36.41 \\
    \end{tabular}
}
\endminipage
\hspace{0.01\textwidth}%\hfill
\minipage{0.14\textwidth}
\includegraphics[width=\linewidth]{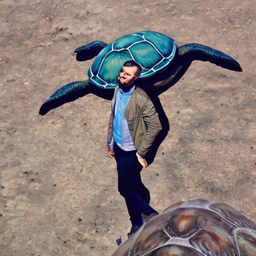}
\vspace{-0.65cm}
\caption*{
    \hspace{-0.43em}
    \fontsize{6}{6}\selectfont
    \begin{tabular}{@{}r@{$\,=\,$}l@{}}
    \acrshort{UNIDET} & 0.90 \\
    \acrshort{HPS}  &   0.288 \\
    \gls{MI}        &   19.11 \\
    \end{tabular}
}
\endminipage
\hspace{0.01\textwidth}%\hfill
\minipage{0.14\textwidth}
\includegraphics[width=\linewidth]{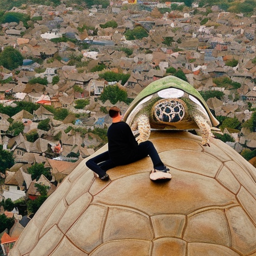}
\vspace{-0.65cm}
\caption*{
    \hspace{-0.43em}
    \fontsize{6}{6}\selectfont
    \begin{tabular}{@{}r@{$\,=\,$}l@{}}
    \acrshort{UNIDET} & 0.79 \\
    \acrshort{HPS}  &   0.272 \\
    \gls{MI}        &   9.93 \\
    \end{tabular}
}
\endminipage
\hspace{0.01\textwidth}%\hfill
\minipage{0.14\textwidth}
\includegraphics[width=\linewidth]{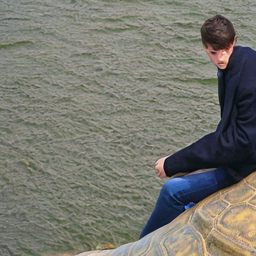}
\vspace{-0.65cm}
\caption*{
    \hspace{-0.43em}
    \fontsize{6}{6}\selectfont
    \begin{tabular}{@{}r@{$\,=\,$}l@{}}
    \acrshort{UNIDET} & 0.68 \\
    \acrshort{HPS}  &   0.231 \\
    \gls{MI}        &   4.90 \\
    \end{tabular}
}
\endminipage
\hspace{0.01\textwidth}%\hfill
\minipage{0.14\textwidth}
\includegraphics[width=\linewidth]{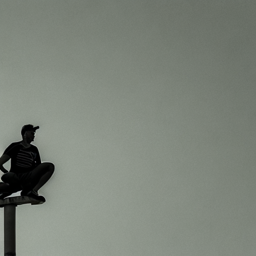}
\vspace{-0.65cm}
\caption*{
    \hspace{-0.43em}
    \fontsize{6}{6}\selectfont
    \begin{tabular}{@{}r@{$\,=\,$}l@{}}
    \acrshort{UNIDET} & 0.00 \\
    \acrshort{HPS}  &   0.180 \\
    \gls{MI}        &   4.81 \\
    \end{tabular}
}
\endminipage

\hspace{0.005\textwidth}%\hfill
\minipage{0.18\textwidth}
{\small \textbf{Non-spatial relation}: \\
``\textit{A dog is chasing \\
after a ball and \\
wagging its tail}''}
\vspace{+0.8cm}
\endminipage
\minipage{0.14\textwidth}
\includegraphics[width=\linewidth]{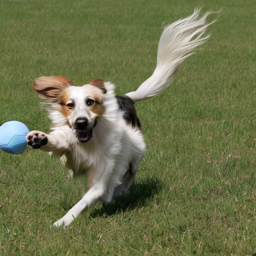}
\vspace{-0.65cm}
\caption*{
    \hspace{-0.43em}
    \fontsize{6}{6}\selectfont
    \begin{tabular}{@{}r@{$\,=\,$}l@{}}
    \acrshort{BLIP} & 0.97 \\
    \acrshort{HPS}  & 0.290 \\
    \gls{MI}        &  15.82 \\
    \end{tabular}
}
\endminipage
\hspace{0.01\textwidth}%\hfill
\minipage{0.14\textwidth}
\includegraphics[width=\linewidth]{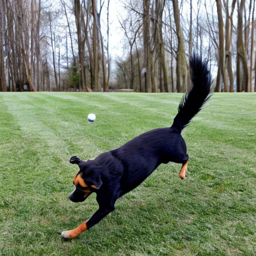}
%\label{fig:awesome_image2}
\vspace{-0.65cm}
\caption*{
    \hspace{-0.43em}
    \fontsize{6}{6}\selectfont
    \begin{tabular}{@{}r@{$\,=\,$}l@{}}
    \acrshort{BLIP} & 0.96 \\
    \acrshort{HPS}  & 0.258 \\
    \gls{MI}        &  11.39 \\
    \end{tabular}
}
\endminipage
\hspace{0.01\textwidth}%\hfill
\minipage{0.14\textwidth}
\includegraphics[width=\linewidth]{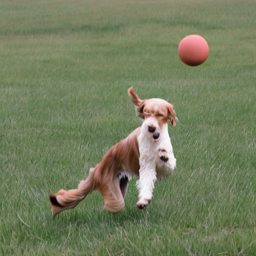}
\vspace{-0.65cm}
\caption*{
    \hspace{-0.43em}
    \fontsize{6}{6}\selectfont
    \begin{tabular}{@{}r@{$\,=\,$}l@{}}
    \acrshort{BLIP} &  0.92 \\
    \acrshort{HPS}  & 0.251  \\
    \gls{MI}        & 8.99  \\
    \end{tabular}
}
\endminipage
\hspace{0.01\textwidth}%\hfill
\minipage{0.14\textwidth}
\includegraphics[width=\linewidth]{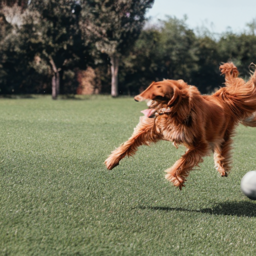}
\vspace{-0.65cm}
\caption*{
    \hspace{-0.43em}
    \fontsize{6}{6}\selectfont
    \begin{tabular}{@{}r@{$\,=\,$}l@{}}
    \acrshort{BLIP} & 0.88   \\
    \acrshort{HPS}  & 0.236  \\
    \gls{MI}        &  7.88 \\
    \end{tabular}
}
\endminipage
\hspace{0.01\textwidth}%\hfill
\minipage{0.14\textwidth}
\includegraphics[width=\linewidth]{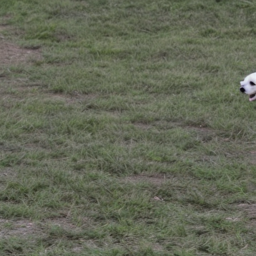}
\vspace{-0.65cm}
\caption*{
    \hspace{-0.43em}
    \fontsize{6}{6}\selectfont
    \begin{tabular}{@{}r@{$\,=\,$}l@{}}
    \acrshort{BLIP} &  0.09  \\
    \acrshort{HPS}  & 0.152  \\
    \gls{MI}        & 5.02 \\
    \end{tabular}
}
\endminipage

\hspace{0.005\textwidth}%\hfill
\minipage{0.18\textwidth}
{\small \textbf{Complex prompt}: \\
``\textit{The red hat \\
was on top of \\
the brown coat}''}
\vspace{+1.2cm}
\endminipage
\minipage{0.14\textwidth}
\includegraphics[width=\linewidth]{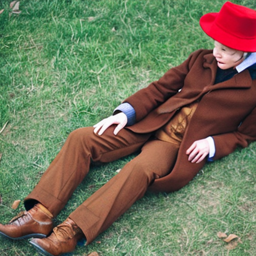}
\vspace{-0.65cm}
\caption*{
    \hspace{-0.43em}
    \fontsize{6}{6}\selectfont
    \begin{tabular}{@{}r@{$\,=\,$}l@{}}
    \acrshort{BLIP} & 0.91 \\
    \acrshort{HPS}  & 0.259 \\
    \gls{MI}        & 18.18 \\
    \end{tabular}
}
\endminipage
\hspace{0.01\textwidth}%\hfill
\minipage{0.14\textwidth}
\includegraphics[width=\linewidth]{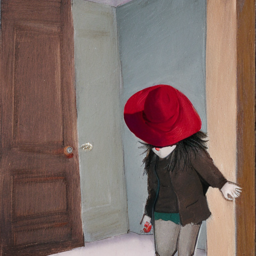}
%\label{fig:awesome_image2}
\vspace{-0.65cm}
\caption*{
    \hspace{-0.43em}
    \fontsize{6}{6}\selectfont
    \begin{tabular}{@{}r@{$\,=\,$}l@{}}
    \acrshort{BLIP} & 0.61 \\
    \acrshort{HPS}  &  0.237 \\
    \gls{MI}        &  6.36 \\
    \end{tabular}
}
\endminipage
\hspace{0.01\textwidth}%\hfill
\minipage{0.14\textwidth}
\includegraphics[width=\linewidth]{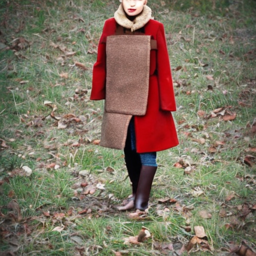}
\vspace{-0.65cm}
\caption*{
    \hspace{-0.43em}
    \fontsize{6}{6}\selectfont
    \begin{tabular}{@{}r@{$\,=\,$}l@{}}
    \acrshort{BLIP} & 0.50 \\
    \acrshort{HPS}  & 0.212 \\
    \gls{MI}        & 5.43 \\
    \end{tabular}
}
\endminipage
\hspace{0.01\textwidth}%\hfill
\minipage{0.14\textwidth}
\includegraphics[width=\linewidth]{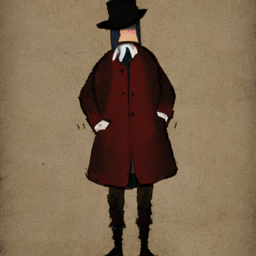}
\vspace{-0.65cm}
\caption*{
    \hspace{-0.43em}
    \fontsize{6}{6}\selectfont
    \begin{tabular}{@{}r@{$\,=\,$}l@{}}
    \acrshort{BLIP} & 0.28 \\
    \acrshort{HPS}  & 0.204 \\
    \gls{MI}        & 5.27 \\
    \end{tabular}
}
\endminipage
\hspace{0.01\textwidth}%\hfill
\minipage{0.14\textwidth}
\includegraphics[width=\linewidth]{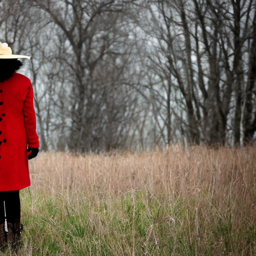}
\vspace{-0.65cm}
\caption*{
    \hspace{-0.43em}
    \fontsize{6}{6}\selectfont
    \begin{tabular}{@{}r@{$\,=\,$}l@{}}
    \acrshort{BLIP} & 0.08 \\
    \acrshort{HPS}  &  0.189 \\
    \gls{MI}        &  4.06 \\
    \end{tabular}
}
\endminipage

\caption{Qualitative analysis of \gls{MI} as an alignment measure (all metrics decrease from left to right). 
}
\label{fig:MI-qualitative-all-categories}
\end{figure}

\vspace{-1.5mm}
\Cref{fig:MI-qualitative-all-categories} is expanding \Cref{fig:MI-qualitative} to include qualitative examples for all categories in \acrshort{T2ICOMPBENCH}.

\vspace{-1mm}

\clearpage
\section[
\acrshort{BLIP}, \acrshort{HPS} and \acrshort{MI} score distributions
\newline
{
\fontsize{8}{8}
\it
\textcolor{blue!50}{
\it Comparing distributions and \acrshort{MI} rank $\cdot$ Related to results in \Cref{sec:mialign}
}}
]{\acrshort{BLIP}, \acrshort{HPS} and \acrshort{MI} score distributions
\label{app:score-distributions-comparison}}
\begin{figure}[!h]
\centering
\includegraphics[width=0.7\textwidth]{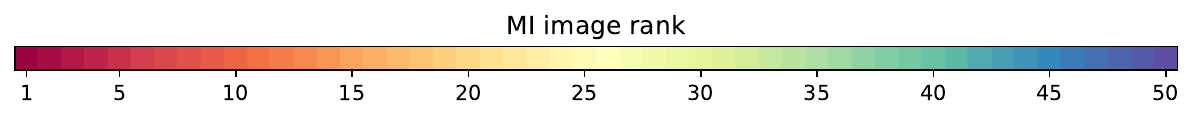}
\includegraphics[width=\textwidth]{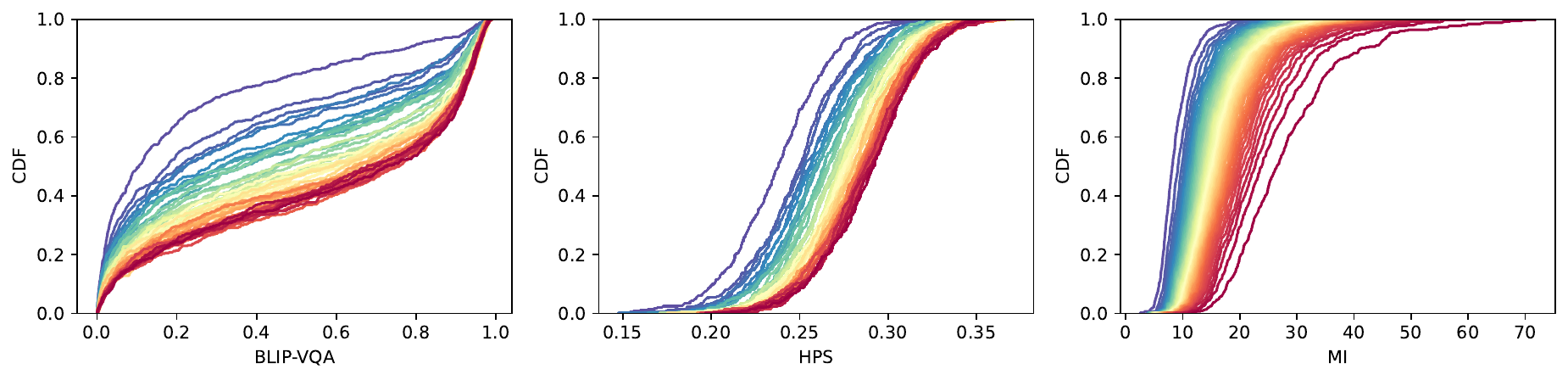}
\caption{CDF of alignment scores. Color reflect images rank based on \acrshort{MI}.
\label{fig:cdfs_scores}}
\end{figure}

The analysis presented in \Cref{sec:mialign} shows that \acrshort{BLIP}, \acrshort{HPS} and \acrshort{MI}
relate to each other. However, two aspects not discussed in \Cref{sec:method} are ($i$)
the support of each metric and ($ii$) how the distribution of the scores compare between well and poor aligned 
images. In this ablation we address both aspects using the following protocol.

We considered all 700 training prompts for the color category (the consideration presented
in this ablation extends to the other \acrshort{T2ICOMPBENCH} categories too), we generated 50 images for each prompt, and 
computed the 3 metrics for each of the 50 images. Last, for each prompt, we rank the images based on \acrshort{MI} (1:highest, 50:lowest) 
-- overall we obtained a 700 prompts $\times$ 50 images $\times$ 4 (3 metrics + 1 rank) tensor.  

We then investigated if/how the \acrshort{MI} rank affects the distribution of the scores for \acrshort{BLIP} and \acrshort{HPS}.
Intuitively, given the highest-ranked (viz lowest-ranked) images based on MI, also \acrshort{BLIP} and \acrshort{HPS} should show very high values
(viz low values). In practice, we first reordered the scores of the three metrics for each prompt based the \acrshort{MI} rank
and then we derived 50 distributions for each metric, one for each column in the tensor collecting the scores of each metric.
\Cref{fig:cdfs_scores} shows the obtained distributions color coded 
based on the \acrshort{MI} rank.

Considering the metrics support, we can notice a few differences among the three metrics.
Specifically, \acrshort{BLIP} is in the [0,1] range and for all rank values,
the whole support is always used. Conversely, despite \acrshort{HPS} is also in the [0, 1]
range,\footnote{\acrshort{HPS} is defined as the cosine similarity between image and text embeddings, similarly to CLIP. As such, theoretically, the score is in [-1, 1] range. However, in practice, and for the \acrshort{T2ICOMPBENCH} dataset, the score is effectively only in the [0, 1] range.} the actual support is more skewed -- this corroborates the discussion presented in \Cref{app:hps_range}.
Last, while \acrshort{MI} is unbounded, the scores are mostly contained in the [0-40] range. 

Considering the relationship between the rank and the scores, all metrics show very similar patterns. 
Specifically, all distributions are very smooth no matter the rank.
Moreover, as expected, for all metrics the distributions smoothly shift horizontally with respect to their rank
-- the color gradient separates very well red/high rank, yellow/middle rank, blue/low rank.

The kendal $\tau$ analysis reported in \Cref{sec:mialign} considers the 1$^{\text{st}}$, 25$^{\text{th}}$, 50$^{\text{th}}$ image
for a prompt, selected by ranking the images based on their \acrshort{MI} score. This is consistent with the analysis
presented in \Cref{fig:cdfs_scores} and based on the figure 
we argue that our selection of 3 pictures (having the highest, mid, lowest scores for each prompt) 
is a reasonable choice for the results reported in \Cref{sec:mialign} as they are representative of the spectrum of values observed by the metrics.

\end{document}